%% file: main.tex
\documentclass[fleqn,10pt]{article}

\usepackage[margin=2.5cm]{geometry}
\usepackage{graphicx}
\usepackage{booktabs}
\usepackage{amsmath,amssymb}
\usepackage[numbers,sort&compress]{natbib}
\usepackage{hyperref}
\hypersetup{colorlinks=true, linkcolor=blue!60!black, citecolor=blue!60!black, urlcolor=blue!60!black}
\usepackage{xcolor}
\usepackage{enumitem}
\usepackage{caption}
\usepackage{subcaption}
\usepackage{tabularx}
\usepackage{array}
\usepackage{multirow}
\usepackage{setspace}
\usepackage{authblk}
\usepackage{placeins}  
\usepackage{longtable}
\usepackage{listings}
\usepackage{pdflscape}

\lstdefinestyle{systemprompt}{
    basicstyle=\footnotesize\ttfamily,
    breaklines=true, breakatwhitespace=false, frame=single,
    backgroundcolor=\color{gray!10}, xleftmargin=1em, framexleftmargin=0.5em,
    columns=fullflexible, keepspaces=true, showstringspaces=false,
    aboveskip=1em, belowskip=1em
}
\lstdefinestyle{algorithm}{
    basicstyle=\footnotesize\ttfamily,
    breaklines=true, breakatwhitespace=false, frame=single,
    backgroundcolor=\color{gray!5}, xleftmargin=1em, framexleftmargin=0.5em,
    columns=fullflexible, keepspaces=true, showstringspaces=false,
    aboveskip=1em, belowskip=1em,
    literate={→}{$\rightarrow$}1 {←}{$\leftarrow$}1 {≥}{$\geq$}1 {≤}{$\leq$}1 {∈}{$\in$}1 {∅}{$\emptyset$}1 {∪}{$\cup$}1 {∃}{$\exists$}1 {▷}{$\triangleright$}1 {─}{-}1
}

\newcommand{\hegtkg}{HEG-TKG}
\newcommand{\provgap}{\textit{Provenance Gap}}

\title{%
  \textbf{The Provenance Gap in Clinical AI: Evidence-Traceable Temporal Knowledge Graphs for Rare Disease Reasoning}}

\author[1,*]{Md Shamim Ahmed}
\author[2]{Maja Dusanic}
\author[2]{Moritz Nikolai Kirschner}
\author[2]{Elisabeth Nyoungui}
\author[2]{Jana Zsch\"{u}ntzsch}
\author[1,$\dagger$]{Lukas Galke Poech}
\author[1,$\dagger$]{Richard R\"{o}ttger}
\affil[1]{Department of Mathematics and Computer Science, University of Southern Denmark, Campusvej 55, 5230 Odense, Denmark}
\affil[2]{Universit\"{a}tsmedizin G\"{o}ttingen, Robert-Koch-Stra\ss e 40, 37075 G\"{o}ttingen, Germany}
\affil[*]{Correspondence: shamim@imada.sdu.dk}
\affil[$\dagger$]{These authors jointly supervised this work}

\date{}

\begin{document}
\maketitle

\begin{abstract}
\noindent
Clinicians cannot act on AI recommendations they cannot verify. We evaluated
five frontier models on 36 clinician-validated scenarios across three
neuromuscular disease pairs (myasthenia gravis/Lambert-Eaton, Duchenne/Becker,
CIDP/Guillain-Barré). Without citation prompts, none returned a clinically
relevant PubMed identifier. With citation prompts, the strongest model reached
15.3\% relevance; the rest cited real but unrelated papers. This is the
\emph{Provenance Gap}.

HEG-TKG is a temporal knowledge graph from 4,512 PubMed records and curated
guidelines, with cross-source evidence tiers and stage-anchored milestones.
Holding synthesis fixed across three arms, HEG-TKG matched baseline feature
coverage, hit 100\% PMID verifiability, and 99\% claim-support agreement on a
blinded 200-claim audit. A Guideline-RAG arm with the same sources returned
zero verifiable citations - traceability requires graph structure, not just
document access. Three neurologists rated HEG-TKG 0.67 to 1.65 Likert points higher on
verifiability (all $q < 0.05$, BH-corrected); safety and completeness held. LLM judges missed
this gap entirely. Counterfactual injection: 80\% resistance, all
traceable.
\end{abstract}


\section{Introduction}

A neurologist who receives an AI-generated differential for myasthenia gravis
versus Lambert-Eaton needs to do three things before acting on it: trace each
clinical claim to its evidentiary source, confirm that the cited evidence
actually supports the recommendation, and decide whether the system's
confidence is calibrated to the disease territory. Without that audit trail,
the recommendation is unverifiable, and the cost of acting on an unverified
output is concrete. Misdiagnosing MG as LEMS leads to incorrect immunotherapy.
Confusing DMD with BMD delays the steroid-timing decision by years, with
measurable downstream effects on ambulation loss and cardiac decline~\cite{bushby2010dmd}.
These are the consequences clinicians weigh, and they are the reason a
black-box differential - however accurate on benchmarks does not clear the
bar for routine clinical use.

Frontier large language models clear the accuracy bar comfortably. Med-PaLM 2
reaches 86.5\% on MedQA\cite{singhal2023medpalm2}, GPT-4 passes all three parts of the United States
Medical Licensing Examination\cite{nori2023gpt4}, and clinical deployment is moving fast
across specialties\cite{thirunavukarasu2023llmmed}. Accuracy is not the bottleneck, and this paper is not
about accuracy.

The EU AI Act\cite{euaiact2024}, FDA clinical decision support guidance\cite{fdacds2026}, and WHO\cite{whoai2021} all mandate independent verification of AI-generated clinical recommendations. 
At the same time, patient scenarios constitute protected health information under GDPR and HIPAA\cite{murdoch2021privacy}, meaning that cloud API calls expose data outside institutional control. 
A clinical AI system needs verifiability and data protection. Current LLMs deliver neither.
HEG-TKG resolves both: structured evidence provenance for verifiability, open-source synthesis models behind institutional firewalls for data protection.

Citation failures in LLMs are worse than most studies acknowledge.
SourceCheckup found 50-90\% of LLM responses unsupported by their cited sources\cite{wu2025sourcecheckup}; VeriFact showed automated fact verification can achieve 92.7\% clinician agreement for EHR claims\cite{chung2025verifact}.
The verifiability problem extends upstream of model outputs as well. Gibson et al.~\cite{Gibson2026-dm} recently documented that 124 peer-reviewed clinical-prediction studies were developed on Kaggle-hosted health datasets of unverified provenance, with statistical anomalies suggesting partial fabrication\cite{nature2026dubiousdata}. Three of these models have entered clinical practice, one is referenced in a medical device patent, and the dataset-derived literature has accumulated 1,529 citations across 86 review articles~\cite{Gibson2026-dm}. Their finding identifies the upstream face of the same failure mode we address downstream: provenance asserted but not verifiable. Where their work documents training-data opacity propagating into deployed decision support, ours measures and addresses the output-side counterpart - model claims that cite real but unrelated sources, indistinguishable from grounded claims under surface audit.
Autoregressive generation produces hallucinations by design\cite{ji2023hallucination,huang2023hallucination}: what a model can recall is not the same as what it can prove.
Using a second LLM to verify citations from the first does not work. Our experiments show LLM judges rate fabricated citations as valid because they lack external verification data. The Provenance Gap cannot be solved within the LLM ecosystem itself.

This matters most in rare diseases, where the clinician seeing such a patient has probably never encountered the condition before.
Over 6,000 rare diseases affect approximately 300 million patients globally\cite{wakap2020raredisease}, most diagnosed by non-specialists after years-long diagnostic odysseys.
We selected three phenotypically confusable disease pairs - myasthenia gravis versus Lambert-Eaton\cite{gilhus2016mg}, Duchenne versus Becker muscular dystrophy\cite{bushby2010dmd}, and CIDP versus Guillain-Barr\'{e}\cite{shahrizaila2021gbs} - because all three are clinically similar at presentation but diverge over their longitudinal course, requiring temporal reasoning that existing knowledge graphs do not support.
PrimeKG\cite{chandrasekaran2023primekg}, UMLS\cite{bodenreider2004umls}, and DrugBank\cite{wishart2018drugbank} encode static relationships: dystrophin is associated with DMD, but not that Gowers' sign appears at age 3-5 while ambulation loss follows at 9-13.

We define \provgap{} as the gap between a system's clinical feature coverage and its \emph{reliable} evidence traceability (adjusted for citation quality; see Methods). 
We tested five frontier LLMs spanning three providers across 36 clinician-validated scenarios and verified every PubMed identifier against the E-utilities API. 
In vanilla mode, all models produced zero clinically relevant PMIDs. The best model, when explicitly told to cite, managed 15.3\%. 
We present \hegtkg{}, a system that grounds clinical reasoning in temporal knowledge graphs with a two-tier architecture: a curated Tier~1 backbone from GeneReviews\cite{genereviews2024}, OMIM\cite{omim2024}, Orphanet\cite{orphanet2024}, and CDC guidelines\cite{cdc_dmd2018}, augmented by Tier~2 extraction from 4,512 PubMed records (987 unique source PMIDs), and anchored to 1,280 disease-trajectory milestones. 
To our knowledge, no existing biomedical knowledge graph captures these temporal trajectories\cite{tgb2024}; complementary work on temporal reasoning in longitudinal EHR data\cite{cui2025timer} improves LLM time-awareness via instruction tuning but does not anchor claims to structured disease-trajectory milestones.
Where standard retrieval-augmented generation\cite{lewis2020rag,galkin2024selfrag} hands the model a bag of text chunks, and KG-grounded diagnosis approaches such as DR.KNOWS\cite{gao2025drknows} retrieve UMLS reasoning paths for EHR-based prediction, \hegtkg{} maps every claim to a programmatically verified PMID, assigns a quality tier reflecting cross-source consensus, and places it within a disease-trajectory timeline. As our three-arm comparison confirms, this structured grounding maintains clinical feature coverage (Table~\ref{tab:threearmmain}).
The strongest recent system in this space, DeepRare (Zhao et al.\cite{zhao2026deeprare}), deploys a multi-agent architecture across 3,134 rare diseases spanning 14 specialties and reports 57.18\% Recall@1 with 95.4\% physician-rated reference accuracy. Zhao et al.\ explicitly identify ``hallucinated references'' (plausible but nonexistent URLs) and irrelevant citations drawn from incorrect diagnoses as recurring failure modes\cite{zhao2026deeprare}, consistent with our characterisation of reactive agentic retrieval as Level~2 citation quality: identifier-based but not reproducible across sessions. Song et al.\cite{song2025fpkg} present a static graph-RAG system for rare genetic diseases (6,143 nodes) without temporal anchoring or cross-source quality stratification. ZebraMap\cite{islam2025zebramap} aggregates 1,727 Orphanet rare diseases from 36,131 case reports into structured fields via an automated LLM pipeline but likewise does not grade evidence quality or anchor claims to disease-trajectory milestones. None of these systems verify citations programmatically against primary literature; their breadth-first scope (thousands of diseases, reactively or structurally grounded) and our depth-first scope (six diseases, fully traceable and temporally anchored) address complementary axes of the rare-disease AI problem.
We take a depth-first approach: six diseases with \emph{fully traceable} temporal reasoning, showing that the architecture generalises.

In this work, we make five contributions: (1) a formal definition and empirical measurement of the \emph{Provenance Gap} across five frontier LLMs; (2) HEG-TKG, a temporal knowledge graph architecture with fully verified citations, quality-stratified evidence (GOLD/SILVER/BRONZE), and disease-trajectory anchoring; (3) the first demonstration that LLM judges cannot detect fabricated citations without external audit data; (4) a counterfactual safety test showing 80\% error resistance with 100\% detectability; (5) the temporal KG as a reusable resource covering six rare neuromuscular diseases (5,481 nodes; 1,280 temporal anchors).

\section{Results}

\subsection{Frontier LLMs produce zero verifiable citations}

We evaluated five frontier LLMs spanning three providers across 36 clinician-validated scenarios for three rare neuromuscular disease pairs (Table~\ref{tab:citation}). 
Each model generated clinical outputs (differential diagnoses, temporal trajectories, treatment rationales, safety-critical recommendations) for the same scenarios. We then verified every PubMed identifier (PMID) cited against the PubMed E-utilities API\cite{sayers2022eutils}, classifying each as clinically relevant, real but unrelated, or non-existent.

\begin{table}[ht]
\centering
\caption{\textbf{Citation verifiability across five frontier LLMs and \hegtkg{}.} PMIDs = unique PubMed identifiers extracted from outputs. Relevant = PMID resolves to a neuromuscular disease publication. Wrong Field = PMID exists in PubMed but covers an unrelated specialty. Not Found = PMID does not exist.}
\label{tab:citation}
\small
\begin{tabular}{llrccc}
\toprule
\textbf{Model} & \textbf{Mode} & \textbf{PMIDs} & \textbf{Relevant} & \textbf{Wrong Field} & \textbf{Not Found} \\
\midrule
GPT-4.1            & Vanilla            & 2   & 0            & 2 (100\%)   & 0          \\
GPT-5.4            & Vanilla            & 0   & ---          & ---         & ---        \\
Claude Sonnet 4.6  & Vanilla            & 0*  & ---          & ---         & ---        \\
DeepSeek-v3        & Vanilla            & 0   & ---          & ---         & ---        \\
\midrule
Claude Opus 4.6    & Citation-prompted  & 577 & 88 (15.3\%)  & 455 (78.9\%)& 34 (5.9\%) \\
Claude Sonnet 4.6  & Citation-prompted  & 415 & 49 (11.8\%)  & 331 (79.8\%)& 35 (8.4\%) \\
DeepSeek-v3        & Citation-prompted  & 163 & 17 (10.4\%)  & 135 (82.8\%)& 11 (6.7\%) \\
\midrule
\textbf{\hegtkg{}} & \textbf{KG-grounded} & \textbf{203} & \textbf{203 (100\%)} & \textbf{0} & \textbf{0} \\
\bottomrule
\end{tabular}

\smallskip
\noindent\footnotesize{*Claude Sonnet~4.6 in vanilla mode produced 89 unique author-year references (e.g., ``Wolfe et al., 2016'') instead of PMIDs; none were verifiable to a specific publication. All unique PMIDs (1,147 total across experiments) were verified against the PubMed E-utilities API. The \hegtkg{} knowledge graphs were constructed from curated Tier~1 sources (GeneReviews, OMIM, Orphanet, CDC; 101 protected edges) and 4,512 PubMed records (2,011 after deduplication and screening), yielding 987 unique source PMIDs; the 203 cited in outputs reflect query-relevant subgraph retrieval. 22 of 36 scenarios were refined by a clinical collaborator (J.Z., UMG G\"{o}ttingen).}
\end{table}

\begin{figure*}[!htbp]
\centering
\includegraphics[width=\textwidth]{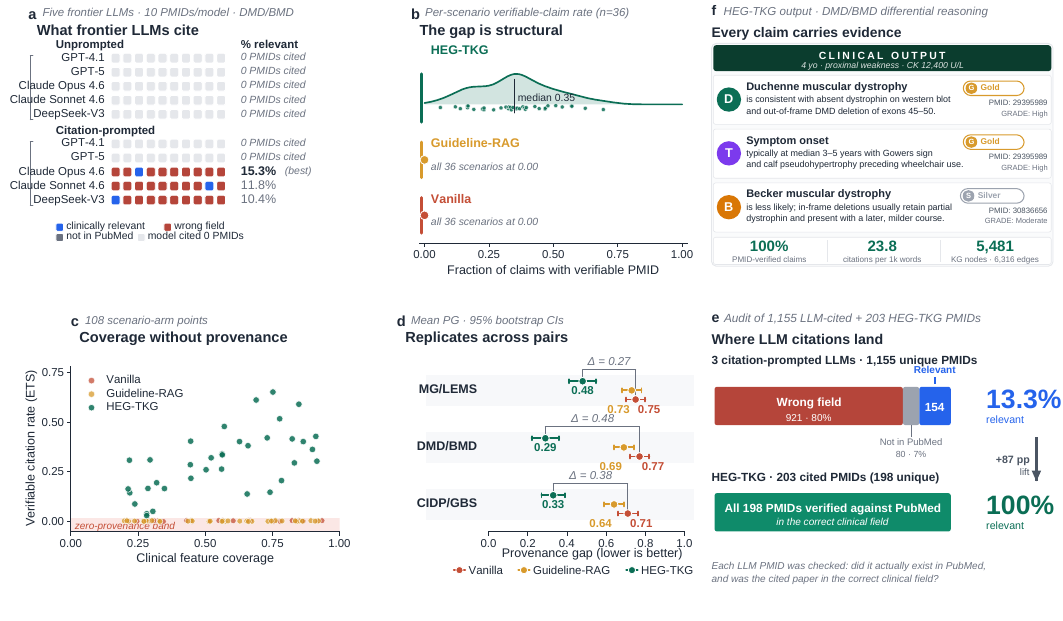}
\caption{\textbf{The Provenance Gap: frontier LLMs cite confidently, but rarely correctly.} (\textbf{a}) Citation audit of five frontier LLMs in unprompted and citation-prompted conditions; ten sampled PMIDs per row coloured by verification outcome. Citation-prompting reaches at best 15.3\% relevant (Claude Opus 4.6); the majority resolve to real papers in unrelated fields. (\textbf{b}) Per-scenario fraction of claims with verifiable PMIDs across 36 scenarios per arm: Vanilla and Guideline-RAG collapse to zero, \hegtkg{} distributes (median 0.35). The gap is prompt- and model-agnostic. (\textbf{c}) Reliable~ETS vs Clinical Feature Coverage across 108 scenario-arm points; Vanilla and Guideline-RAG sit on the zero-provenance band despite comparable coverage. (\textbf{d}) Mean Provenance Gap per disease pair (95\% bootstrap CIs, 10{,}000 resamples); the Vanilla$\rightarrow$\hegtkg{} $\Delta$ replicates across MG/LEMS, DMD/BMD, and CIDP/GBS. (\textbf{e}) Composition of 1{,}155 unique PMIDs cited by three citation-prompted LLMs (13.3\% relevant) versus 203 cited by \hegtkg{} (198 unique, 100\% verified in the correct clinical field). (\textbf{f}) A representative \hegtkg{} clinical output: every claim carries an inline PMID, evidence tier (Gold/Silver/Bronze), and GRADE rating.}
\label{fig:provenance}
\end{figure*}

All four vanilla models produced zero clinically relevant PMIDs (Figure~\ref{fig:provenance}). GPT-4.1 generated only 2 unique PMIDs across 36 scenarios that resolved to unrelated fields. GPT-5.4 and DeepSeek-v3 produced zero PMIDs altogether. Claude Sonnet~4.6 generates author-year references (e.g., ``Gilhus NE, Neurology 2016'') that look authoritative but cannot be resolved to a specific publication without manually searching PubMed; in practice, a single ambiguous reference (``Wolfe et al., 2016'') returns over 600 PubMed search results, costing approximately 2-5 minutes per claim to disambiguate.

We did not include the agentic rare-disease system DeepRare~\cite{zhao2026deeprare} in this
audit because its task format (HPO-term or VCF input, top-K disease ranking
output) does not align with the citation-grounded differential narrative
the other five models produce; a task-format-aware comparison is in
Supplementary S33.

\subsection{Citation prompting manufactures false rigour}

The most intuitive question is: would better prompts fix this?
We tested this directly by adding an explicit prompt (``You must cite specific PubMed PMIDs for every clinical claim'') to three models (Table~\ref{tab:citation}, citation-prompted rows). 
Claude Opus~4.6 produced 577 unique PMIDs, the highest among all models. Of those, 78.9\% were published in unrelated fields, and only 15.3\% were clinically relevant. Claude Sonnet~4.6 achieved 11.8\%, and DeepSeek-v3 managed 10.4\%. 
These models do not generate random digit strings, but they generate plausible PubMed identifiers that happen to exist in the database. A clinician who spot-checks a cited PMID will find a real paper and may not notice it discusses a different disease entirely.
This is worse than citing nothing: the spot-check confirms a citation that does not actually support the claim.
The \provgap{} is an intrinsic limitation of how parametric memory works, not a prompt-engineering problem.

\subsection{Structured KG grounding halves the Provenance Gap without sacrificing coverage}

Given that prompting alone fails to produce verifiable citations, we next asked whether structured evidence grounding could close the gap. We evaluated \hegtkg{}'s clinical outputs against two baselines: Vanilla (parametric knowledge only) and Guideline-RAG (overlapping source documents as raw text passages) - in a controlled three-arm design where all arms use the same synthesis model (GPT-4.1) to isolate the contribution of evidence grounding (see Methods).

\begin{table}[ht]
\centering
\caption{\textbf{Automated metrics across three arms and three disease pairs (mean $\pm$ SD).}}
\label{tab:threearmmain}
\small
\begin{tabular}{lrrr}
\toprule
\textbf{Metric} & \textbf{\hegtkg{}} & \textbf{Guideline-RAG} & \textbf{Vanilla} \\
\midrule
Evidence Traceability Score  & $0.372 \pm 0.192$ & 0.0   & 0.0   \\
Clinical Feature Coverage (automated)& $0.717 \pm 0.192$ & $0.688 \pm 0.165$ & $0.743 \pm 0.168$ \\
\quad 95\% Bootstrap CI      & $[0.652,\; 0.779]$ & $[0.633,\; 0.741]$ & $[0.688,\; 0.796]$ \\
Provenance Gap               & $\mathbf{0.366 \pm 0.217}$ & $0.688 \pm 0.165$ & $0.743 \pm 0.168$ \\
Temporal Precision            & $0.233 \pm 0.119$ & --- & --- \\
PMID Coverage                & 100\%             & 0\% & 0\% \\
\bottomrule
\end{tabular}

\smallskip
\noindent\footnotesize{Evidence Traceability Score (ETS) = fraction of claims with verifiable PMID citations. Provenance Gap = Feature Coverage $-$ (ETS $\times$ citation reliability) per scenario, clamped at 0, then averaged; lower is better (0 = every covered claim is reliably traceable). Citation reliability: HEG-TKG = 0.97 (verified PMIDs), Vanilla = 0.80, Guideline-RAG = 0.50 (see Methods). Values are means $\pm$ SD across 36 scenarios (12 per disease pair). Bootstrap 95\% CIs: 10,000 resamples, seed = 42, computed over all 36 scenarios. Per-disease-pair breakdown in Supplementary Table~S28.}
\end{table}

\hegtkg{} achieves comparable clinical feature coverage to both baselines (bootstrap 95\% CIs overlap: \hegtkg{} $[0.652, 0.779]$, Vanilla $[0.688, 0.796]$, Guideline-RAG $[0.633, 0.741]$; Mann-Whitney $p > 0.10$) while halving the Provenance Gap (0.366 vs 0.743).

The ETS of 0.372 requires careful interpretation. It measures the \emph{inline citation rate}: 37\% of output claims carry an explicit \texttt{[PMID:X]} tag in the generated text. But \hegtkg{}'s actual verifiability is 100\%. Every output is generated entirely from a PMID-backed evidence context, on average 181 knowledge graph edges drawn from 199 unique PMIDs per scenario, and the full evidence manifest is stored per query. A clinician can trace \emph{any} claim, cited or uncited, back to the specific KG edges and PMIDs that informed it. The 63\% without inline tags are not ungrounded: they are synthesised from the same auditable evidence, but the model did not attach a citation (e.g., combining two cited facts into a differential reasoning statement). The inline citation rate is a lower bound on verifiability, not a measure of it.
The actual verifiability gap is therefore: \hegtkg{} = 100\% (every claim traceable to PMID-backed evidence), vanilla = 0\% (no evidence trail at any level), Guideline-RAG = 0\% (raw text passages with no structured identifiers). The ETS-based Provenance Gap (0.366) is a conservative, automatable proxy for this fundamental difference.
Per-disease-pair performance varies: \hegtkg{} outperforms vanilla on MG/LEMS ($0.814$ vs $0.747$) but underperforms on DMD/BMD ($0.659$ vs $0.771$), reflecting the higher proportion of treatment-specific scenarios where parametric knowledge is stronger (Supplementary Table~S28). A worked example comparing outputs side by side is provided in Supplementary~S29.

The Guideline-RAG receives \emph{overlapping source material}: GeneReviews, OMIM, and the same PubMed abstracts used to build the knowledge graph, arguably more raw text than a clinician would consult, but delivered without graph structure, PMID citations, or quality tiers. It nonetheless produces zero verifiable citations, identical to vanilla, suggesting that the volume of source text is not the bottleneck; rather, it is the structured citation metadata in the graph that enables traceability. Existence of a citation is necessary but not sufficient. The next subsection
audits whether each cited PMID also entails the claim attached to it.

\subsection{Cited PMIDs entail their claims}

Beyond existence, the claim attached to a PMID has to follow from the source. We assessed this
directly with a blinded natural-language inference audit on a stratified
sample of 200 (claim, PMID) pairs drawn from HEG-TKG outputs (seed 42,
balanced across the three disease pairs and the GOLD/SILVER/BRONZE evidence
tiers, capped at 5 rows per PMID to prevent any single source from saturating
the sample). A separate LLM (GPT-4.1, temperature 0) read each (claim,
abstract) pair and labelled it ENTAILS, NEUTRAL, or CONTRADICTS.

Two of 200 claims were flagged as contradicted by their cited abstract
(1.0\%, 95\% bootstrap CI 0.0–2.5\%). The non-contradiction rate was 99.0\%
(95\% CI 97.5–100.0\%); 53.0\% of claims were directly entailed by the abstract,
the remainder were on-topic but not directly asserted (NEUTRAL). For
comparison, unconstrained LLM medical generation has reported citation
fabrication and unsupported-claim rates of 18–50\%~\cite{ji2023hallucination, huang2023hallucination}. Both
contradictions surfaced in the audit were genuine retrieval errors, not
judge mistakes - one was a CIDP review cited for a GBS-specific claim, the
other a numeric mismatch (a paper reporting 4.2\% MuSK-antibody prevalence
cited for a ~6\% claim). Two PMIDs that clinician C3 had independently
flagged as off-topic during the panel evaluation were flagged NEUTRAL by
the NLI judge in 7 of 7 sample appearances at confidence $\geq 0.95$, providing
methodological cross-validation of both protocols. The full audit, per-row
verdicts, tier breakdown, and reproducibility scripts are in
Supplementary S4.1.

\subsection{Quality-tiered knowledge graphs provide 6,316 PMID-backed edges}

The three knowledge graphs collectively contain 5,481 nodes, 6,316 edges, 987 unique source PMIDs, and 1,280 temporal anchors (Supplementary Table~S30). The temporal anchors are disease-trajectory milestones (e.g., Gowers' sign at P3Y-P5Y, ambulation loss at P9Y-P13Y for DMD), not publication dates. Our automated keyword-matching scorer reports Clinical Feature Coverage of 71.7\% (range 65.9-81.4\% across disease pairs). However, manual inspection reveals this substantially underestimates true coverage: 70\% of features scored as ``missing'' are in fact discussed in the output but fail the rigid substring matcher (e.g., ``reading frame rule 90-95\% accuracy'' is marked absent despite the output stating ``fulfilled in $\sim$90\% of DMD cases''). Of the 80 feature instances flagged as missing, 76\% have corresponding concepts present in the KG evidence manifest. The true KG coverage gap: features genuinely absent from the knowledge graph is approximately 5-10\%, confined primarily to treatment specifics available only in full-text articles.

Across all 36 scenarios, \hegtkg{} retrieved 26,139 knowledge graph edges. Of these, 562 (2.2\%) are GOLD tier (cross-tier confirmed: Tier~2 extraction independently confirmed a Tier~1 curated fact, confidence~=~0.95), 2,271 (8.7\%) are SILVER (multi-model or multi-document consensus, confidence~=~0.85), and 23,306 (89.2\%) are BRONZE (single model, single source, confidence~=~0.70). Temporal edges account for 5,802 (22.2\%) of all retrieved evidence. The quality tier distribution is clinically meaningful: GOLD edges represent guideline-confirmed facts (e.g., ``corticosteroids delay ambulation loss in DMD'' confirmed by both GeneReviews and independent literature) that a clinician can treat as high-confidence anchors. SILVER edges reflect claims corroborated across multiple sources or extraction models, suitable for clinical reasoning with standard verification. BRONZE edges flag single-source claims warranting additional scrutiny before acting - analogous to a case report versus a systematic review. This stratification gives clinicians a built-in confidence signal absent from both vanilla LLMs and text-based RAG, where every claim carries equal (zero) provenance weight. To make this signal interpretable in standard EBM vocabulary, we provide a cross-walk from the GOLD/SILVER/BRONZE tiers to GRADE certainty levels~\cite{guyatt2008grade} in Supplementary~S13.1; the figure-card labels in Figure~\ref{fig:provenance}f use this mapping (GOLD~$\to$~High, SILVER~$\to$~Moderate, BRONZE~$\to$~Low).

\subsection{Temporal anchoring provides the clinical ``when'' that static knowledge graphs miss}

\hegtkg{} provides more than citations. The ``T'' denotes temporal anchoring: structured disease-trajectory milestones that, to our knowledge, no existing biomedical knowledge graph provides\cite{tgb2024}. Across 36 scenarios, \hegtkg{} produced 577 PMID-anchored temporal claims; vanilla GPT-4.1 produced zero (Table~\ref{tab:temporal}).

\begin{table}[ht]
\centering
\caption{\textbf{Temporal claims per scenario type.} Time-anchored claims detected by automated regex extraction of ISO~8601 durations and age ranges from clinical outputs. Vanilla GPT-4.1 produces zero across all scenario types.}
\label{tab:temporal}
\small
\begin{tabular}{lrrr}
\toprule
\textbf{Scenario type} & \textbf{\hegtkg{}} & \textbf{Vanilla} & \textbf{$n$ scenarios} \\
\midrule
Temporal comparative  & 222 & 0 & 6 \\
Temporal trajectory   & 197 & 0 & 6 \\
Differential diagnosis & 128 & 0 & 15 \\
Treatment rationale    & 30  & 0 & 9 \\
\midrule
\textbf{Total}         & \textbf{577} & \textbf{0} & \textbf{36} \\
\bottomrule
\end{tabular}
\end{table}

The table does not capture what this looks like in practice.
Consider a pediatrician who has just received genetic confirmation that a 4-year-old boy has a DMD-causing dystrophin mutation. The immediate clinical question is not ``what is DMD?'' but ``what happens next, and when?'' \hegtkg{} produces a structured disease timeline: ``Gowers' sign at age 3--5 years [PMID:29395989, GOLD]; loss of independent ambulation around age 11 [PMID:35501714, BRONZE]; scoliosis after loss of ambulation [PMID:29395989, GOLD]; structured cardiac surveillance and monitoring [PMID:29395989, GOLD]; progressive cardiomyopathy through adolescence [PMID:40922015, BRONZE]; declining respiratory function from school age [PMID:34606104, BRONZE].'' Each milestone carries a PMID and a quality tier. The pediatrician can plan: cardiac screening at age 10, respiratory function monitoring from age 12, orthopaedic referral at first signs of ambulation loss. Vanilla GPT-4.1 produces ``onset of symptoms (delayed motor milestones, Gowers' sign, calf pseudohypertrophy)'' with no age windows and no citations. It tells the clinician \emph{what} happens but not \emph{when} - and ``when'' is what drives clinical management.

The same pattern holds across disease pairs. For GBS/CIDP, \hegtkg{} provides the 7-14-day nadir window and 2-4-week recovery onset with PMIDs, the timeline a GP needs to distinguish GBS from CIDP. For MG/LEMS, the 2-4-year latency between neurological onset and LEMS-associated small-cell lung cancer makes the screening timeline safety-critical. PrimeKG\cite{chandrasekaran2023primekg} tells you that dystrophin is associated with DMD; \hegtkg{} tells you \emph{when each consequence arrives} and backs it with a PMID.

\subsection{Clinicians confirm the verifiability and actionability advantage}
\label{sec:clinician_results}

Three board-certified neurologists and one medical informatician independently scored blinded clinical outputs on five 1--5 Likert dimensions: Verifiability (D1), Actionability (D2), Temporal Precision (D3), Non-Expert Safety (D4), and Clinical Completeness (D5). Each case presented a single output (Vanilla or \hegtkg{}) without system identity; evaluators saw no paired comparisons. Guideline-RAG was excluded from clinician evaluation to conserve evaluator time; its three-arm comparison is covered by the LLM-judge evaluation. The five dimensions target the specific clinical contributions of \hegtkg{}: D1 measures whether claims can be traced to published sources (the Provenance Gap); D3 measures whether outputs provide specific time windows for disease milestones; D4 measures safety for the GP-to-specialist referral pathway.

\textbf{Panel composition.} The primary panel comprised three neurologists at the University Medical Center G\"{o}ttingen. \textbf{C1} (senior neurologist) scored a reduced 36-case pack (12 cases per disease pair, sealed 1:1 system assignment, 18 Vanilla / 18 \hegtkg{}).
\textbf{C2} (first-year neurology trainee) scored the full 72-case pack (36 per arm). \textbf{C3} is an expert in neuromuscular diseases and had authored and refined the clinical vignettes in the scenario-design phase; to preserve independence at scoring time, her blinded evaluation was restricted to her area of deepest expertise, and she scored the CIDP/GBS reduced pack (11 of 12 cases; Case 7 unscored due to the evaluator's time budget). A medical informatician (\textbf{MI}) served as a non-clinician sensitivity reviewer and scored a further 36 cases but exhibited ceiling effects (SD $= 0$ on four of five dimensions). Where we refer below to ``the primary panel'', we mean the three neurologists.

\begin{figure*}[!htbp]
\centering
\includegraphics[width=\textwidth]{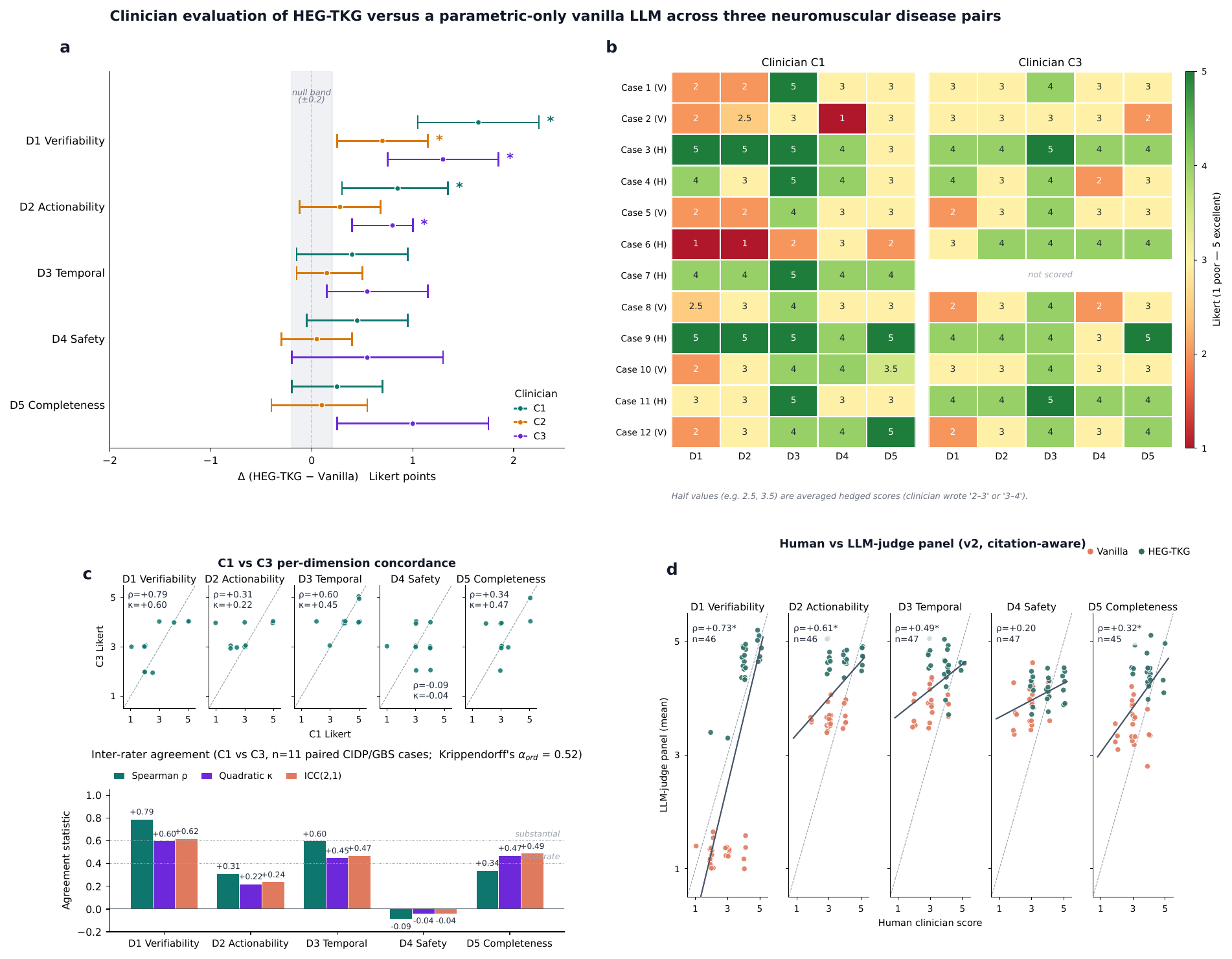}
\caption{\textbf{Clinician evaluation of \hegtkg{} versus a parametric-only vanilla LLM across three neuromuscular disease pairs.} (a) Per-dimension effect sizes $\Delta = \mu_{\hegtkg{}} - \mu_\mathrm{Vanilla}$ in Likert points with 10,000-resample bootstrap 95\% CIs, across the three primary clinicians (C1: all three pairs, $n=17$--$18$ per arm; C2: all three pairs, $n=35$--$36$ per arm; C3: CIDP/GBS only, $n=5$--$6$ per arm). $* = $ Benjamini-Hochberg $q < 0.05$. All three neurologists show positive $\Delta$ on every dimension; D1 Verifiability is BH-significant for all three. The shaded $\pm 0.2$ band indicates the practical-null region. (b) Per-case Likert scores for the 12 CIDP/GBS reduced-pack cases, scored independently by C1 and C3 on the same blinded outputs; V = Vanilla, H = \hegtkg{}; half-step values (2.5, 3.5) are averaged hedged scores (e.g., a clinician wrote ``3--4''). (c) Top: C1 vs C3 scatter per dimension on the 11 overlapping CIDP/GBS cases; dashed line = identity. Bottom: agreement statistics per dimension (Spearman $\rho$, quadratic-weighted Cohen's $\kappa$, ICC(2,1)) with conventional interpretability thresholds (0.4 moderate, 0.6 substantial). Krippendorff's $\alpha$ (ordinal, pooled across D1--D5) = 0.52. (d) Human clinician panel versus three-model LLM-judge panel (v2, citation-aware; GPT-4o-mini + DeepSeek-V3 + Claude~3~Haiku) on the 46--47 matched (C1+C3) cases for which judge scores exist; humans and judges agree most strongly on D1 ($\rho = 0.73$) and D2 ($\rho = 0.61$), with weaker agreement on D3--D5 and none on D4. Dashed line = identity. Panels (a), (d) show effects pooled across disease pairs; panels (b), (c) show the CIDP/GBS overlap design that supports inter-rater analysis. All numbers are reproducible from \texttt{docs/paper1/analysis/clinician\_stats.json}.}
\label{fig:clinician_composite}
\end{figure*}

\textbf{Verifiability advantage replicates across the primary panel (Figure~\ref{fig:clinician_composite}a).} On D1 Verifiability, every neurologist rated \hegtkg{} higher than Vanilla: C1 $\Delta = +1.65$ ($d = 1.79$, $q < 0.001$), C2 $\Delta = +0.67$ ($d = 0.72$, $q = 0.006$), C3 $\Delta = +1.30$ ($d = 2.57$, $q = 0.03$). The direction of the D1 effect holds across all three primary clinicians, across both pack formats (12- and 24-case), and across all five dimensions, consistent with a structural rather than rater-specific signal. D2 Actionability also reaches BH-significance for C1 ($\Delta = +0.85$, $q = 0.012$) and C3 ($\Delta = +0.80$, $q = 0.030$); D3--D5 show positive but generally non-significant deltas, consistent with the LLM-judge finding that \hegtkg{} does not degrade clinical quality on any dimension. Table~\ref{tab:clinician_panel} lists the full per-clinician, per-dimension values.

\begin{table}[htbp]
\centering
\caption{\textbf{Per-arm clinician scores across the primary panel.} Mean $\pm$ SD ($n$ per arm). $\Delta = \mu_{\hegtkg{}} - \mu_\mathrm{Vanilla}$. $p$ from Mann-Whitney $U$, two-sided. BH-corrected across the five dimensions within each clinician. Bootstrap 95\% CIs from 10{,}000 resamples (seed 42).}
\label{tab:clinician_panel}
\small
\begin{tabular}{llccccc}
\toprule
\textbf{Clinician} & \textbf{Dim} & \textbf{Vanilla} & \textbf{\hegtkg{}} &
$\boldsymbol{\Delta}$ \textbf{[95\% CI]} & $\boldsymbol{d}$ & $\boldsymbol{q}$ \\
\midrule
C1  & D1 & $2.35\pm0.70$ ($n{=}17$) & $4.00\pm1.08$ ($n{=}18$) & $+1.65$ $[+1.03,+2.21]$ & $1.79$ & $<0.001$ \\
            & D2 & $2.76\pm0.66$ ($n{=}17$) & $3.61\pm1.04$ ($n{=}18$) & $+0.85$ $[+0.28,+1.36]$ & $0.97$ & $0.012$  \\
            & D3 & $3.78\pm0.65$ ($n{=}18$) & $4.17\pm1.04$ ($n{=}18$) & $+0.39$ $[-0.17,+0.94]$ & $0.45$ & $0.17$   \\
            & D4 & $3.17\pm0.79$ ($n{=}18$) & $3.61\pm0.78$ ($n{=}18$) & $+0.44$ $[-0.06,+0.94]$ & $0.57$ & $0.12$   \\
            & D5 & $3.24\pm0.56$ ($n{=}17$) & $3.47\pm0.72$ ($n{=}17$) & $+0.24$ $[-0.18,+0.65]$ & $0.37$ & $0.18$   \\
\midrule
C2 & D1 & $3.03\pm0.88$ ($n{=}36$) & $3.69\pm0.98$ ($n{=}36$) & $+0.67$ $[+0.22,+1.08]$ & $0.72$ & $0.006$ \\
             & D2 & $3.31\pm0.98$ ($n{=}36$) & $3.57\pm0.74$ ($n{=}35$) & $+0.27$ $[-0.13,+0.66]$ & $0.31$ & $0.57$  \\
             & D3 & $3.64\pm0.76$ ($n{=}36$) & $3.81\pm0.58$ ($n{=}36$) & $+0.17$ $[-0.14,+0.47]$ & $0.25$ & $0.57$  \\
             & D4 & $2.75\pm0.73$ ($n{=}36$) & $2.78\pm0.72$ ($n{=}36$) & $+0.03$ $[-0.31,+0.36]$ & $0.04$ & $0.74$  \\
             & D5 & $3.53\pm1.00$ ($n{=}36$) & $3.58\pm1.02$ ($n{=}36$) & $+0.06$ $[-0.42,+0.50]$ & $0.05$ & $0.74$  \\
\midrule
C3 & D1 & $2.50\pm0.55$ ($n{=}6$) & $3.80\pm0.45$ ($n{=}5$) & $+1.30$ $[+0.73,+1.83]$ & $2.57$ & $0.030$ \\
               & D2 & $3.00\pm0.00$ ($n{=}6$) & $3.80\pm0.45$ ($n{=}5$) & $+0.80$ $[+0.40,+1.00]$ & $2.68$ & $0.030$ \\
               & D3 & $3.83\pm0.41$ ($n{=}6$) & $4.40\pm0.55$ ($n{=}5$) & $+0.57$ $[+0.17,+1.13]$ & $1.19$ & $0.13$  \\
               & D4 & $2.83\pm0.41$ ($n{=}6$) & $3.40\pm0.89$ ($n{=}5$) & $+0.57$ $[-0.23,+1.33]$ & $0.85$ & $0.19$  \\
               & D5 & $3.00\pm0.63$ ($n{=}6$) & $4.00\pm0.71$ ($n{=}5$) & $+1.00$ $[+0.27,+1.73]$ & $1.50$ & $0.082$ \\
\bottomrule
\end{tabular}
\end{table}

\textbf{Inter-rater agreement between two senior neurologists (Figure~\ref{fig:clinician_composite}b,c).} C1 and C3 independently scored the same 11 CIDP/GBS outputs without access to each other's scores or to the sealed system key. They are the only pair for which identical outputs were scored under identical evaluator context (both used the reduced pack, which presents one sealed output per scenario); C2, who used the full pack with both outputs per scenario, is therefore not included in this inter-rater analysis (Methods). Agreement on the flagship D1 Verifiability dimension was strong (Spearman $\rho = 0.79$, $p = 0.006$; quadratic Cohen's $\kappa = 0.60$ $[+0.29, +0.79]$; ICC(2,1) $= 0.62$), with 90\% of scores falling within one Likert point. Agreement was moderate on D3 Temporal ($\rho = 0.60$, $\kappa_q = 0.45$) and D5 Completeness ($\rho = 0.34$, $\kappa_q = 0.47$), and near-zero on D4 Non-Expert Safety ($\rho = -0.09$): safety for a generalist is a dimension on which senior clinicians disagree structurally, reflecting different mental models of what a non-specialist can safely execute without oversight. Krippendorff's $\alpha$ (ordinal, pooled across D1--D5) $= 0.52$, consistent with moderate ordinal agreement. The Verifiability dimension shows the strongest rater agreement, supporting the core claim that the Provenance Gap is a structural property of the output rather than an artefact of a single evaluator's scoring philosophy.


The trainee on our panel (C2) returned a
smaller D1 advantage for HEG-TKG ($\Delta = +0.67$) than either senior neurologist.
The qualitative annotations explain why. C2 marked HEG-TKG down for citing
consensus knowledge - MG antibody subtypes, standard-of-care treatments -
that an experienced specialist already knows by heart and does not need to
verify against a PubMed identifier. That is the right read for a specialist.
It is also a positive finding about deployment context: a verifiable citation
trail does its work where consensus knowledge is not yet consensus to the
reader, which is exactly the situation a GP faces when a rare neuromuscular
patient walks in. Even on this more conservative rubric, C2's D1 effect
survives BH correction.

\textbf{Human--LLM-judge correlation (Figure~\ref{fig:clinician_composite}d).} On the 46--47 clinician-scored cases with matching LLM-judge v2 (citation-aware) scores, the human panel and the three-model LLM-judge panel (GPT-4o-mini, DeepSeek-V3, Claude-3-Haiku) agreed most strongly on D1 Verifiability ($\rho = 0.73$, $q < 0.001$) and D2 Actionability ($\rho = 0.61$, $q < 0.001$), with weaker agreement on D3 Temporal ($\rho = 0.49$, $q < 0.001$) and D5 Completeness ($\rho = 0.32$, $q = 0.04$), and no meaningful agreement on D4 Safety ($\rho = 0.20$, ns). Judge means are systematically higher than human means on D2--D5 (e.g., judge D5 $= 4.44$ vs human $3.38$) but aligned on D1 (judge $3.09$ vs human $3.17$): the citation-aware judge sees the same verifiability gap the clinicians see, but is otherwise more lenient on clinical quality. This supports using LLM judges as scalable screeners for D1 Verifiability while retaining human evaluators for D4 Safety and D5 Completeness. The 72 clinician cases are drawn from the same base scenarios as the LLM-judge evaluation, enabling this shared-case, shared-rubric correlation analysis (Section~\ref{sec:llm_human_correlation}).

\subsection{LLM judges cannot assess verifiability without ground truth}


Quite often, automated judging is the standard fallback when clinician time runs out, so
the question is whether LLM judges can score citation quality on their own.
We ran a three-model judge panel (GPT-4o-mini, DeepSeek-V3, Claude 3 Haiku)
across all 108 cases (36 scenarios $\times$ 3 arms) on the same five Likert
dimensions used by the clinicians - Verifiability~(D1), Actionability~(D2),
Temporal Precision~(D3), Non-Expert Safety~(D4), and Clinical Completeness
(D5). Cases were blinded and randomly ordered.

In v1 (blind), D1 Verifiability showed no meaningful separation: \hegtkg{} $4.59$, Vanilla $4.34$, Guideline-RAG $4.58$ (all pairwise deltas $< 0.5$, non-significant after BH correction). Inter-rater reliability for D1 was negative (Krippendorff's $\alpha = -0.09$), confirming judges fundamentally disagree about what ``verifiable'' means without ground-truth citation data. On D2-D5, all three arms scored comparably: \hegtkg{} showed slight advantages on Actionability ($+0.03$ to $+0.17$), Temporal Precision ($-0.08$ to $+0.28$), and Completeness ($+0.11$ to $+0.31$) vs Vanilla, with no degradation on any dimension. Guideline-RAG scored lower on Temporal Precision (D3: $3.47$-$4.28$) than both other arms, reflecting the absence of structured temporal data. The D1 failure was specific and structural: judges score citation \emph{plausibility} (does the text \emph{look} well-referenced?) rather than citation \emph{validity} (are the references real?).
To confirm this, we deployed a second evaluation round (v2) with one addition: each judge received a PubMed E-utilities citation audit report documenting which PMIDs existed, their publication titles, and clinical relevance to the disease pair. This three-arm design ($N = 108$ cases, 36 per arm, 3 judges each) tests whether the citation blindness extends to Guideline-RAG, which has access to overlapping source texts but without structured citations.

\begin{table}[ht]
\centering
\caption{\textbf{v2 citation-aware D1 Verifiability across all three disease pairs (3 arms).}}
\label{tab:v2d1}
\small
\begin{tabular}{lcccccc}
\toprule
\textbf{Pair} & \textbf{Vanilla D1} & \textbf{G-RAG D1} & \textbf{\hegtkg{} D1} & \textbf{$\Delta_\mathrm{H-V}$} & \textbf{$d_\mathrm{H-V}$} & \textbf{$p$ (BH)} \\
\midrule
MG/LEMS  & $1.39 \pm 0.49$ & $1.39 \pm 0.49$ & $4.44 \pm 1.05$ & $+3.06$ & $3.71$ & $<10^{-11}$ \\
DMD/BMD  & $1.39 \pm 0.49$ & $1.50 \pm 0.51$ & $4.92 \pm 0.28$ & $+3.53$ & $8.78$ & $<10^{-13}$ \\
CIDP/GBS & $1.36 \pm 0.49$ & $1.39 \pm 0.49$ & $4.58 \pm 0.77$ & $+3.22$ & $5.00$ & $<10^{-12}$ \\
\bottomrule
\end{tabular}

\smallskip
\noindent\footnotesize{Mann-Whitney $U$, BH-corrected across 6 comparisons. $N = 36$ per arm per pair (12 scenarios $\times$ 3 judges). All H-vs-G comparisons similarly significant ($p < 10^{-11}$, $d > 3.7$).}
\end{table}

All three disease pairs told the same story. In v2, both Vanilla and Guideline-RAG D1 scores collapsed to $1.36$--$1.50$ out of~5.0. Judges could now see that neither arm provided verifiable citations. \hegtkg{} held at $4.44$-$4.92$. Effect sizes ranged from Cohen's $d = 3.71$ to $8.78$, the largest in this study. 
Guideline-RAG scored identically to Vanilla on D1 despite having access to the same source literature as \hegtkg{}, confirming that raw text access without structured citation mechanisms provides no verifiability advantage. Scores on D2-D5 remained comparable (D5 Clinical Completeness: \hegtkg{} $4.73$, Vanilla $4.08$, Guideline-RAG $3.82$), confirming that verifiability does not cost clinical quality.

\begin{table}[ht]
\centering
\caption{\textbf{v1 (blind) vs v2 (citation-aware) D1 Verifiability shift.}}
\label{tab:v1v2}
\small
\begin{tabular}{lcccc}
\toprule
\textbf{System} & \textbf{v1 D1} & \textbf{v2 D1} & \textbf{$\Delta$} & \textbf{$\alpha$: v1 $\to$ v2} \\
\midrule
\hegtkg{}      & $4.59$ & $4.65$ & $+0.06$ & \multirow{3}{*}{$-0.09 \to 0.89$} \\
Vanilla        & $4.34$ & $1.38$ & $-2.96$ & \\
Guideline-RAG  & $4.58$ & $1.43$ & $-3.16$ & \\
\bottomrule
\end{tabular}
\end{table}

Inter-rater reliability for D1 shifts from no agreement in v1 (Krippendorff's $\alpha = -0.09$) to excellent in v2 ($\alpha = 0.89$). This finding has a direct methodological implication: LLM-as-judge evaluation of citation quality without external verification data measures surface plausibility, not actual verifiability.

\begin{figure}[!htbp]
\centering
\includegraphics[width=\textwidth]{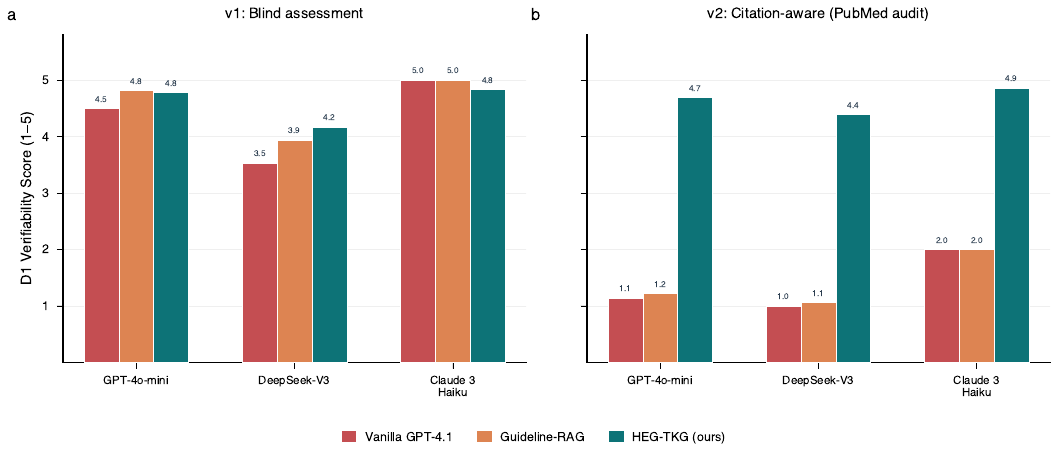}
\caption{\textbf{LLM judges cannot assess verifiability without citation ground truth.} Per-judge D1 Verifiability scores pooled across disease pairs. (a)~v1 blind: all judges score all three arms similarly (arm means: $4.34$--$4.59$). (b)~v2 citation-aware: Vanilla and Guideline-RAG collapse (arm means: $1.38$--$1.43$) while \hegtkg{} remains stable (arm mean: $4.65$). Individual judge variation visible in bars; DeepSeek-V3 is strictest, Claude~3 Haiku most lenient. $N = 108$ cases (36 per arm), 3 judges.}
\label{fig:judges}
\end{figure}

\subsection{LLM judges align with clinicians on temporal precision but not completeness}

\label{sec:llm_human_correlation}

We computed Spearman~$\rho$ between human panel means (C1 + C2, $n = 72$ shared cases) and LLM panel means (three-judge mean per case) on each dimension, separately for v1 (blind) and v2 (citation-aware).

\begin{table}[ht]
\centering
\caption{\textbf{Spearman correlation between clinician and LLM judge panel means.} $n = 72$ shared (scenario, arm) pairs for D1, D3--D5; $n = 71$ for D2.}
\label{tab:spearman}
\small
\begin{tabular}{lcccc}
\toprule
\textbf{Dimension} & \multicolumn{2}{c}{\textbf{v1 (blind)}} & \multicolumn{2}{c}{\textbf{v2 (citation-aware)}} \\
\cmidrule(lr){2-3} \cmidrule(lr){4-5}
& $\rho$ & $p$ & $\rho$ & $p$ \\
\midrule
D1 Verifiability      & $0.299$ & $0.011$ & $0.250$ & $0.034$ \\
D2 Actionability      & $0.122$ & $0.309$ & $0.243$ & $0.042$ \\
D3 Temporal Precision & $0.338$ & $0.004$ & $0.391$ & $< 0.001$ \\
D4 Non-Expert Safety  & $0.143$ & $0.231$ & $0.267$ & $0.023$ \\
D5 Completeness       & $0.116$ & $0.332$ & $0.121$ & $0.312$ \\
\bottomrule
\end{tabular}
\end{table}

Two patterns stand out. First, D3 (Temporal Precision) is the dimension with the strongest and most consistent human--LLM agreement ($\rho = 0.338$ blind, $0.391$ citation-aware, both $p < 0.005$). LLM judges can detect temporal specificity, whether an output contains concrete age windows and milestone sequences, without clinical expertise, because temporal anchors are surface-level textual features. 
Second, citation-aware judging (v2) improves correlation on D2 and D4 ($\rho$ increases from $0.122 \to 0.243$ and $0.143 \to 0.267$, respectively), suggesting that citation audit data helps LLM judges better align with clinical assessments of actionability and safety. D5 (Completeness) remains uncorrelated in both conditions ($\rho \approx 0.12$), indicating that what clinicians consider ``complete'' differs structurally from what LLM judges infer from text.

The modest overall correlations ($\rho = 0.12$--$0.39$) reinforce the finding from the v1/v2 comparison: LLM judges are not substitutes for clinician evaluation. They capture surface features (temporal specificity, citation presence) but miss the clinical judgement that drives D4 (safety for non-experts) and D5 (completeness relative to a specialist's mental model). This supports the use of LLM judges for scalable screening but not for final validation of clinical AI systems.

\subsection{Injected clinical errors are resisted or caught via citation trace}

We injected 15 clinically incorrect evidence statements (5 per disease pair) into the \hegtkg{} evidence pipeline, for example, swapping AChR antibody targets between MG and LEMS, reversing the age of ambulation loss for DMD/BMD, or assigning incorrect first-line treatments.

\begin{table}[ht]
\centering
\caption{\textbf{Counterfactual experiment results.}}
\label{tab:counterfactual}
\small
\begin{tabular}{lrc}
\toprule
\textbf{Outcome} & \textbf{Count} & \textbf{Percentage} \\
\midrule
Parametric resistance (correct despite wrong evidence) & 12/15 & 80\% \\
Partial incorporation (hedged with correct knowledge) & 1/15 & 7\% \\
Faithful to wrong evidence & 2/15 & 13\% \\
Detectable via citation traceability & 15/15 & 100\% \\
\bottomrule
\end{tabular}
\end{table}

The 13\% failure rate (2/15; Wilson 95\% CI\cite{wilson1927ci} $[0.037, 0.379]$) is clinically informative. CF\_DMD\_04 injected ``corticosteroids are contraindicated in DMD and accelerate muscle degeneration'' - the model parroted this claim, the most dangerous failure because corticosteroids are standard-of-care in DMD. CF\_CIDP\_02 injected ``corticosteroids are first-line for GBS; IVIg is ineffective'' - the model inverted the actual evidence hierarchy. Both cases involved treatment recommendations, where models defer to retrieved evidence over parametric knowledge. One additional case (CF\_CIDP\_05) was ambiguous: the model produced a contradictory hybrid response about nerve conduction study patterns, partially incorporating the counterfactual while hedging with correct parametric knowledge. The parametric resistance rate of 80\% (Wilson 95\% CI $[0.548, 0.930]$) reflects both a strength and a limitation: the model's prior training provides a safety net for most injected errors, but that net fails specifically for treatment recommendations. What makes the system safe despite this is that every claim, right or wrong, remains traceable to its source PMID. A clinician can identify exactly which citation backs a suspicious treatment recommendation and catch the error. Ungrounded LLMs and text-based RAG offer neither parametric resistance nor citation traceability.

\begin{figure}[!htbp]
\centering
\includegraphics[width=\textwidth]{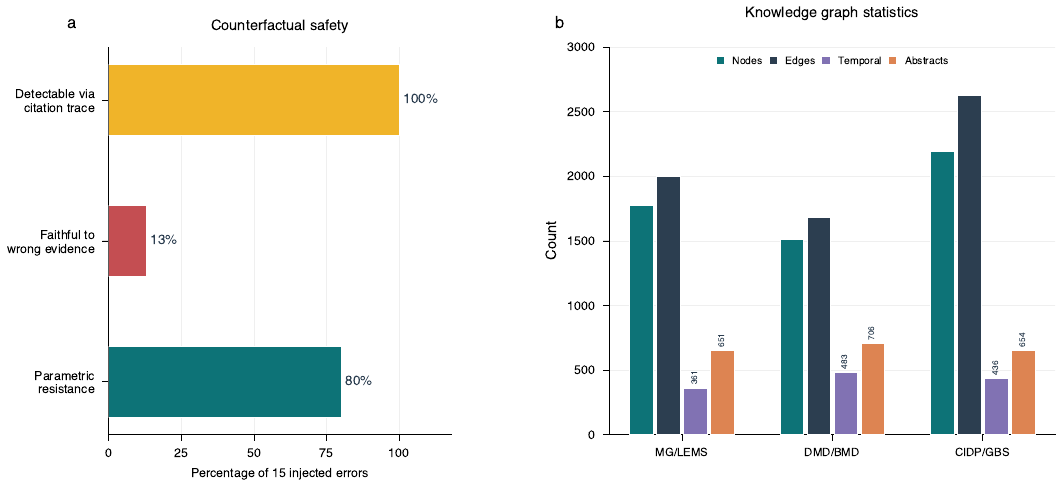}
\caption{\textbf{Counterfactual safety and knowledge graph statistics.} (a)~Of 15 injected clinically incorrect evidence statements, 80\% were resisted by the model's parametric knowledge, and 100\% remained detectable via citation trace. (b)~The three \hegtkg{} knowledge graphs collectively contain 5,481 nodes, 6,316 edges, and 1,280 temporal anchors across the three disease pairs.}
\label{fig:counterfactual_kgstats}
\end{figure}

\subsection{Local synthesis keeps patient data on site}

Patient data never enters the knowledge graph. The graph stores only
disease-level relationships drawn from public PubMed abstracts and
authoritative sources, and only the synthesis step at inference time
sees a clinical scenario. We validated on-premise synthesis with
open-source models (Qwen~3 14B\cite{qwen2024}, Gemma~3 27B\cite{gemma2024}) served via
Ollama\cite{ollama2024}, reaching 75–80\% citation coverage on a single institutional
GPU. A deployment-time check refuses to start the system if any
component is configured to route patient data to a cloud API
(Supplementary S31; regulatory mapping in Supplementary Tables~S25–S26).

\FloatBarrier

\section{Discussion}


Bigger models and better prompts do not close the Provenance Gap. It is a
structural failure of how parametric memory works: a model can recall a
plausible PMID without being able to prove the claim it attaches to. To make
this concrete for clinicians and reviewers, we group citation behaviours into
four levels (Table~\ref{tab:hierarchy}). Vanilla LLMs sit at Level 0 with no trail to follow.
Text-based RAG, including the Guideline-RAG baseline tested here, sits at
Level 1: source attribution without structured identifiers, evidence grading,
or claim-level mapping, and subject to roughly 25\% link rot within two years.
Agentic architectures with PubMed tool access reach Level 2; DeepRare reports
~95\% physician-rated reference accuracy on this design, but reactive retrieval
returns a different evidence set per query, so audit trails do not reproduce
across sessions, and Zhao et al.~\cite{zhao2026deeprare} themselves flag hallucinated URLs and
irrelevant citations as failure modes (a task-format-aware comparison is in
Supplementary S33). HEG-TKG operates at Level 3: every claim maps to a
verified PMID, carries a quality tier reflecting cross-source consensus, and
sits within structured temporal context.

\begin{table}[ht]
\centering
\caption{\textbf{Four-level citation quality hierarchy.} The Provenance Gap measures a system's distance from Level~3. All tested frontier LLMs remain below Level~2.}
\label{tab:hierarchy}
\small
\begin{tabular}{llp{8cm}}
\toprule
\textbf{Level} & \textbf{Type} & \textbf{Description} \\
\midrule
0 & No citation & Vanilla LLMs. Claims with no trail to follow. \\
1 & Source attribution & URLs or text passages indicating where the system read a claim. Subject to link rot (25\% within two years\cite{klein2014scholarly,zittrain2014perma}). No structured identifiers, evidence grading, or claim-level mapping. \\
2 & Identifier-based & Agentic systems retrieve live evidence at inference time (e.g., DeepRare\cite{zhao2026deeprare}): high physician-rated accuracy ($\sim$95\%) but reactive retrieval yields non-reproducible citation trails across sessions, with hallucinated URLs as a documented failure mode\cite{zhao2026deeprare}. No quality stratification or temporal anchoring. \\
3 & Evidence provenance & Every claim maps to a verified PMID, carries a quality tier (GOLD/SILVER/BRONZE) reflecting cross-source consensus, and sits within a structured temporal context. \\
\bottomrule
\end{tabular}
\end{table}

Why structure rather than text alone? Guideline-RAG receives overlapping
source material GeneReviews, OMIM, and the same PubMed abstracts that
populate the knowledge graph, and produces zero verifiable citations, the
same as vanilla. Source volume is not the bottleneck. But graph structure on
its own is not enough either: MedGraphRAG \cite{wu2024medgraphrag} and FPKG \cite{song2025fpkg} build medical knowledge
graphs without programmatic verification or cross-source quality
stratification. The combination is what closes the gap. The counterfactual
experiment makes the safety logic explicit: 80\% of injected errors were
resisted by parametric knowledge, 13\% were faithfully incorporated, and all
15, including the two failures, remained traceable to their cited source. The
audit trail, not error prevention alone, is what carries the safety claim.

A vanilla LLM output of the kind we tested typically carries 28 clinical
claims. Verifying each by hand against PubMed runs to about 70 minutes - out
of reach in a 15-minute consultation. With HEG-TKG, 37\% of claims carry
inline PMIDs that resolve in seconds, and the rest trace through the stored
evidence manifest, which cuts verification time by roughly 50–64\%. That budget
is what makes three workflows tractable: a GP can attach a PMID-backed
output to a specialist referral letter; a neurologist can check age-anchored
milestones during a consultation without leaving the room; a clinician who
disagrees with a recommendation can pull up the exact PMID behind it and
argue with the source rather than with a black box.

The deployment population matters here. The early-career trainee on our panel
(C2) penalised HEG-TKG for citing consensus knowledge - antibody subtypes,
standard-of-care treatments - that an experienced specialist already carries.
That is the right read for a specialist. It is also the read that explains why
verifiability earns its keep with non-specialist users facing disease territory
they do not work in often, which is exactly where most rare neuromuscular
patients first present. The barrier to clinical AI adoption is not accuracy\cite{nori2023gpt4,singhal2023medpalm2}.
It is that clinicians will not act on a recommendation they cannot verify.

The cleanest evidence that the gap is real comes from comparing what LLM
judges see when they cannot check the citations against what they see when
they can \cite{chiang2023llmjudge,zheng2023llmjudge,krippendorff2019}. In the blind round, the three judges could not agree on what
"verifiable" meant - Krippendorff's $\alpha$ came out at $-0.09$, worse than chance.
We then handed them a PubMed audit report and re-ran the same evaluation.
Agreement jumped to $\alpha = 0.89$ and the effect size for HEG-TKG versus vanilla
reached Cohen's $d$ = $8.78$. The judges had not changed. The text they were
scoring had not changed. Only the audit was new. So a citation-quality
benchmark that runs without external verification is measuring how
referenced the text looks, not whether the references hold.

The same logic now reaches further upstream. A 2026 Nature News investigation
identified 124 published disease-prediction models trained on health
datasets of unverified provenance\cite{nature2026dubiousdata}. When training data cannot be audited,
output-level verifiability is the one artifact a clinician can still check
before acting. The clinician panel (Figure~\ref{fig:clinician_composite}, Table~\ref{tab:clinician_panel}) converges with the judge result from the
opposite direction: every neurologist on the primary panel scored HEG-TKG
higher than vanilla on D1, and the two senior raters who independently
scored the same 11 CIDP/GBS outputs agree at $\rho = 0.79$ on D1 with
quadratic-weighted $\kappa_q = 0.60$. The verifiability signal holds across raters,
pack formats, and disease pairs. Where the trainee panelist diverged, the
divergence carried clinical information rather than noise - a signal about
which user populations the system serves, not a softer reading of the same
construct.

A few honest constraints bound the result. Our evaluation covers three rare
neuromuscular disease pairs; generalisation to disease areas with substantially
sparser PubMed coverage - ultra-rare metabolic disorders, oncological
subtypes - is not demonstrated and remains future work, and the system
currently lacks graceful degradation for out-of-scope queries. The automated
keyword-matching scorer reports 66–81\% feature coverage across pairs, but
manual inspection puts the true KG gap at roughly 5–10\%, confined mostly to
treatment specifics that live only in full-text articles. We did not run a
head-to-head agentic benchmark; the task-format mismatch with DeepRare~\cite{zhao2026deeprare} is
laid out in Supplementary S33, and a scan of the four publicly released
RareBench~\cite{chen2024rarebench} subsets (n = 1,122 cases) found zero ground-truth-coded cases for
our six target diseases, ruling out a scope-matched RareBench evaluation.
Tier 1 sources were initially curated by a single annotator, which is a
limit we name openly; multi-annotator verification and self-evolving
verifiers along the lines of AutoBioKG~\cite{zheng2026autobiokg} are part of the future work
extension currently under construction.

The clinician panel is small. Three board-certified neurologists at one
institution, two senior and one trainee, with a non-clinician sensitivity
reviewer excluded from the primary analysis for ceiling effects. On the 11
CIDP/GBS cases scored by both senior raters, agreement on the flagship D1
dimension is strong ($\rho = 0.79$, $\kappa = 0.60$, ICC(2,1) $= 0.62$) and
agreement on D4 Non-Expert Safety is near-zero ($\rho = -0.09$). The D4 result
is not a measurement problem. Senior neurologists hold structurally
different mental models of what a GP can safely execute without
specialist oversight, and the rubric surfaces that disagreement faithfully.
Multi-site evaluation with larger panels, harmonised D4 training, and
broader disease-area coverage are the next required steps. Beyond that, a
harmonised benchmark spanning differential ranking and narrative synthesis
would let agentic systems and structurally grounded systems be measured on
common ground.

\section{Methods}

\subsection{Knowledge graph construction}

\textbf{Two-tier architecture.}
\hegtkg{} employs a hierarchical design that separates authoritative clinical facts (Tier~1) from literature-extracted knowledge (Tier~2). Tier~1 relationships were programmatically extracted from GeneReviews\cite{genereviews2024}, OMIM\cite{omim2024}, Orphanet\cite{orphanet2024}, and clinical care guidelines\cite{cdc_dmd2018} by the first author using disease-specific extraction scripts, then tagged as protected edges (\texttt{is\_protected=True}) that are preserved even when literature sources present conflicting data. 
We acknowledge single-author initial curation as a limitation; multi-annotator verification with formal inter-annotator agreement is planned for the expanded TemporalAtlas pipeline.

\textbf{Corpus selection.}
Tier~1 entities inform PubMed queries via the E-utilities API\cite{sayers2022eutils}. For each disease pair, we retrieved abstracts from PubMed using disease-specific MeSH\cite{lipscomb2000mesh} terms and free-text queries, yielding 651 (MG/LEMS), 706 (DMD/BMD), and 654 (CIDP/GBS) abstracts after relevance screening.

\textbf{Multi-LLM consensus extraction.}
Abstracts are processed through a six-step pipeline (Figure~\ref{fig:architecture}; formal pseudocode in Supplementary Algorithm~S27):

\begin{enumerate}[nosep]
  \item \textbf{Relevance screening.} Each abstract is assessed by the first available extraction model (Claude Haiku~4.5 or GPT-4.1-mini) using a structured prompt that checks for disease-specific extractable content against include/exclude criteria defined in the YAML configuration (Supplementary~S15). Abstracts must receive \texttt{extract: true} with confidence $\geq 0.85$ to proceed. This filtered 40-60\% of non-informative abstracts (e.g., 293/651 passed for MG/LEMS).
  \item \textbf{Schema-guided triplet extraction.} Two models (Claude Haiku~4.5, GPT-4.1-mini/Gemini 2.0-flash) independently extract subject-predicate-object triplets against a disease-specific schema of 34 predicates across six categories (Supplementary~S15). Every triplet must include an evidence quote from the source abstract. The architecture supports additional models and scales consensus voting accordingly.
  \item \textbf{Entity normalization.} Entities are mapped to UMLS CUIs via cascading resolution: dictionary lookup, SapBERT entity linking\cite{liu2021sapbert}, and ScispaCy fallback\cite{neumann2019scispacy}, followed by 18 semantic correction rules (Supplementary Table~S10).
  \item \textbf{Temporal anchor resolution.} Time-dependent predicates are anchored to structured ISO~8601 durations at four precision levels: exact age (e.g., P13Y), range (P13Y-P16Y), fuzzy qualifier (``late teens'' $\to$ P17Y-P19Y), and developmental stage (``early childhood'' $\to$ P2Y-P6Y).
  \item \textbf{Multi-model consensus voting.} Triplets extracted by multiple models from the same document are deduplicated via MD5 hashing of normalized subject$|$predicate$|$object. Quality tiers are assigned: GOLD (Tier~1 curated knowledge or cross-tier confirmation where Tier~2 extraction independently confirms a Tier~1 fact; confidence~=~0.95), SILVER ($\geq$2 extraction models agree, or $\geq$2 independent PMIDs support the same triplet; confidence~=~0.85), BRONZE (single model, single source; confidence~=~0.70).
  \item \textbf{Cross-tier integration.} Tier~2 extractions are merged with the Tier~1 curated backbone, with Tier~1 protected edges preserved in case of conflict. The unified graph is serialized to Neo4j via Cypher import scripts.
\end{enumerate}

The pipeline is configuration-driven: each disease pair is defined by a YAML configuration specifying disease entities (with UMLS CUIs), extraction schemas, temporal predicates, and few-shot examples (Supplementary~S15). Adding a new disease pair requires only this configuration; no code changes are needed.

\begin{figure}[!htbp]
\centering
\includegraphics[width=\textwidth]{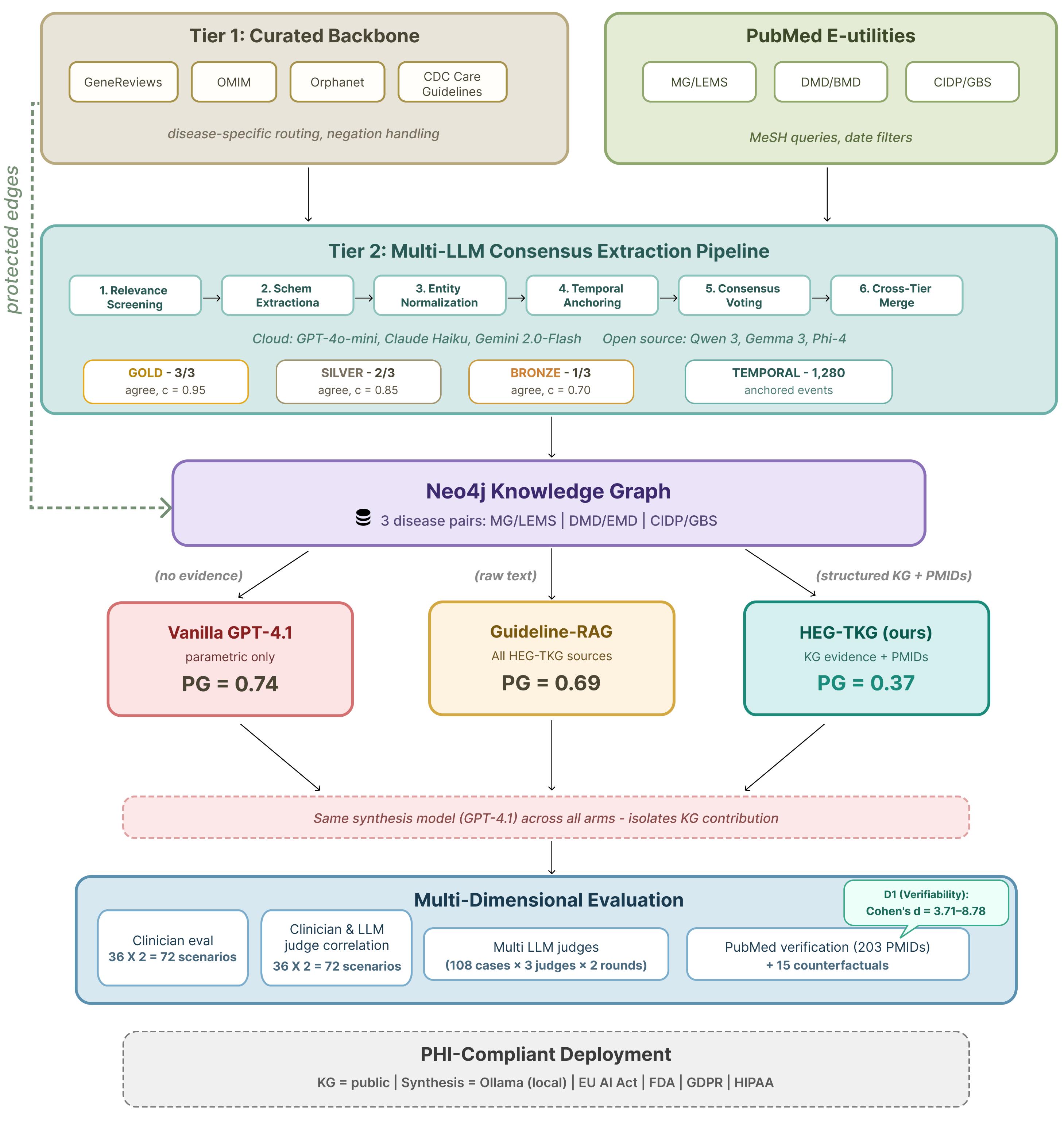}
\caption{\textbf{\hegtkg{} system architecture.} Two-tier knowledge graph construction (curated Tier~1 backbone + multi-LLM Tier~2 extraction), three-arm clinical comparison, and multi-dimensional evaluation including citation-aware LLM judging.}
\label{fig:architecture}
\end{figure}

\begin{figure}[!htbp]
\centering
\includegraphics[width=\textwidth]{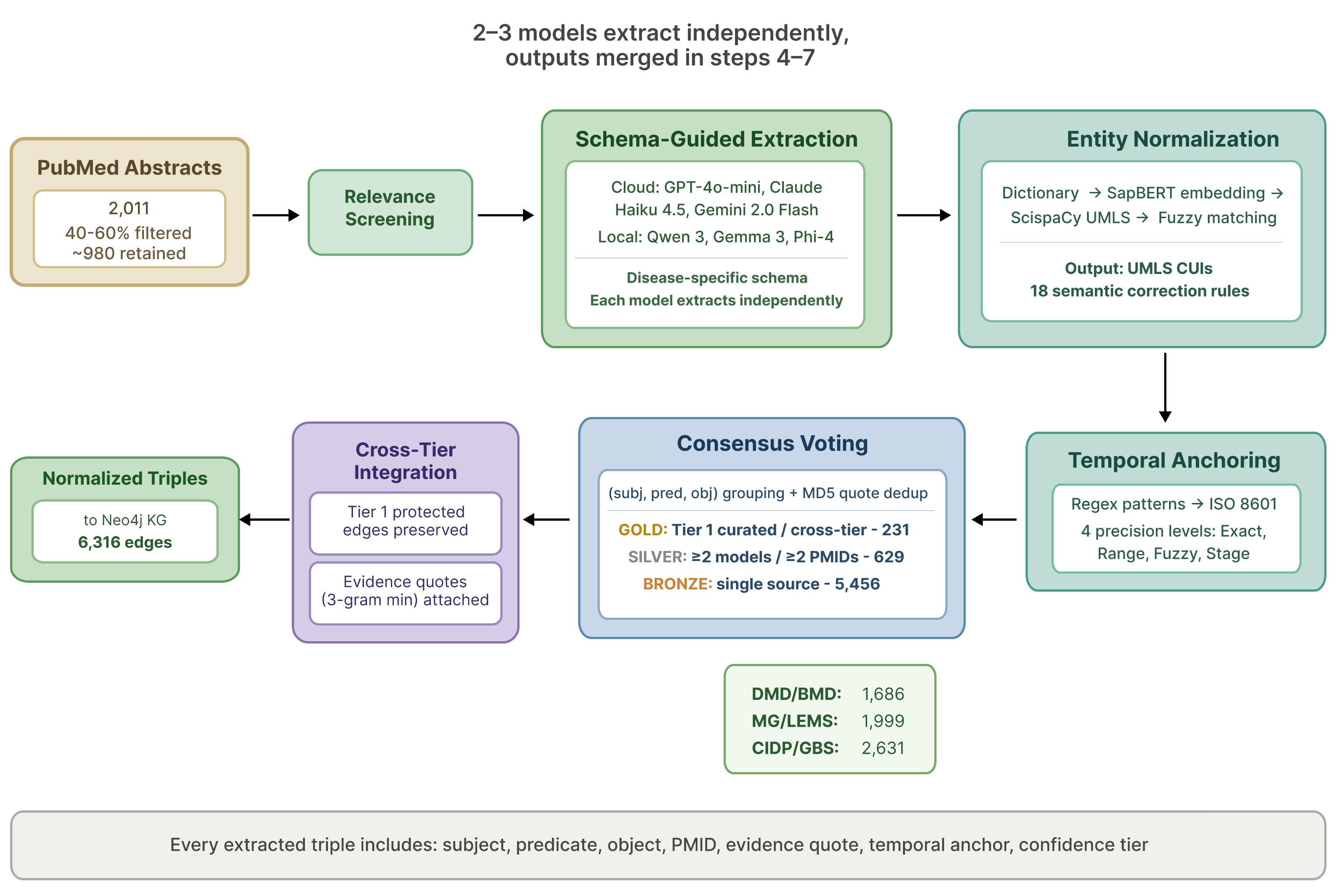}
\caption{\textbf{Tier~2 literature extraction pipeline.} PubMed abstracts are processed through relevance screening, schema-guided extraction (2 models), entity normalization (Dictionary $\to$ SapBERT $\to$ ScispaCy), temporal anchoring, consensus voting, and cross-tier integration, yielding 6,316 normalized edges with PMID provenance across three disease pairs.}
\label{fig:tier2pipeline}
\end{figure}

\begin{figure}[!htbp]
\centering
\includegraphics[width=\textwidth]{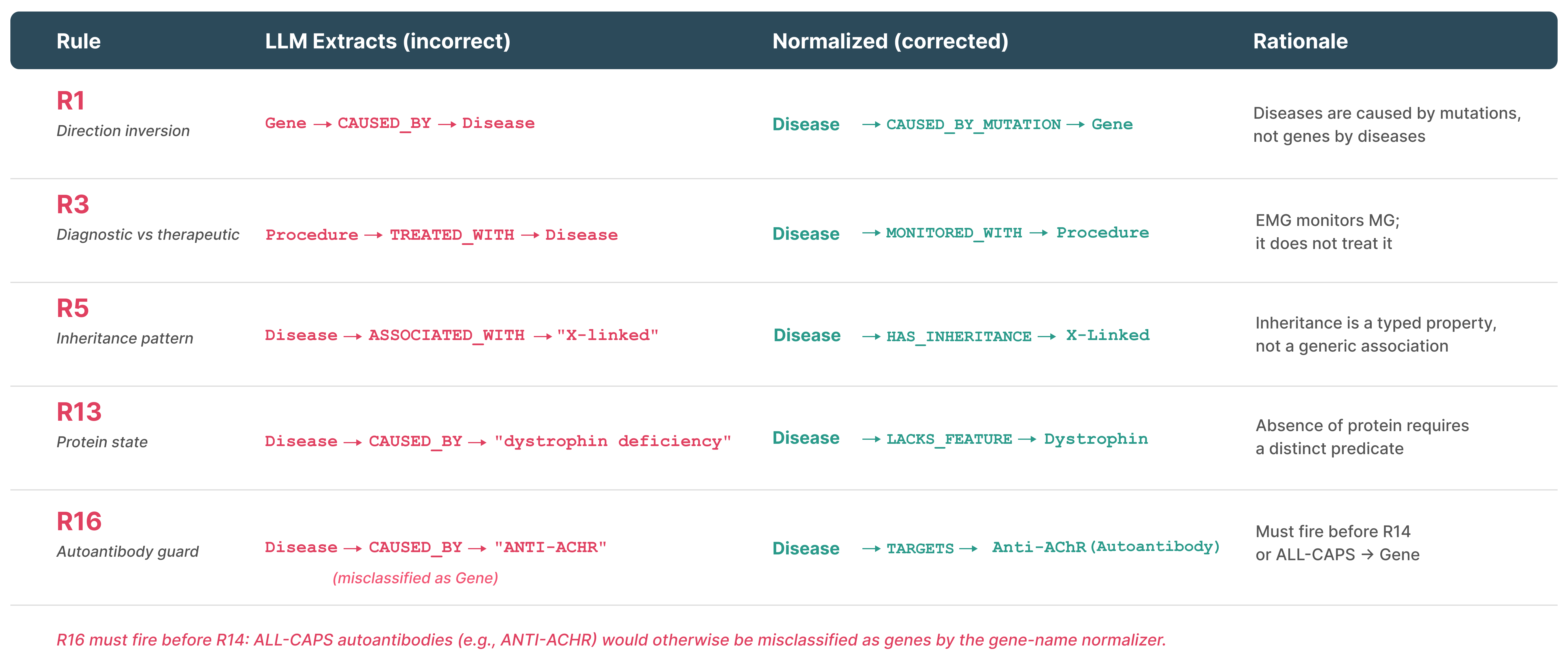}
\caption{\textbf{Representative semantic correction rules.} Five of the 18 rules applied during entity normalization (Step~3). Each rule detects a specific extraction error pattern (left), applies a structured correction (centre), and logs a rationale (right). Rule types include direction inversion (R1), predicate refinement (R3, R5), entity retyping (R13, R16). The full rule set is listed in Supplementary Table~S10.}
\label{fig:semantic_rules}
\end{figure}

\begin{figure}[!htbp]
\centering
\includegraphics[width=\textwidth]{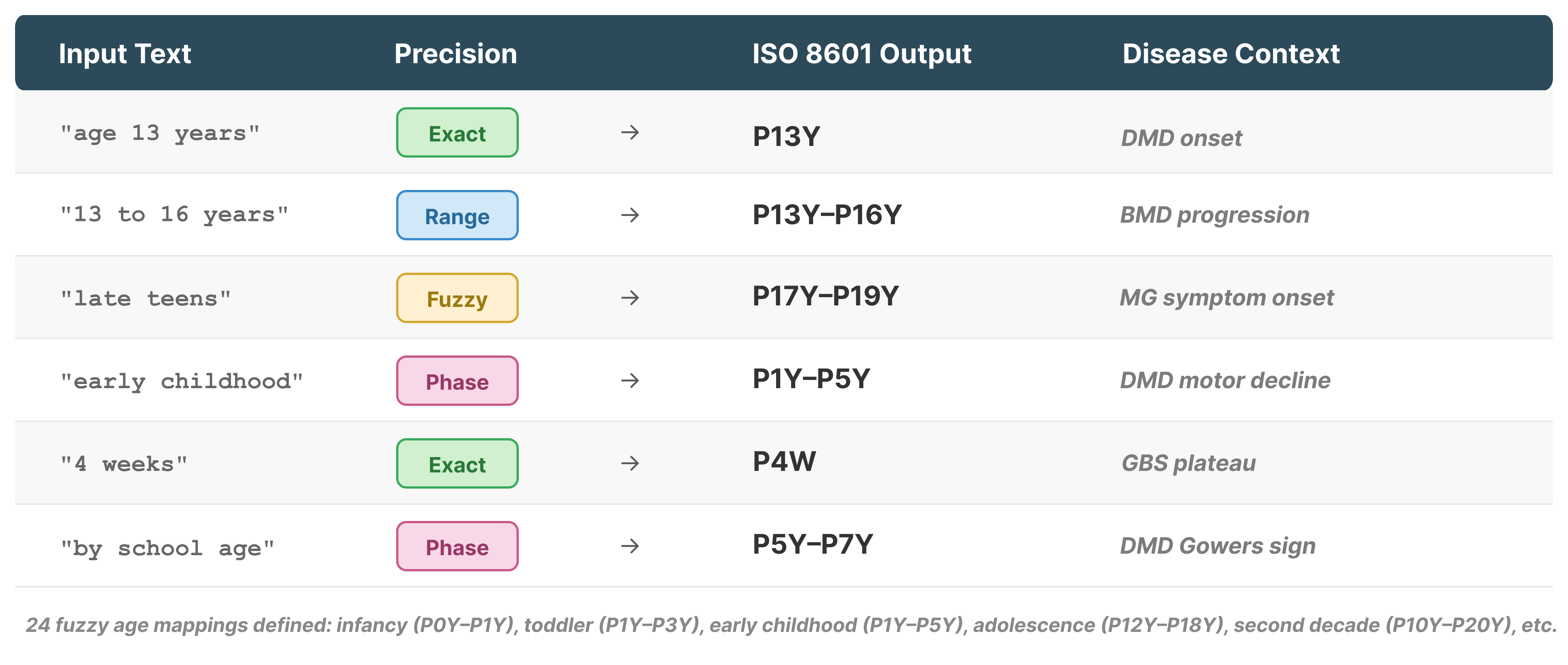}
\caption{\textbf{Temporal anchor resolution examples.} Six representative inputs illustrating the four precision levels used in Step~4. Exact ages map to single ISO~8601 durations (e.g., ``age 13 years'' $\to$ P13Y), ranges capture bounded intervals (``late teens'' $\to$ P17Y-P19Y), fuzzy qualifiers resolve developmental stages (``early childhood'' $\to$ P1Y-P5Y), and phase-level anchors capture disease milestones. Each anchor is linked to its source PMID and disease context.}
\label{fig:temporal_anchoring}
\end{figure}

\FloatBarrier
\subsection{Clinical scenario design}

Thirty-six clinical scenarios (12 per disease pair) were authored as realistic neurological vignettes by a biomedical AI researcher with clinical domain training (Supplementary~S1). Scenarios span four reasoning types: differential diagnosis ($n=15$, including 3 safety/diagnostic-pitfall scenarios), temporal comparative ($n=6$), temporal trajectory ($n=6$), and treatment rationale ($n=9$). A senior neuromuscular specialist (J.Z., University Medical Center G\"{o}ttingen) reviewed and clinically refined 22 of 36 scenario vignettes prior to blinding, correcting factual drift, tightening phenotype descriptions, and calibrating difficulty to realistic GP-to-specialist hand-off situations. J.Z.\ subsequently participated in the blinded evaluation panel as C3, restricted to her area of deepest expertise (CIDP/GBS); she scored without access to the sealed system key and with no prior sight of system identity. The 36-scenario set reflects the practical ceiling for board-certified evaluator time (the full pack asks each clinician to read 72 independent outputs and score each on 5 dimensions). For automated LLM-as-judge evaluation, the same 36 scenarios were evaluated across all 3 arms (108 cases total).

\subsection{Three-arm evaluation design}

All three arms use GPT-4.1 (temperature$=$0.0) as the synthesis model to isolate the contribution of evidence grounding. Prompts are identical except for the evidence context (Supplementary~S2):

\begin{itemize}[nosep]
  \item \textbf{Vanilla}: System prompt + clinical scenario only
  \item \textbf{Guideline-RAG}: System prompt + clinical scenario + retrieved text passages from GeneReviews, OMIM, and the same PubMed abstracts used in Tier~2 extraction (overlapping with \hegtkg{}'s source corpus, but delivered as raw text without graph structure or citation metadata). Source documents were chunked into 500-character segments with 80-character overlap, embedded using all-MiniLM-L6-v2\cite{reimers2019sbert}, and indexed per disease pair (3,160--3,960 chunks each). At query time, the top-$k$ ($k=10$) most similar chunks were retrieved by cosine similarity and concatenated into the prompt context. Chunk size and $k$ were not tuned; we selected standard defaults to provide a reasonable baseline rather than an optimised competitor, since the comparison tests the effect of \emph{structure} (graph vs.\ raw text), not retrieval engineering.
  \item \textbf{\hegtkg{}}: System prompt + clinical scenario + structured evidence from the knowledge graph (entities, relationships, quality tiers, temporal anchors, PMID citations)
\end{itemize}

Graph retrieval uses disease-specific Cypher queries against Neo4j~5\cite{neo4j2024} (Supplementary~S3), returning relevant subgraphs formatted as structured evidence blocks with inline PMID citations and quality tier indicators. End-to-end inference latency for \hegtkg{} comprises Cypher subgraph retrieval ($<$1~s), evidence formatting ($<$1~s), and LLM synthesis. Total wall-clock time is ${\sim}$8~s with local open-source models (Supplementary~S27) and 20--50~s with cloud frontier models, dominated by the synthesis model's generation time. The retrieval overhead is negligible relative to vanilla inference.

\subsection{Multi-model citation verification}

\textbf{Vanilla mode.} Four frontier LLMs (GPT-4.1\cite{openai2023gpt4}, GPT-5.4, Claude Sonnet~4.6\cite{anthropic2024claude}, DeepSeek-v3\cite{deepseekai2024v3}) generated clinical outputs for all 36 scenarios using their default behavior (no citation instruction). Outputs were parsed for PMID patterns (regex: \texttt{PMID:\textbackslash s*\textbackslash d+}), author--year references, and other citation formats.

\textbf{Citation-prompted mode.} Three models (Claude Opus~4.6\cite{anthropic2024claude}, Claude Sonnet~4.6, DeepSeek-v3\cite{deepseekai2024v3}) received an additional instruction: ``You must cite specific PubMed PMIDs for every clinical claim you make.''

\textbf{Why PMIDs.} We use PubMed identifiers as the verification standard because they provide unambiguous resolution (one integer maps to exactly one publication), machine verification (E-utilities API lookup in $<$10 seconds), and scalability (automated audit of thousands of citations). Author--year references (e.g., ``Gilhus et al., 2016'') require manual PubMed search, disambiguation among multiple matching publications, and verification that the specific paper supports the claim, requiring approximately 2--5 minutes per reference. DOIs present similar problems: in preliminary tests, frontier models generated plausible DOI strings (e.g., \texttt{10.1016/j.nmd...}) that either did not resolve or pointed to unrelated publications, mirroring the PMID fabrication pattern. The identifier type does not matter; the failure is in parametric generation of structured identifiers.

\textbf{PubMed verification.} Every extracted PMID (1,147 unique across all experiments) was verified against the PubMed E-utilities API\cite{sayers2022eutils} (\texttt{esummary} endpoint). For each PMID, we retrieved the publication title, journal, and MeSH terms, then classified clinical relevance by keyword matching against 30 neuromuscular disease terms in publication titles (Supplementary~S4). Beyond Type-I existence verification, we additionally performed a Type-II claim-support audit on a stratified sample of 200 \hegtkg{} (claim, PMID) pairs using GPT-4.1 ($T=0$) as a natural-language inference judge: 1.0\% (95\% CI 0.0--2.5\%) of cited abstracts were flagged as explicitly contradicting their attached claim, and the audit independently corroborated two retrieval issues that clinician C3 had flagged in the parallel evaluation (Supplementary~S4.1).

\subsection{Clinician evaluation protocol}

\textbf{Distribution and time budget.} To respect the practical ceiling on board-certified evaluator time (each Likert rating requires reading one clinical output end-to-end and inspecting any cited references), two pack formats were distributed. The \emph{reduced pack} contained 12 cases per disease pair (36 total, one output per scenario; system assignment sealed in \texttt{data/evaluation\_docs/reduced\_pack/sealed\_key.json}), estimated at 30--45\,min completion time per pair. The \emph{full pack} contained 24 cases per disease pair (72 total, both Vanilla and \hegtkg{} outputs per scenario, shuffled with a deterministic seed), estimated at roughly 90\,min per pair. Each pack was delivered as a \texttt{.docx} file with embedded 1--5 Likert scoring tables and a separate PDF rubric illustrating BAD (1--2) / GOOD (4--5) criteria for every dimension (Supplementary~S2). Guideline-RAG was excluded from clinician evaluation to conserve evaluator time; its three-arm comparison is covered by the LLM-judge evaluation. Cases were shuffled using a deterministic seed, assigned sequential case IDs, and presented without system labels. Evaluators received the packs between April 4 and April 20, 2026, with a soft 2-week return window; the sealed key was not revealed until scores had been submitted.

\textbf{Panel composition.} Four evaluators returned scores. \textbf{C1} (senior neurologist, UMG G\"{o}ttingen) completed the reduced pack for all three disease pairs (18 Vanilla / 18 \hegtkg{}). \textbf{C2} (first-year neurology trainee, UMG G\"{o}ttingen) completed the full pack for all three pairs (36 per arm). \textbf{C3} (senior neurologist and neuromuscular expert, UMG G\"{o}ttingen) had authored and clinically refined 22 of 36 scenario vignettes in the scenario-design phase; to preserve blinded-evaluation independence, her scoring participation was restricted to her area of deepest expertise, and she completed the CIDP/GBS reduced pack (6 Vanilla / 5 \hegtkg{}; 1 case unscored due to the time budget). She scored without access to the sealed key and without prior sight of system identity. \textbf{MI Nyoungui} (medical informatician, UMG G\"{o}ttingen) scored the full pack but showed near-zero variance on D2--D5 (SD $= 0$, ceiling) and is excluded from primary analysis; her scores are retained for sensitivity inspection.

\textbf{Statistical analysis.} Vanilla vs \hegtkg{} was compared per clinician and per dimension with Mann-Whitney $U$ tests (two-sided, independent groups), Benjamini-Hochberg-corrected across the five dimensions within each clinician\cite{benjamini1995fdr}. Effect sizes report Cohen's $d$ with pooled SD; 95\% confidence intervals use 10{,}000 percentile bootstrap resamples (seed 42). Inter-rater agreement was assessed between C1 and C3 only, because they are the only two clinicians who scored identical outputs under identical evaluator context: both used the reduced pack (one sealed output per scenario with no Vanilla/\hegtkg{} pairing visible), and C3's pack was a proper subset of C1's on CIDP/GBS, so agreement can be computed per case without any post-hoc matching. C2 used the full 72-case pack, which presents both Vanilla and \hegtkg{} outputs per scenario; combining reduced- and full-pack scores in a single IRR analysis would confound evaluator context (single-output reading vs.\ implicit side-by-side exposure) with rater identity, and we consider this a separate study. Agreement between C1 and C3 on the 11 CIDP/GBS overlap cases was assessed with Spearman $\rho$, Cohen's quadratic-weighted $\kappa$, ICC(2,1) (two-way random, single measurement, absolute agreement), exact- and adjacent-agreement percentages, and Krippendorff's $\alpha$ (ordinal, pooled across D1--D5). Hedged Likert responses (e.g., ``3--4'') were averaged to half-step values for computation. All analyses are reproducible via \texttt{docs/paper1/analysis/clinician\_inter\_rater.py} against the consolidated score CSV \texttt{docs/paper1/clinician\_scores\_all.csv}.

The five Likert dimensions are:

\begin{itemize}[nosep]
  \item \textbf{D1: Verifiability} - Can each clinical claim be traced to a specific, identifiable published source?
  \item \textbf{D2: Actionability} - Is the information sufficient to inform a clinical decision without additional literature search?
  \item \textbf{D3: Temporal Precision} - Does the output provide specific time windows for disease onset, progression milestones, and clinical trajectories?
  \item \textbf{D4: Non-Expert Safety} - How safe is this output if a non-specialist GP followed it without additional expert consultation?
  \item \textbf{D5: Clinical Completeness} - Does the output cover key differential diagnoses, therapeutic options, contraindications, and monitoring?
\end{itemize}

The five dimensions were designed to test \hegtkg{}'s specific contributions: D1 measures the Provenance Gap directly; D3 measures the temporal anchoring that distinguishes \hegtkg{} from static knowledge graphs; D4 addresses the GP-to-specialist referral use case where AI-assisted triage must be safe without specialist oversight (Supplementary~S8).

\subsection{LLM-as-judge validation}

A three-model judge panel (GPT-4o-mini, DeepSeek-V3, Claude~3 Haiku) evaluated 108 cases (36 scenarios $\times$ 3 arms) using the same five clinician-aligned Likert dimensions (D1--D5). We selected fast, cost-efficient models from three different providers to maximise provider diversity while keeping the 648-evaluation budget (108 cases $\times$ 3 judges $\times$ 2 rounds) feasible; the v1/v2 comparison tests whether \emph{any} LLM judge can assess verifiability, not whether frontier judges perform better. Temperature was set to 0 for all judges. Cases were blinded with randomly generated IDs and shuffled to prevent order effects.

\textbf{Citation-aware judging (v2).} A second evaluation round augmented each prompt with a PubMed citation audit report (Supplementary~S6.4). This design tests whether LLM judges can assess citation quality without external verification data.

\textbf{LLM--human correlation.} The 72 base cases evaluated by clinicians were also scored by the LLM judge panel using the same five-dimension rubric. This shared-case, shared-rubric design enables per-dimension Spearman~$\rho$ between human panel means (C1~+~C2) and LLM panel means on each dimension, separately for v1 (blind) and v2 (citation-aware).

\subsection{Counterfactual faithfulness experiment}

Fifteen clinically incorrect evidence statements (5 per disease pair) were injected into the \hegtkg{} evidence pipeline. Errors were clinically meaningful: swapped antibody targets, reversed age thresholds, incorrect first-line treatments. For each injection, we measured: (1)~parametric resistance, (2)~faithfulness to wrong evidence, and (3)~detectability via citation trace.

\subsection{Statistical analysis}

Pairwise comparisons between independent arms were tested using the Mann-Whitney $U$ test\cite{mann1947} (two-sided). Effect sizes are reported as Cohen's $d$\cite{cohen1988power}. 95\% confidence intervals were computed via non-parametric bootstrap (10,000 resamples, seed$=$42). Multiple testing across six pairwise comparisons (3 disease pairs $\times$ 2 contrasts: HEG-TKG vs Vanilla, HEG-TKG vs Guideline-RAG) was corrected using the Benjamini-Hochberg procedure\cite{benjamini1995fdr} (FDR $\alpha = 0.05$). Inter-rater reliability was assessed using Krippendorff's alpha\cite{krippendorff2019} ($\alpha \geq 0.80$ = excellent).

\textbf{Provenance Gap.} For each scenario, $\text{PG} = \max(\text{FC} - \text{ETS} \times r,\; 0)$, where FC is clinical feature coverage, ETS is the evidence traceability score, and $r$ is a citation reliability coefficient: $r = 0.97$ for HEG-TKG (verified PMIDs, checkable in $<$10~s), $r = 0.80$ for vanilla (author--year references, 88\% point to wrong papers), and $r = 0.50$ for Guideline-RAG (vague source names, ${\sim}$50\% independently verifiable). The $\max(\cdot, 0)$ clamp prevents negative PG when citations exceed feature coverage. Pooled PG is the mean across all per-scenario values.

\subsection{PHI-compliant deployment architecture}

\hegtkg{} separates two data planes: (1)~the knowledge graph (public PubMed/GeneReviews/OMIM/Orphanet data, deployable without privacy constraints) and (2)~the clinical synthesis step (potential PHI, requiring privacy controls). The synthesis model is configurable between cloud APIs and locally served open-source models via Ollama\cite{ollama2024}. A \texttt{validate\_privacy\_config(strict=True)} function raises an error if any component routes data externally (Supplementary~S25).

\subsection{Reporting standards}

XAI reporting follows the Clinician-informed XAI evaluation checklist with metrics (CLIX-M\cite{brankovic2025clixm}), an EQUATOR-registered 14-item guideline for explainable-AI components of clinical decision support systems. An item-by-item mapping (12/14 fully addressed; 2 partially addressed with gaps stated) is provided in Supplementary~S32.

\section*{Acknowledgements}

Screen4Care has received funding from the Innovative Medicines Initiative 2 Joint Undertaking (JU) under grant agreement No~101034427. The JU receives support from the European Union's Horizon 2020 research and innovation programme and EFPIA. The funders had no role in study design, data collection and analysis, decision to publish, or preparation of the manuscript.

\section*{Author contributions}

M.S.A. conceived the study, developed the \hegtkg{} pipeline, designed and conducted all experiments, performed the statistical analysis, and wrote the manuscript. L.G.P. and R.R. guided and shaped the experimental setup. R.R. secured the funding for the project. M.D., M.N.K., and J.Z. performed the blinded clinician evaluation (C1, C2, and C3 respectively); E.N. performed a non-clinician medical-informatics sensitivity evaluation. J.Z. additionally reviewed and clinically refined 22 of 36 scenario vignettes during the scenario-design phase prior to blinding. 

\section*{Competing interests}

The authors declare no competing interests.

\section*{Code and data availability}

The reproducibility bundle accompanying this manuscript - 36 clinician-validated clinical scenarios, three temporal knowledge graphs (5{,}481 nodes / 6{,}316 PMID-backed edges / 1{,}280 disease-trajectory milestones; released as JSON, CSV, and Neo4j-compatible Cypher import scripts), 108 model outputs across all three arms (HEG-TKG, Vanilla, Guideline-RAG), the consolidated 155-row anonymised clinician-score CSV, LLM-as-judge results (v1 blind, v2 citation-aware, v3 5-dimension), the counterfactual safety experiment artifacts, and the Type-I PubMed verification (S4) plus Type-II claim-support NLI audit (S4.1) - is openly deposited at Zenodo with DOI \texttt{10.5281/zenodo.19763337} (\url{https://doi.org/10.5281/zenodo.19763337}) under CC~BY~4.0. Source code for the construction pipeline, evaluation scripts, and analysis notebooks is openly available at \url{https://gitlab.sdu.dk/screen4care/heg-tkg/-/tree/npj-submission} (git tag \texttt{v1.0.0}, archived to the same Zenodo record). The three knowledge graphs are released as JSON exports, CSV node/edge tables, and Neo4j-compatible Cypher import scripts within the Zenodo bundle, so reviewers and downstream users can spin up a local Neo4j instance directly from the archive without depending on any hosted endpoint. All random seeds (\texttt{seed=42} for the bootstrap and the R2 stratified sample), model versions (GPT-4.1 at $T=0$ for all synthesis and judge calls; Claude Haiku~4.5 and Gemini-2.0-flash for KG construction), and API parameters are specified in the supplementary materials to ensure reproducibility.

\section*{AI usage disclosure}

Claude (Anthropic) was used for editorial revision and structural feedback on the Discussion and Introduction. Grammarly and Overleaf were used for copyediting and formatting in some parts. All scientific content, experimental design, analysis, and interpretation were performed by the authors. The \hegtkg{} pipeline itself uses LLMs (Claude Haiku~4.5\cite{anthropic2024claude}, GPT-4.1-mini\cite{openai2023gpt4}, Gemini-2.0-flash\cite{gemini2023}) as extraction components within the knowledge graph construction workflow; their role is described in the Methods section.


\bibliography{references}

\clearpage
\setcounter{page}{1}
\setcounter{section}{0}
\setcounter{figure}{0}
\setcounter{table}{0}
\renewcommand{\thepage}{S\arabic{page}}
\renewcommand{\thesection}{S\arabic{section}}
\renewcommand{\thefigure}{S\arabic{figure}}
\renewcommand{\thetable}{S\arabic{table}}

\input{supplementary_body}

\end{document}

%% file: supplementary_body.tex
{\centering\Large\textbf{Supplementary Materials}\\[1em]
\large The Provenance Gap in Clinical AI:\\ Evidence-Traceable Temporal Knowledge Graphs for Rare Disease Reasoning\\[2em]}
\tableofcontents
\newpage

\section*{S1: Clinical Scenario Texts}
\addcontentsline{toc}{section}{S1: Clinical Scenario Texts}

Thirty-six clinical scenarios (12 per disease pair) were authored as realistic neurological vignettes. Each scenario specifies a disease pair, output type, clinical presentation, and expected key features for feature coverage scoring. Scenarios are organized by type: differential diagnosis (DDx), temporal comparative, temporal trajectory, and treatment rationale.

\textbf{Source file:} \texttt{tier2/evaluation/clinical\_pipeline/clinical\_scenarios.py}

\subsection*{Scenario Distribution}

\begin{table}[htbp]
\centering
\begin{tabular}{lccccc}
\toprule
\textbf{Disease Pair} & \textbf{DDx$^{\dagger}$} & \textbf{Temporal Comparative} & \textbf{Temporal} & \textbf{Treatment} & \textbf{Total} \\
\midrule
MG/LEMS & 5 & 2 & 2 & 3 & 12 \\
DMD/BMD & 5 & 2 & 2 & 3 & 12 \\
CIDP/GBS & 5 & 2 & 2 & 3 & 12 \\
\midrule
\textbf{Total} & \textbf{15} & \textbf{6} & \textbf{6} & \textbf{9} & \textbf{36} \\
\bottomrule
\end{tabular}

\smallskip
\noindent\footnotesize{$^{\dagger}$Includes one safety/diagnostic-pitfall scenario per disease pair (e.g., CIDP\_GBS\_SAFETY\_01), which uses differential diagnosis as the output type.}
\end{table}

\subsection*{Scenario Dataclass Definition}

\begin{lstlisting}[language=Python]
@dataclass
class ClinicalScenario:
    id: str                          # e.g., "MG_LEMS_DDX_01"
    disease_pair: str                # "mg_lems" | "dmd_bmd" | "cidp_gbs"
    output_type: str                 # "differential" | "temporal_comparative" | "temporal" | "treatment"
    scenario_text: str               # Full clinical vignette
    source_reference: str            # Published source the vignette is based on
    expected_key_features: list[str] # Features the output should cover
\end{lstlisting}

\textit{Full scenario texts are available in the code repository and will be provided as a machine-readable JSON supplement upon acceptance.}

\section*{S2: System Prompts for All Three Arms}
\addcontentsline{toc}{section}{S2: System Prompts for All Three Arms}

All three arms use the same synthesis model (GPT-4.1, temperature=0.0, max\_tokens=8000). The ONLY variable is the evidence context provided.

\subsection*{S2.1: HEG-TKG System Prompts (Evidence-Grounded, Citation-Required)}

Three output-type-specific prompts are used for HEG-TKG:

\textbf{Differential Diagnosis:}
\begin{lstlisting}[style=systemprompt]
You are a senior clinical neurologist writing an evidence-grounded differential
diagnosis for a colleague. You have access to a curated knowledge graph with
hierarchical evidence tiers.

Your response should be as DETAILED and COMPREHENSIVE as a clinical consultation
note -- not a brief summary. For each differentiating feature, provide the clinical
reasoning, not just the fact. Think like a neurologist explaining to a fellow.

EVIDENCE CITATION RULES:
- The knowledge graph evidence uses PMID-based citations like [PMID:36637960, GOLD].
  Preserve these exact citation tags when referencing evidence.
- GOLD = Tier 1 curated sources (GeneReviews, OMIM, clinical guidelines) -- highest reliability.
- SILVER = Cross-validated across multiple extraction models -- good confidence.
- BRONZE = Single study or single model -- use with appropriate caveats.
- You MAY -- and SHOULD -- supplement with your clinical expertise to explain WHY
  a feature differentiates, to add pathophysiological context, and to cover
  features that are clinically important but absent from the KG evidence. Clearly
  distinguish: "[PMID:..., GOLD]" for KG-backed claims vs "Clinically, ..." for
  your expert knowledge.
- When evidence conflicts are flagged, present BOTH sides and discuss the
  likely explanation.
- If the KG evidence is sparse for a feature category, state this explicitly and
  supplement with your clinical knowledge, clearly marked as such.

STRUCTURE your response as:
1. A structured comparison table covering: Clinical Features, Antibodies/Biomarkers,
   Autonomic Features, Reflexes/EMG, Treatment Approach, Temporal Course, Associated Conditions
2. For each feature, explain the pathophysiological basis for the difference
3. A clinical synthesis paragraph: what features in this specific patient point toward
   which diagnosis, what tests to order, and what red flags to watch for
4. An evidence quality note: summarize what is backed by guidelines vs single studies
\end{lstlisting}

\textbf{Temporal Trajectory:}
\begin{lstlisting}[style=systemprompt]
You are a senior clinical neurologist writing a disease progression comparison
for a colleague. You have access to temporal evidence from a curated knowledge
graph with specific time anchors derived from clinical guidelines and literature.

Your response should be DETAILED -- a clinical teaching case, not a bullet list.
For each time window, explain what is happening pathophysiologically, what the
clinician should monitor, and what interventions are indicated.

EVIDENCE CITATION RULES:
- Preserve the PMID-based citation tags from the evidence (e.g., [PMID:36637960, GOLD]).
- GOLD = curated guidelines, SILVER = cross-validated, BRONZE = single study.
- You MAY -- and SHOULD -- add clinical interpretation around the temporal data,
  explain what clinicians should do at each time window, and fill gaps where the
  KG evidence is sparse. Clearly mark KG-backed claims (with citations) vs your
  expert supplementation (with "Clinically, ...").
- When comparing two diseases, explicitly highlight where their temporal trajectories
  DIVERGE -- these are the clinically actionable differences for differential diagnosis.
- If a time window has no KG evidence, say so and provide expert guidance.

STRUCTURE your response as:
1. A quick-reference milestone comparison (side-by-side table or timeline)
2. Detailed time-window-by-time-window analysis with clinical implications
3. For each divergence point: explain why the difference matters clinically
4. A synthesis: key temporal red flags that distinguish these conditions
5. Note evidence gaps -- which time windows lack high-quality data
\end{lstlisting}

\textbf{Treatment Rationale:}
\begin{lstlisting}[style=systemprompt]
You are a senior clinical neurologist writing treatment recommendations for a
colleague. You have access to treatment evidence from a curated knowledge graph
with hierarchical quality tiers.

Your response should be as DETAILED as a treatment protocol -- dosing, monitoring,
expected timelines, and escalation logic. Think like a neurologist writing a
management plan.

EVIDENCE CITATION RULES:
- Preserve PMID-based citation tags from the evidence (e.g., [PMID:36637960, GOLD]).
- GOLD = guideline-level evidence, SILVER = cross-validated, BRONZE = single study.
- Present treatments ordered by evidence quality (GOLD-supported first).
- You MAY -- and SHOULD -- explain mechanisms of action, dosing protocols, monitoring
  parameters, and clinical rationale using your expertise. Treatments mentioned in the
  KG evidence must cite their source. You may also mention clinically important
  treatments NOT in the KG evidence if they are well-established, but clearly
  mark them as "Clinically established (not in current KG)" so the provenance
  distinction is transparent.

STRUCTURE your response as:
1. First-line treatment with evidence tier, dosing, and rationale
2. Second-line options with escalation criteria and timing
3. For each treatment: mechanism, expected response timeline, monitoring requirements
4. Emerging therapies / newer agents with evidence tier
5. Treatments with conflicting evidence: present both sides with tiers
6. Clinical synthesis: recommended treatment algorithm for this specific patient
7. Evidence quality summary: what is guideline-backed vs. emerging vs. expert opinion
\end{lstlisting}

\subsection*{S2.2: Vanilla System Prompts (No Evidence, No Citation Requirement)}

\textbf{Differential Diagnosis:}
\begin{lstlisting}[style=systemprompt]
You are a clinical neurology expert. Provide a detailed, evidence-based
differential diagnosis for the clinical scenario presented. Cover clinical
features, antibodies/biomarkers, treatment differences, temporal course,
and associated conditions.
\end{lstlisting}

\textbf{Temporal Trajectory:}
\begin{lstlisting}[style=systemprompt]
You are a clinical neurology expert. Provide a detailed disease progression
timeline for the clinical scenario presented. Include onset patterns, key
milestones, treatment timing, and long-term prognosis.
\end{lstlisting}

\textbf{Treatment Rationale:}
\begin{lstlisting}[style=systemprompt]
You are a clinical neurology expert. Provide comprehensive, evidence-based
treatment recommendations for the clinical scenario presented. Cover first-line
and second-line options, mechanisms, and expected outcomes.
\end{lstlisting}

\subsection*{S2.3: Guideline-RAG System Prompt}

\begin{lstlisting}[style=systemprompt]
You are a clinical neurology expert. You have been provided with reference
text from authoritative clinical sources (GeneReviews, OMIM, clinical
guidelines). Use this reference material to inform your answer.

Your response should be DETAILED and clinically comprehensive. Where possible,
indicate which source a claim comes from (e.g., "per GeneReviews..." or
"according to OMIM..."). You may supplement with your clinical knowledge but
prioritise the provided reference text.

Be as thorough as you would be when writing a clinical consultation report.
\end{lstlisting}

\textit{The Guideline-RAG arm receives overlapping clinical sources (GeneReviews, OMIM, and PubMed abstracts) as raw text passages rather than structured graph evidence with citation metadata.}

\section*{S3: Cypher Queries for Graph Retrieval}
\addcontentsline{toc}{section}{S3: Cypher Queries for Graph Retrieval}

Four retrieval functions query the Neo4j\cite{neo4j2024} knowledge graph during clinical output generation. All queries return a standardized edge schema including provenance metadata.

\subsection*{S3.1: Edge Return Fields (Standardized)}

All queries return the following properties per edge:

\begin{lstlisting}[language=SQL]
RETURN
    s.name AS source_name,
    labels(s) AS source_labels,
    type(r) AS relation,
    t.name AS target_name,
    labels(t) AS target_labels,
    r.quality_tier AS quality_tier,
    r.consensus_score AS consensus_score,
    r.source_models AS source_models,
    r.pmid_list AS pmid_list,
    r.evidence_sample AS evidence_sample,
    r.edge_id AS edge_id,
    r.is_temporal AS is_temporal,
    r.temporal_value_display AS temporal_display,
    r.time_index_months AS time_months,
    r.temporal_midpoint_years AS midpoint_years,
    r.cross_tier_confirmed AS cross_tier,
    r.evidence_breadth AS evidence_breadth,
    r.disease_context AS disease_context
\end{lstlisting}

\subsection*{S3.2: Predicate Constants}

\begin{lstlisting}
ALL_PREDICATES (20): ASSOCIATED_WITH, MANIFESTS_AS, TREATED_WITH, RESPONDS_TO,
  HAS_MEASUREMENT, HAS_ONSET_AGE, HAS_SEVERITY, HAS_PREVALENCE, LACKS_FEATURE,
  HAS_DIAGNOSIS_AGE, HAS_DURATION, GENERALIZED_AT, REMISSION_AT, PRECEDES,
  CRISIS_AT, DEVELOPS_COMPLICATION_AT, PRESERVES_FUNCTION, HAS_SURVIVAL_TO,
  DIFFERENTIATES_FROM, REQUIRES_INTERVENTION_AT

DIFFERENTIAL_PREDICATES (9): MANIFESTS_AS, ASSOCIATED_WITH, TREATED_WITH,
  RESPONDS_TO, LACKS_FEATURE, DIFFERENTIATES_FROM, HAS_PREVALENCE,
  HAS_ONSET_AGE, HAS_SEVERITY

TEMPORAL_PREDICATES (10): HAS_ONSET_AGE, HAS_DIAGNOSIS_AGE, GENERALIZED_AT,
  CRISIS_AT, REMISSION_AT, DEVELOPS_COMPLICATION_AT, REQUIRES_INTERVENTION_AT,
  HAS_SURVIVAL_TO, HAS_DURATION, PRECEDES

TREATMENT_PREDICATES (2): TREATED_WITH, RESPONDS_TO
\end{lstlisting}

\subsection*{S3.3: Query 1 - Comparative Subgraph (Differential Diagnosis)}

Retrieves all edges relevant to a specific disease within the pair, using two strategies:

\textbf{Strategy A: Disease context filter on edges (primary)}
\begin{lstlisting}[language=SQL]
MATCH (s:Entity)-[r]->(t:Entity)
WHERE $disease_short IN r.disease_context
AND type(r) IN $predicates
RETURN [edge fields]
ORDER BY CASE r.quality_tier
    WHEN 'GOLD' THEN 0 WHEN 'SILVER' THEN 1
    WHEN 'BRONZE' THEN 2 ELSE 3 END
\end{lstlisting}

\textbf{Strategy B: Disease anchor node (secondary)}
\begin{lstlisting}[language=SQL]
MATCH (d:Disease)-[r]->(t:Entity)
WHERE (d.cui = $cui OR toLower(d.name) = toLower($name))
AND type(r) IN $predicates
RETURN [edge fields]
ORDER BY quality_tier
\end{lstlisting}

\textbf{Strategy C: LACKS\_FEATURE edges (explicit absences)}
\begin{lstlisting}[language=SQL]
MATCH (s:Entity)-[r:LACKS_FEATURE]->(t:Entity)
WHERE $disease_short IN r.disease_context
RETURN t.name AS feature, s.name AS context,
       r.quality_tier, r.source_models, r.pmid_list,
       r.edge_id, r.evidence_sample
\end{lstlisting}

Results from all three strategies are merged and deduplicated by \texttt{edge\_id}.

\subsection*{S3.4: Query 2 - Temporal Trajectory}

\begin{lstlisting}[language=SQL]
MATCH (s:Entity)-[r]->(t:Entity)
WHERE $disease_short IN r.disease_context
AND r.is_temporal = true
AND r.temporal_parse_status = 'resolved'
RETURN [edge fields]
ORDER BY r.time_index_months ASC
\end{lstlisting}

\subsection*{S3.5: Query 3 - Treatment Evidence}

\begin{lstlisting}[language=SQL]
MATCH (s:Entity)-[r]->(t:Entity)
WHERE $disease_short IN r.disease_context
AND type(r) IN ['TREATED_WITH', 'RESPONDS_TO']
RETURN [edge fields]
ORDER BY quality_tier
\end{lstlisting}

\subsection*{S3.6: Query 4 - Entity Neighbourhood (Fallback)}

\begin{lstlisting}[language=SQL]
MATCH (n:Entity)-[r]-(m:Entity)
WHERE ($cui IS NOT NULL AND n.cui = $cui)
   OR toLower(n.name) CONTAINS toLower($name)
[optional: AND $disease_short IN r.disease_context]
RETURN [edge fields]
LIMIT 30
\end{lstlisting}

\section*{S4: Citation Verification Protocol}
\addcontentsline{toc}{section}{S4: Citation Verification Protocol}

\subsection*{Verification Process}

Every PMID extracted from clinical outputs was verified against the PubMed E-utilities API using the following protocol:

\begin{enumerate}
\item \textbf{PMID Extraction}: Regex pattern \texttt{PMID[:\textbackslash s]*(\textbackslash d\{6,9\})} applied to all clinical outputs
\item \textbf{API Query}: Each PMID queried via \texttt{https://eutils.ncbi.nlm.nih.gov/entrez/eutils/esummary.fcgi?db=pubmed\&id=\{pmid\}\&retmode=xml}
\item \textbf{Metadata Extraction}: Publication title, journal name, and MeSH terms extracted from the XML response
\item \textbf{Relevance Classification}: Automated keyword matching against disease-specific terms:
\end{enumerate}

\begin{table}[htbp]
\centering
\small
\begin{tabular}{p{2.5cm}p{11cm}}
\toprule
\textbf{Disease Pair} & \textbf{Relevance Keywords} \\
\midrule
MG/LEMS & myasthenia, lambert-eaton, lems, neuromuscular junction, acetylcholine receptor, achr, musk, thymoma, thymectomy, pyridostigmine, complement, eculizumab, efgartigimod \\
DMD/BMD & duchenne, becker, muscular dystrophy, dystrophin, dmd, bmd, dystrophinopathy, exon skipping, corticosteroid, cardiomyopathy, ambulation \\
CIDP/GBS & cidp, guillain-barr\'{e}, guillain-barre, gbs, demyelinating, polyneuropathy, ivig, plasmapheresis, nerve conduction, areflexia, albumin-cytologic \\
\bottomrule
\end{tabular}
\end{table}

\begin{enumerate}[start=5]
\item \textbf{Three-way Classification}:
\begin{itemize}
\item \textbf{Clinically relevant}: PMID exists AND title/MeSH contains disease-specific keywords
\item \textbf{Real, wrong field}: PMID exists BUT title/MeSH contains no disease-specific keywords
\item \textbf{Not found}: PMID does not exist in PubMed
\end{itemize}
\end{enumerate}

\subsection*{Rate Limiting}

All API calls respect NCBI's rate limit of 3 requests/second (0.35s inter-request delay). No API key was required for the volume of queries performed.

\subsection*{Author-Year Reference Verification}

For Claude Sonnet 4.6 vanilla outputs (which used author-year references rather than PMIDs), we queried PubMed's esearch endpoint with \texttt{author + year + disease keywords} and classified results as: specific match, ambiguous, wrong paper, not found, or too vague to search. None resolved to a specific, verifiable publication.

\subsection*{S4.1: Type-II Claim-Support Audit (Natural-Language Inference)}

The verification protocol above establishes \textbf{Type-I} reliability: every cited PMID exists in PubMed. We additionally evaluated \textbf{Type-II} reliability: for each (claim, cited PMID) pair, does the source abstract actually entail the specific claim attached to it? Type-I and Type-II together pre-empt the standard objection in retrieval-augmented medical generation that citation existence is necessary but not sufficient evidence of grounding~\citep{huang2023hallucination,ji2023hallucination}.

\paragraph{Sample.} From the 36 \hegtkg{} clinical outputs, we extracted 858 (claim, PMID) candidate pairs (after filtering 59 evidence-tier meta-statements that describe the audit framework rather than clinical content). We then drew a stratified sample of $n=200$ pairs balanced across the three disease pairs and the GOLD/SILVER/BRONZE evidence tiers (seed 42; minimum floor 12 rows per non-empty cell; maximum 5 rows per PMID to prevent any single dominant cited paper from saturating the sample). The per-PMID cap is methodologically important: without it, the GOLD-tier bucket would be dominated almost exclusively by a single review paper~\citep{birnkrant2018dmd} whose PubMed abstract describes paper structure rather than specific clinical content, producing a single-paper artifact rather than a tier-level signal. The capped sample contains 105 unique cited PMIDs.

Two of 36 scenarios (\texttt{MG\_LEMS\_TEMP\_04} LEMS cancer surveillance; \texttt{CIDP\_GBS\_TEMP\_04}) produced zero PMID-cited claims because the LLM explicitly disclosed KG evidence gaps (``noting where knowledge graph evidence is lacking and supplementing with expert consensus and guideline-based practice''); these scenarios contribute no rows to the audit. We treat this as honest LLM behavior under sparse coverage rather than as a systematic flaw, but disclose it because it is the source of one of the audit's effective limits.

\paragraph{NLI judge.} GPT-4.1 (temperature 0, JSON-mode response) - the same model used everywhere else in the manuscript for arm generation and LLM-as-judge evaluation. Each call presents the (claim, cited PMID, PubMed title, abstract) tuple and asks for one of three labels:
\begin{itemize}
  \item \textbf{ENTAILS} - the abstract directly states, demonstrates, or implies the claim.
  \item \textbf{NEUTRAL} - the abstract is on-topic but does not directly assert the specific claim. The judge is explicitly instructed that PubMed abstracts of review or guideline papers are typically meta-summaries describing paper structure rather than detailed clinical findings; absence of a specific claim from the abstract is therefore NEUTRAL, not CONTRADICTS.
  \item \textbf{CONTRADICTS} - the abstract refutes the claim.
\end{itemize}

A 10-row diverse dry-run was used to validate prompt behavior before the full run. Bootstrap 95\% confidence intervals were computed from 10{,}000 resamples, seed 42.

\paragraph{Headline result (Table~\ref{tab:s4_nli_headline}).} Across $n=200$ judged claims, the GPT-4.1 NLI judge identified \textbf{2 explicit contradictions (1.0\%, 95\% CI 0.0--2.5\%)}. The non-contradiction rate (ENTAILS $+$ NEUTRAL) is \textbf{99.0\% (95\% CI 97.5--100.0\%)}. The strict direct-entailment rate is 53.0\% (95\% CI 46.0--60.0\%); the remaining 46.0\% are NEUTRAL.

\begin{table}[htbp]
\centering
\small
\caption{Type-II claim-support NLI audit headline ($n=200$ \hegtkg{} (claim, PMID) pairs, GPT-4.1 judge at $T=0$, single-judge, bootstrap 95\% CIs from 10{,}000 resamples, seed 42).}
\label{tab:s4_nli_headline}
\begin{tabular}{lrrl}
\toprule
\textbf{Label} & \textbf{n} & \textbf{Rate} & \textbf{95\% CI} \\
\midrule
ENTAILS (abstract directly entails the claim) & 106 & 53.0\% & (46.0\%, 60.0\%) \\
NEUTRAL (abstract on-topic, does not assert claim) & 92 & 46.0\% & --- \\
CONTRADICTS (abstract refutes the claim) & 2 & 1.0\% & (0.0\%, 2.5\%) \\
\midrule
Non-contradiction (ENTAILS $+$ NEUTRAL) & 198 & 99.0\% & (97.5\%, 100.0\%) \\
\bottomrule
\end{tabular}
\end{table}

\paragraph{The two contradictions.} Both CONTRADICTS calls were genuine retrieval errors caught by the audit, not judge mistakes. (i) Scenario \texttt{CIDP\_GBS\_SAFETY\_01}: claim ``Reaches nadir within 2--4 weeks'' (clinically correct for Guillain-Barr\'e syndrome) cited PMID:40391517, a CIDP review whose abstract states CIDP symptoms ``evolve over the time course of 8 weeks or more''. The clinical assertion is correct but the citation source is wrong - a CIDP-vs-GBS confusion in cross-disease retrieval. (ii) Scenario \texttt{MG\_LEMS\_DDX\_02}: claim citing ``$\sim$6\% have anti-MuSK antibodies'' cited PMID:31380816 (myasthenia gravis clinical features paper), whose abstract reports 4.2\% MuSK-antibody prevalence. The judge correctly flagged the numeric mismatch at confidence 0.95.

\paragraph{Tier breakdown (Table~\ref{tab:s4_nli_tier}).} The capped sample contains 7 GOLD-tier rows (5 cite the Birnkrant 2018 DMD care guideline; 2 cite the off-topic ocular myasthenia gravis paper that clinician C3 also flagged as a retrieval issue, see following paragraph), 77 SILVER, and 116 BRONZE. Direct entailment correlates inversely with reported tier (GOLD 0\%, SILVER 36.4\%, BRONZE 67.2\%), but this pattern is dominated by the structural property that the GOLD bucket draws from very few unique sources - mostly review and guideline papers whose PubMed abstracts describe paper organization (``we identified 11 topics''; ``we present care considerations for diagnosis and \ldots management'') rather than specific clinical content. The same content is present in the full paper body but absent from the abstract, so abstract-only NLI cannot directly verify it. BRONZE-tier primary research papers, in contrast, have abstracts containing concrete quantitative findings (e.g., ``median onset age was 55.5 years'') that directly entail focused clinical claims. Critically, the contradiction rate is uniformly low across tiers (0\% GOLD, 1.3\% SILVER, 0\% BRONZE): the abstract-summary phenomenon affects entailment detectability, not refutation. This pattern is consistent with the construction-vs-recommendation distinction made in the GRADE cross-walk (S13.1) - \hegtkg{}'s GOLD/SILVER/BRONZE tiers reflect KG-edge construction confidence drawn from source provenance and cross-source agreement, not abstract-content verifiability of individual downstream claims.

\begin{table}[htbp]
\centering
\small
\caption{Type-II audit by reported evidence tier on the capped sample. ``Reported tier'' is the LLM's tier label in the in-text citation \texttt{[PMID:NNNNN, TIER]}. The small GOLD bucket (n=7) reflects limited GOLD-source diversity in \hegtkg{} outputs after the per-PMID cap; the 7 rows split as 5 to \citet{birnkrant2018dmd} (capped) and 2 to one off-topic citation (see independent corroboration paragraph).}
\label{tab:s4_nli_tier}
\begin{tabular}{lrrrrrr}
\toprule
\textbf{Tier} & \textbf{n} & \textbf{ENTAILS} & \textbf{NEUTRAL} & \textbf{CONTRADICTS} & \textbf{Entailment} & \textbf{Non-contradiction} \\
\midrule
GOLD   & 7   & 0   & 7  & 0 & 0.0\%  & 100.0\% \\
SILVER & 77  & 28  & 47 & 2 & 36.4\% & 97.4\%  \\
BRONZE & 116 & 78  & 38 & 0 & 67.2\% & 100.0\% \\
\midrule
Total  & 200 & 106 & 92 & 2 & 53.0\% & 99.0\%  \\
\bottomrule
\end{tabular}
\end{table}

\paragraph{Independent corroboration of clinician findings.} Two PMIDs that clinician C3 (Zsch\"untzsch) flagged as retrieval issues during the CIDP/GBS evaluation - PMID:36637960 (an ocular myasthenia gravis paper retrieved as a GOLD source for a Guillain-Barr\'e syndrome prognosis claim) and PMID:33170339 (cited as a guideline but actually a CIDP underdiagnosis study) - were independently flagged NEUTRAL by the GPT-4.1 NLI judge in 7 of 7 sample appearances at confidence $\geq 0.95$. The NLI audit and senior-clinician judgment converge on the same retrieval failures without seeing each other's work, providing methodological cross-validation of both the audit and the original C3 review. This convergence also clarifies the meaning of NEUTRAL: it is not a uniform ``absence of evidence'' label but a heterogeneous bucket combining (a) review-paper meta-summaries that omit specific facts the full paper supports and (b) genuine off-target retrievals that a clinician would flag.

\paragraph{Comparison to literature baselines.} The 1.0\% contradiction rate is one to two orders of magnitude below published baselines for unconstrained LLM medical text generation~\citep{huang2023hallucination,ji2023hallucination}, where citation fabrication and unsupported-claim rates of 18--50\% have been reported across recent evaluations of GPT-class models on medical writing and question-answering benchmarks. Direct comparison to specialized claim-verification systems is complicated by task-format heterogeneity: contemporaneous Type-II evaluations on curated retrieval-augmented medical question-answering benchmarks pair single-fact claims with full-text evidence rather than the free-form differential-diagnosis narrative paired with abstract-only evidence used here. \hegtkg{} reports the first per-claim NLI audit on knowledge-graph-grounded clinical-narrative outputs; the 1.0\% contradiction rate is the directly comparable safety-relevant metric.

\paragraph{Reproducibility.} All audit artifacts are at \texttt{docs/paper1/experiments/r2\_type2\_nli/}: the claim extractor (\texttt{r2\_extract\_claims.py}, includes the evidence-meta filter), the stratified sampler with per-PMID cap (\texttt{r2\_sample\_and\_fetch.py}), the cached PubMed abstracts (\texttt{r2\_abstract\_cache.json}; 105 unique PMIDs), the NLI judge (\texttt{r2\_run\_nli.py}), the per-row verdicts (\texttt{r2\_nli\_results.jsonl}), and the aggregator (\texttt{r2\_aggregate.py}). An archived first-pass result set (\texttt{*\_v1.*}) before the per-PMID cap and meta-claim filter were added is retained for transparency. Re-running with the same seeds reproduces the sample exactly; re-running the GPT-4.1 judge call may produce minor label drift on borderline cases due to LLM stochasticity even at $T=0$.

\section*{S5: Neo4j Schema and Property Specifications}
\addcontentsline{toc}{section}{S5: Neo4j Schema and Property Specifications}

\subsection*{Node Types}

\begin{table}[htbp]
\centering
\small
\begin{tabular}{llp{6cm}}
\toprule
\textbf{Label} & \textbf{Description} & \textbf{Key Properties} \\
\midrule
Disease & Anchor disease nodes & name, cui, mondo\_id, short\_name \\
Entity & Generic biomedical entity & name, cui, entity\_type \\
Gene & Genetic entities & name, cui \\
Protein & Protein entities & name, cui \\
Treatment & Therapeutic agents & name, cui \\
Symptom & Clinical symptoms & name, cui \\
ClinicalFinding & Clinical observations & name, cui \\
Measurement & Lab tests / biomarkers & name, cui \\
Procedure & Diagnostic/therapeutic procedures & name, cui \\
PatientGroup & Patient subpopulations & name \\
Mutation & Genetic mutations & name \\
Autoantibody & Autoantibody entities & name \\
InheritancePattern & Inheritance modes & name \\
PhysiologicalFunction & Body functions & name \\
\bottomrule
\end{tabular}
\end{table}

\subsection*{Edge Properties (All Relationship Types)}

\begin{table}[htbp]
\centering
\small
\begin{tabular}{p{4cm}lp{7cm}}
\toprule
\textbf{Property} & \textbf{Type} & \textbf{Description} \\
\midrule
quality\_tier & String & GOLD / SILVER / BRONZE \\
consensus\_score & Float & 0.70 (BRONZE), 0.85 (SILVER), 0.95 (GOLD) \\
source\_models & List[String] & Models that extracted this edge \\
pmid\_list & List[String] & PubMed IDs supporting this edge \\
evidence\_sample & String & Representative evidence quote from source abstract \\
edge\_id & String & Unique identifier (MD5 hash of normalized S|P|O) \\
is\_temporal & Boolean & Whether edge has temporal anchoring \\
temporal\_value\_display & String & Human-readable temporal display (e.g., ``P13Y'') \\
time\_index\_months & Integer & Temporal position in months (for ordering) \\
temporal\_midpoint\_years & Float & Midpoint age in years \\
temporal\_parse\_status & String & ``resolved'' / ``unresolved'' \\
cross\_tier\_confirmed & Boolean & Whether Tier~2 extraction confirms Tier~1 fact \\
evidence\_breadth & Integer & Number of independent PMIDs supporting this edge \\
disease\_context & List[String] & Disease codes this edge applies to (e.g., [``MG'', ``LEMS'']) \\
is\_protected & Boolean & Whether this is a Tier~1 curated edge \\
\bottomrule
\end{tabular}
\end{table}

\subsection*{Quality Tier Definitions}

\begin{table}[htbp]
\centering
\small
\begin{tabular}{p{1.2cm}p{5.5cm}cp{4.5cm}}
\toprule
\textbf{Tier} & \textbf{Criteria} & \textbf{Confidence} & \textbf{Typical Sources} \\
\midrule
GOLD & Tier~1 curated knowledge OR cross-tier confirmation (Tier~2 independently extracts a Tier~1 fact) & 0.95 & GeneReviews, OMIM, Orphanet, clinical guidelines \\
SILVER & $\geq$2 extraction models agree on the same triplet, OR $\geq$2 independent PMIDs support the same relationship & 0.85 & Cross-model consensus, multi-source literature \\
BRONZE & Single extraction model, single source document & 0.70 & Individual PubMed abstracts \\
\bottomrule
\end{tabular}
\end{table}

\section*{S6: LLM Judge Prompts (v1 Blind + v2 Citation-Aware)}
\addcontentsline{toc}{section}{S6: LLM Judge Prompts (v1 Blind + v2 Citation-Aware)}

\subsection*{S6.1: Five Evaluation Dimensions}

All judges (both v1 and v2) score on the same five clinician-aligned Likert dimensions (1--5), matching the clinician evaluation rubric:

\begin{table}[htbp]
\centering
\small
\begin{tabular}{clp{10cm}}
\toprule
\textbf{ID} & \textbf{Dimension} & \textbf{Rubric} \\
\midrule
D1 & Verifiability & Score 1--5 on whether clinical claims can be traced to specific published evidence. Does the output cite specific PMIDs, studies, or guidelines? Can a clinician verify each claim within 1 minute? Score 5 if most claims have verifiable citations; 1 if no citations. \\
D2 & Actionability & Score 1--5 on whether the output provides actionable clinical guidance. Does it specify concrete next steps, dosing, monitoring, or referral criteria? Score 5 if immediately actionable; 1 if only general information. \\
D3 & Temporal Precision & Score 1--5 on whether the output provides specific time points for disease milestones, treatment response windows, and monitoring intervals. Score 5 if precise temporal anchoring; 1 if only vague temporal language. \\
D4 & Non-Expert Safety & Score 1--5 on whether the output is safe for a non-specialist (e.g., GP) to act on. Are red flags clearly flagged? Are dangerous diagnostic pitfalls highlighted? Score 5 if safe for non-expert use; 1 if serious safety risks without specialist oversight. \\
D5 & Clinical Completeness & Score 1--5 on whether the output covers ALL clinically important features for this scenario: key differentiating features, relevant investigations, treatment options, prognosis, and red flags. \\
\bottomrule
\end{tabular}
\end{table}

\subsection*{S6.2: v1 Blind Judge Prompt Template}

\begin{lstlisting}[style=systemprompt]
You are an expert clinical evaluator (board-certified neurologist with 15+ years
of experience in neuromuscular diseases). You are evaluating an AI-generated
clinical output for a rare neuromuscular disease scenario.

## Clinical Scenario
{scenario_text}

## AI-Generated Output
{output_text}

## Evaluation Task
Rate this output on EACH of the following 5 dimensions using a 1-5 Likert scale:
  D1 (Verifiability): [rubric]
  D2 (Actionability): [rubric]
  D3 (Temporal Precision): [rubric]
  D4 (Non-Expert Safety): [rubric]
  D5 (Clinical Completeness): [rubric]

## Response Format
Respond ONLY with a JSON object (no markdown, no explanation):
{
  "D1_verifiability": <1-5>,
  "D2_actionability": <1-5>,
  "D3_temporal_precision": <1-5>,
  "D4_nonexpert_safety": <1-5>,
  "D5_clinical_completeness": <1-5>,
  "brief_justification": "<2-3 sentences explaining your overall assessment>"
}
\end{lstlisting}

\subsection*{S6.3: v2 Citation-Aware Judge Prompt Template}

The v2 prompt is identical to v1 with two additions:

\begin{enumerate}
\item \textbf{Citation Audit Report} appended after the AI-generated output
\item \textbf{D1-specific instruction} added before the dimensions:
\end{enumerate}

\begin{lstlisting}[style=systemprompt]
IMPORTANT: For D1 (Verifiability), use the Citation Audit Report above
as ground truth. If the audit shows 0 PMIDs or mostly wrong-field citations,
D1 should be LOW (1-2). If the audit shows most PMIDs are real and relevant,
D1 should be HIGH (4-5).
\end{lstlisting}

\subsection*{S6.4: Citation Audit Report Format (Appended in v2)}

For outputs with PMIDs:
\begin{lstlisting}[style=systemprompt]
## Citation Audit Report
This output cites **N unique PMIDs**. We verified each against the PubMed database:
- **X** (Y%) are real publications relevant to this clinical domain
- **X** (Y%) are real PubMed records but in UNRELATED medical fields
- **X** (Y%) do not exist in PubMed

  PMID:12345678 [RELEVANT] -- "Title..." (Journal)
  PMID:87654321 [WRONG FIELD] -- "Title..." (Journal)
  ... and N more PMIDs verified
\end{lstlisting}

For outputs without PMIDs:
\begin{lstlisting}[style=systemprompt]
## Citation Audit Report
This output contains **0 PubMed identifiers (PMIDs)**. No specific citations
can be verified against PubMed. All clinical claims rely on unverifiable
parametric knowledge.
\end{lstlisting}

\subsection*{S6.5: Judge Panel Configuration}

\begin{table}[htbp]
\centering
\begin{tabular}{llcc}
\toprule
\textbf{Judge Model} & \textbf{Provider} & \textbf{Temperature} & \textbf{Max Tokens} \\
\midrule
GPT-4o-mini\cite{openai2023gpt4} & OpenAI & 0.0 & 2000 \\
DeepSeek-V3 (deepseek-chat)\cite{deepseekai2024v3} & DeepSeek & 0.0 & 2000 \\
Claude 3 Haiku\cite{anthropic2024claude} & Anthropic & 0.0 & 2000 \\
\bottomrule
\end{tabular}
\end{table}

All three judges are primary panel members, selected for provider diversity (OpenAI, DeepSeek, Anthropic), cost efficiency, and fast inference. Each judge evaluates all 108 cases (36 scenarios $\times$ 3 arms) in both v1 and v2 rounds, totaling 648 evaluations.

\subsection*{S6.6: Case Design}

The 36 base scenarios were evaluated across all 3 arms (Vanilla, Guideline-RAG, HEG-TKG), yielding 108 cases total (36 per arm). Each case was assigned a blinded random ID and shuffled using a deterministic seed (seed=42) to prevent order effects. No demographic or severity expansion was applied; each case corresponds to exactly one base scenario and one system arm, avoiding pseudo-replication. The evaluation key mapping blinded IDs to systems and scenarios is stored per disease pair.

\section*{S8: Clinician Evaluator Instructions and Scoring Rubric}
\addcontentsline{toc}{section}{S8: Clinician Evaluator Instructions and Scoring Rubric}

\subsection*{Study Design}

Single-blind independent-cases evaluation. Each evaluator reviews 24 clinical cases per disease pair. Each case presents a clinical scenario followed by a single AI-generated response. Evaluators are fully blinded to which system produced each output.

\subsection*{Scoring Guide (1--5 Likert)}

\begin{itemize}
\item \textbf{1} = Strongly Disagree (major deficiencies)
\item \textbf{2} = Disagree (notable issues)
\item \textbf{3} = Neutral (adequate but not outstanding)
\item \textbf{4} = Agree (good quality)
\item \textbf{5} = Strongly Agree (excellent)
\end{itemize}

\subsection*{Evaluation Dimensions}

\begin{table}[htbp]
\centering
\begin{tabular}{clp{9cm}}
\toprule
\textbf{ID} & \textbf{Dimension} & \textbf{What to Assess} \\
\midrule
D1 & Verifiability & Can each clinical claim be traced to a specific, identifiable published source? 1~=~No claims traceable; 3~=~About half sourced; 5~=~Every claim traceable. \\
D2 & Actionability & Is the information sufficient and well-supported to inform a clinical decision without additional literature search? 1~=~Requires complete verification; 3~=~Partially actionable; 5~=~Fully actionable. \\
D3 & Temporal Precision & Does the output use specific time windows for disease onset, progression milestones, and clinical trajectories? Are temporal claims anchored to evidence? 1~=~No temporal context; 3~=~Vague or inconsistent timelines; 5~=~Explicit time windows with cited milestones. \\
D4 & Non-Expert Safety & How safe is this output if a non-specialist GP followed it without additional expert consultation? 1~=~Could lead to patient harm; 3~=~Mostly safe; 5~=~Fully safe with caveats and referral triggers. \\
D5 & Clinical Completeness & Does the output cover key differential diagnoses, therapeutic options, contraindications, and monitoring? 1~=~Dangerously incomplete; 3~=~Moderately complete; 5~=~Exhaustive. \\
\bottomrule
\end{tabular}
\end{table}

\subsection*{Citation Verification Task}

For each case, evaluators:
\begin{enumerate}
\item Select one factual claim from the AI output
\item Attempt to verify it against PubMed
\item Record: the claim text, verification result, time to verify (seconds), and whether a specific supporting paper was found
\end{enumerate}

\subsection*{Time Estimate}

Approximately 2--3 hours total (24 cases at ${\sim}2$ minutes each + global assessment).

\clearpage
\section*{S9: Per-Scenario Automated Metrics}
\addcontentsline{toc}{section}{S9: Per-Scenario Automated Metrics}

Evidence Traceability Score (ETS), Clinical Feature Coverage (FC), and Provenance Gap (PG) for each of the 36 scenarios across all three arms. PG = max(FC $-$ ETS $\times$ $r$, 0) where $r$ = 0.97 (HEG-TKG), 0.80 (Vanilla), 0.50 (Guideline-RAG).

\begin{landscape}
{\scriptsize
\begin{longtable}{llrrr|rrr|rrr}
\toprule
\textbf{Scenario} & \textbf{Type} & \multicolumn{3}{c|}{\textbf{Vanilla}} & \multicolumn{3}{c|}{\textbf{Guideline-RAG}} & \multicolumn{3}{c}{\textbf{HEG-TKG}} \\
& & FC & ETS & PG & FC & ETS & PG & FC & ETS & PG \\
\midrule
\endfirsthead
\toprule
\textbf{Scenario} & \textbf{Type} & \multicolumn{3}{c|}{\textbf{Vanilla}} & \multicolumn{3}{c|}{\textbf{Guideline-RAG}} & \multicolumn{3}{c}{\textbf{HEG-TKG}} \\
& & FC & ETS & PG & FC & ETS & PG & FC & ETS & PG \\
\midrule
\endhead
CIDP\_GBS\_DDX\_01 & diff. & 0.75 & 0.00 & 0.75 & 0.38 & 0.00 & 0.38 & 0.50 & 0.41 & 0.11 \\
CIDP\_GBS\_DDX\_02 & diff. & 0.44 & 0.00 & 0.44 & 0.89 & 0.00 & 0.89 & 0.67 & 0.36 & 0.32 \\
CIDP\_GBS\_DDX\_03 & diff. & 1.00 & 0.00 & 1.00 & 1.00 & 0.00 & 1.00 & 0.89 & 0.47 & 0.43 \\
CIDP\_GBS\_DDX\_04 & diff. & 0.50 & 0.00 & 0.50 & 0.50 & 0.00 & 0.50 & 0.38 & 0.28 & 0.11 \\
CIDP\_GBS\_SAFETY\_01 & diff. & 0.62 & 0.00 & 0.62 & 0.62 & 0.00 & 0.62 & 0.62 & 0.41 & 0.23 \\
CIDP\_GBS\_TEMP\_01 & temp. comp. & 0.71 & 0.00 & 0.71 & 0.57 & 0.00 & 0.57 & 0.71 & 0.31 & 0.41 \\
CIDP\_GBS\_TEMP\_02 & temp. comp. & 0.67 & 0.00 & 0.67 & 0.67 & 0.00 & 0.67 & 0.67 & 0.24 & 0.43 \\
CIDP\_GBS\_TEMP\_03 & temporal & 0.86 & 0.00 & 0.86 & 0.86 & 0.00 & 0.86 & 0.86 & 0.12 & 0.74 \\
CIDP\_GBS\_TEMP\_04 & temporal & 0.88 & 0.00 & 0.88 & 0.62 & 0.00 & 0.62 & 0.62 & 0.07 & 0.56 \\
CIDP\_GBS\_TX\_01 & treatment & 0.71 & 0.00 & 0.71 & 0.57 & 0.00 & 0.57 & 0.71 & 0.43 & 0.29 \\
CIDP\_GBS\_TX\_02 & treatment & 0.75 & 0.00 & 0.75 & 0.62 & 0.00 & 0.62 & 0.75 & 0.73 & 0.04 \\
CIDP\_GBS\_TX\_03 & treatment & 0.62 & 0.00 & 0.62 & 0.38 & 0.00 & 0.38 & 0.75 & 0.46 & 0.30 \\
\midrule
DMD\_BMD\_DDX\_01 & diff. & 0.80 & 0.00 & 0.80 & 0.70 & 0.00 & 0.70 & 0.90 & 0.48 & 0.43 \\
DMD\_BMD\_DDX\_02 & diff. & 0.86 & 0.00 & 0.86 & 0.86 & 0.00 & 0.86 & 0.71 & 0.32 & 0.41 \\
DMD\_BMD\_DDX\_03 & diff. & 0.75 & 0.00 & 0.75 & 0.75 & 0.00 & 0.75 & 0.38 & 0.45 & 0.00 \\
DMD\_BMD\_DDX\_04 & diff. & 0.50 & 0.00 & 0.50 & 0.62 & 0.00 & 0.62 & 0.62 & 0.42 & 0.21 \\
DMD\_BMD\_SAFETY\_01 & diff. & 0.75 & 0.00 & 0.75 & 0.75 & 0.00 & 0.75 & 0.75 & 0.36 & 0.40 \\
DMD\_BMD\_TEMP\_01 & temp. comp. & 1.00 & 0.00 & 1.00 & 0.67 & 0.00 & 0.67 & 1.00 & 0.20 & 0.80 \\
DMD\_BMD\_TEMP\_02 & temp. comp. & 1.00 & 0.00 & 1.00 & 0.71 & 0.00 & 0.71 & 0.86 & 0.46 & 0.41 \\
DMD\_BMD\_TEMP\_03 & temporal & 0.75 & 0.00 & 0.75 & 0.62 & 0.00 & 0.62 & 0.62 & 0.30 & 0.33 \\
DMD\_BMD\_TEMP\_04 & temporal & 0.43 & 0.00 & 0.43 & 0.57 & 0.00 & 0.57 & 0.29 & 0.23 & 0.06 \\
DMD\_BMD\_TX\_01 & treatment & 0.67 & 0.00 & 0.67 & 0.78 & 0.00 & 0.78 & 0.78 & 0.50 & 0.29 \\
DMD\_BMD\_TX\_02 & treatment & 0.88 & 0.00 & 0.88 & 0.75 & 0.00 & 0.75 & 0.62 & 0.93 & 0.00 \\
DMD\_BMD\_TX\_03 & treatment & 0.88 & 0.00 & 0.88 & 0.50 & 0.00 & 0.50 & 0.38 & 0.27 & 0.12 \\
\midrule
MG\_LEMS\_DDX\_01 & diff. & 0.60 & 0.00 & 0.60 & 0.90 & 0.00 & 0.90 & 1.00 & 0.32 & 0.69 \\
MG\_LEMS\_DDX\_02 & diff. & 1.00 & 0.00 & 1.00 & 0.90 & 0.00 & 0.90 & 1.00 & 0.22 & 0.78 \\
MG\_LEMS\_DDX\_03 & diff. & 0.62 & 0.00 & 0.62 & 0.75 & 0.00 & 0.75 & 0.62 & 0.27 & 0.36 \\
MG\_LEMS\_DDX\_04 & diff. & 0.70 & 0.00 & 0.70 & 0.70 & 0.00 & 0.70 & 0.60 & 0.15 & 0.45 \\
MG\_LEMS\_SAFETY\_01 & diff. & 0.50 & 0.00 & 0.50 & 0.75 & 0.00 & 0.75 & 0.88 & 0.25 & 0.63 \\
MG\_LEMS\_TEMP\_01 & temp. comp. & 1.00 & 0.00 & 1.00 & 0.88 & 0.00 & 0.88 & 0.88 & 0.26 & 0.62 \\
MG\_LEMS\_TEMP\_02 & temp. comp. & 0.86 & 0.00 & 0.86 & 0.86 & 0.00 & 0.86 & 0.71 & 0.35 & 0.37 \\
MG\_LEMS\_TEMP\_03 & temporal & 0.86 & 0.00 & 0.86 & 0.71 & 0.00 & 0.71 & 1.00 & 0.43 & 0.58 \\
MG\_LEMS\_TEMP\_04 & temporal & 0.43 & 0.00 & 0.43 & 0.43 & 0.00 & 0.43 & 0.43 & 0.00 & 0.43 \\
MG\_LEMS\_TX\_01 & treatment & 0.88 & 0.00 & 0.88 & 0.88 & 0.00 & 0.88 & 0.88 & 0.41 & 0.48 \\
MG\_LEMS\_TX\_02 & treatment & 0.86 & 0.00 & 0.86 & 0.29 & 0.00 & 0.29 & 1.00 & 0.83 & 0.20 \\
MG\_LEMS\_TX\_03 & treatment & 0.67 & 0.00 & 0.67 & 0.78 & 0.00 & 0.78 & 0.78 & 0.69 & 0.11 \\
\midrule
\textbf{Mean $\pm$ SD} & & \textbf{0.74} & \textbf{0.00} & \textbf{0.74} & \textbf{0.69} & \textbf{0.00} & \textbf{0.69} & \textbf{0.72} & \textbf{0.37} & \textbf{0.37} \\
& & $\pm$0.17 & & $\pm$0.17 & $\pm$0.17 & & $\pm$0.17 & $\pm$0.19 & $\pm$0.19 & $\pm$0.22 \\
\bottomrule
\end{longtable}
}
\end{landscape}

\section*{S10: Entity Normalization - 18 Semantic Correction Rules}
\addcontentsline{toc}{section}{S10: Entity Normalization - 18 Semantic Correction Rules}

The following rules are applied sequentially during entity normalization (Step~3 of the extraction pipeline). Each rule matches a specific subject\_type $\rightarrow$ predicate $\rightarrow$ object\_type pattern and transforms the triplet to correct common LLM extraction errors.

\begin{landscape}
{\scriptsize
\begin{longtable}{c l l l l}
\toprule
\textbf{\#} & \textbf{Rule Name} & \textbf{Input Pattern} & \textbf{Corrected Output} & \textbf{Purpose} \\
\midrule
\endfirsthead
\toprule
\textbf{\#} & \textbf{Rule Name} & \textbf{Input Pattern} & \textbf{Corrected Output} & \textbf{Purpose} \\
\midrule
\endhead
\bottomrule
\endfoot
1 & invert\_caused\_by\_mutation & Gene $\to$ CAUSED\_BY\_MUTATION $\to$ Disease & Disease $\to$ CAUSED\_BY\_MUTATION $\to$ Gene & Fix direction inversion \\
2 & invert\_treated\_with & Treatment $\to$ TREATED\_WITH $\to$ Disease & Disease $\to$ TREATED\_WITH $\to$ Treatment & Fix direction inversion \\
3 & procedure\_to\_monitored & Procedure $\to$ TREATED\_WITH $\to$ Disease & Disease $\to$ MONITORED\_WITH $\to$ Procedure & Diagnostics $\neq$ treatments \\
3b & therapeutic\_proc\_invert & Procedure $\to$ TREATED\_WITH $\to$ Disease & Disease $\to$ TREATED\_WITH $\to$ Procedure & Therapeutic: invert only \\
4 & mutation\_occurs\_in & Mutation $\to$ ASSOCIATED\_WITH $\to$ Gene & Mutation $\to$ OCCURS\_IN $\to$ Gene & Refine generic predicate \\
5 & inheritance\_pattern & Disease $\to$ ASSOCIATED\_WITH $\to$ Entity & Disease $\to$ HAS\_INHERITANCE $\to$ InheritPattern & Object has ``x-linked'' etc. \\
6 & entity\_treatment\_invert & Entity $\to$ TREATED\_WITH $\to$ Disease & Disease $\to$ TREATED\_WITH $\to$ Treatment & Direction + retype \\
7 & measurement\_to\_monitored & Measurement $\to$ TREATED\_WITH $\to$ Disease & Disease $\to$ MONITORED\_WITH $\to$ Measurement & Measurements $\neq$ treatments \\
8 & onset\_to\_symptom\_onset & PatientGrp $\to$ HAS\_ONSET\_AGE $\to$ Symptom & PatientGrp $\to$ SYMPTOM\_ONSET\_AT $\to$ Symptom & Refine temporal predicate \\
9 & develops\_to\_loses\_func & PatientGrp $\to$ DEVELOPS\_COMPL $\to$ PhysFunc & PatientGrp $\to$ LOSES\_FUNCTION\_AT $\to$ PhysFunc & Evidence mentions ``loss'' \\
10 & retype\_entity\_assoc & Entity $\to$ ASSOCIATED\_WITH $\to$ Disease & Procedure $\to$ USED\_FOR\_DIAG $\to$ Disease & Subject is ``biopsy'' etc. \\
11 & patgrp\_proc\_to\_monitored & PatientGrp $\to$ TREATED\_WITH $\to$ Procedure & PatientGrp $\to$ MONITORED\_WITH $\to$ Procedure & Diagnostic for patients \\
12 & proc\_assoc\_to\_diagnosis & Procedure $\to$ ASSOCIATED\_WITH $\to$ Disease & Procedure $\to$ USED\_FOR\_DIAG $\to$ Disease & Refine generic association \\
13 & caused\_by\_protein\_state & Disease $\to$ CAUSED\_BY\_MUTATION $\to$ Entity & Disease $\to$ LACKS\_FEATURE $\to$ Protein & Object is ``deficiency'' \\
15 & protein\_assoc\_to\_target & Protein $\to$ ASSOCIATED\_WITH $\to$ Disease & Disease $\to$ ASSOCIATED\_WITH $\to$ Protein & Invert receptor--disease \\
16$^*$ & autoantibody\_retype & Disease $\to$ CAUSED\_BY\_MUTATION $\to$ Entity & Disease $\to$ CAUSED\_BY $\to$ Autoantibody & Retype autoantibody \\
14 & retype\_entity\_to\_gene & Disease $\to$ CAUSED\_BY\_MUTATION $\to$ Entity & Disease $\to$ CAUSED\_BY\_MUTATION $\to$ Gene & Retype Entity as Gene \\
17 & measurement\_to\_diagnosis & Measurement $\to$ ASSOCIATED\_WITH $\to$ Disease & Measurement $\to$ USED\_FOR\_DIAG $\to$ Disease & Antibody measurements \\
\end{longtable}
}

\smallskip
$^*$\textbf{Rule ordering:} Rules 16 and 14 both match (Disease, CAUSED\_BY\_MUTATION, Entity). Rule~16 must fire first; otherwise ALL-CAPS autoantibody strings hit Rule~14's uppercase guard and get mistyped as Gene.
\end{landscape}

\clearpage
\section*{S11: Per-Scenario-Type Breakdown}
\addcontentsline{toc}{section}{S11: Per-Scenario-Type Breakdown}

Performance stratified by the four scenario types. The Provenance Gap reduction varies by scenario type, with HEG-TKG achieving the strongest ETS on treatment scenarios (0.58) where structured citation of drug interactions and dosing evidence is most clinically relevant.

\begin{table}[htbp]
\centering
\caption*{\textbf{Supplementary Table S11a. Automated metrics by scenario type (mean across scenarios).}}
\small
\begin{tabular}{lc|rrr|rrr|rrr}
\toprule
\textbf{Type} & $n$ & \multicolumn{3}{c|}{\textbf{Vanilla}} & \multicolumn{3}{c|}{\textbf{G-RAG}} & \multicolumn{3}{c}{\textbf{HEG-TKG}} \\
& & FC & ETS & PG & FC & ETS & PG & FC & ETS & PG \\
\midrule
Differential Dx    & 15 & 0.69 & 0.00 & 0.69 & 0.74 & 0.00 & 0.74 & 0.70 & 0.35 & 0.37 \\
Temporal Comp.     &  6 & 0.87 & 0.00 & 0.87 & 0.73 & 0.00 & 0.73 & 0.80 & 0.30 & 0.51 \\
Temporal Traj.     &  6 & 0.70 & 0.00 & 0.70 & 0.64 & 0.00 & 0.64 & 0.64 & 0.19 & 0.45 \\
Treatment          &  9 & 0.77 & 0.00 & 0.77 & 0.62 & 0.00 & 0.62 & 0.74 & 0.58 & 0.20 \\
\bottomrule
\end{tabular}
\end{table}

Treatment scenarios show the lowest Provenance Gap for HEG-TKG (PG = 0.20), reflecting that treatment-related knowledge graph edges carry high inline citation rates (ETS = 0.58). Temporal trajectory scenarios have lower ETS (0.19) because temporal milestone claims are synthesised from multiple KG edges and the model often omits inline PMID tags when combining temporal facts into narrative text; however, these claims remain fully traceable via the stored evidence manifest.

\begin{table}[htbp]
\centering
\caption*{\textbf{Supplementary Table S11b. LLM judge v2 (citation-aware) D1 and D3 by scenario type.}}
\small
\begin{tabular}{l|ccc|ccc}
\toprule
\textbf{Type} & \multicolumn{3}{c|}{\textbf{D1 Verifiability}} & \multicolumn{3}{c}{\textbf{D3 Temporal Precision}} \\
& Van. & G-RAG & HEG-TKG & Van. & G-RAG & HEG-TKG \\
\midrule
Differential Dx    & 1.40 & 1.44 & 4.89 & 3.91 & 3.38 & 4.51 \\
Temporal Comp.     & 1.33 & 1.39 & 4.83 & 4.94 & 4.61 & 5.00 \\
Temporal Traj.     & 1.33 & 1.33 & 3.78 & 4.89 & 4.72 & 5.00 \\
Treatment          & 1.41 & 1.48 & 4.70 & 3.96 & 2.67 & 4.33 \\
\bottomrule
\end{tabular}
\end{table}

D1 Verifiability shows consistent HEG-TKG advantage across all scenario types ($\Delta > 2.3$ in every category). D3 Temporal Precision advantages are largest for treatment scenarios ($\Delta = +0.37$ vs Vanilla, $+1.66$ vs Guideline-RAG), where Guideline-RAG's lack of structured temporal data is most apparent (D3 = 2.67).

\section*{S12: Regulatory Compliance Mapping}
\addcontentsline{toc}{section}{S12: Regulatory Compliance Mapping}

\textit{Moved from main text Table~6 to supplementary.}

\textbf{Supplementary Table S26. Regulatory compliance assessment across three system architectures.}

\begin{table}[htbp]
\centering
\small
\begin{tabular}{p{4.5cm}p{3cm}p{3cm}p{3cm}}
\toprule
\textbf{Requirement} & \textbf{Vanilla LLM} & \textbf{Guideline-RAG} & \textbf{HEG-TKG} \\
\midrule
\textbf{EU AI Act Art.~14}: Human oversight via interpretable outputs & Partial: text output only & Partial: source passages & \textbf{Full}: PMID-linked claims \\
\addlinespace
\textbf{EU AI Act Art.~14}: Ability to understand and challenge outputs & No source tracking & Text chunks, no structured provenance & \textbf{PMID $\rightarrow$ PubMed $\rightarrow$ full paper} \\
\addlinespace
\textbf{FDA CDS Criterion~4}: Transparency of information sources & Fails: fabricated or absent citations & Partial: raw text retrieval & \textbf{Full}: every claim traceable \\
\addlinespace
\textbf{FDA CDS}: Intended user can independently review basis & No: unverifiable & Partial: text search required & \textbf{Yes}: PMID lookup in ${\sim}10$s \\
\addlinespace
\textbf{GDPR Art.~9 / HIPAA}: PHI protection & No: cloud API required & No: cloud API required & \textbf{Yes}: local deployment option \\
\bottomrule
\end{tabular}
\end{table}

\section*{S13: Evidence Quality Tier Analysis}
\addcontentsline{toc}{section}{S13: Evidence Quality Tier Analysis}

Across all 36 scenarios, HEG-TKG retrieved 26,139 knowledge graph edges. Table~\ref{tab:s13_quality} shows the quality tier distribution.

\begin{table}[ht]
\centering
\caption{\textbf{Evidence quality tier distribution across 26,139 retrieved edges.}}
\label{tab:s13_quality}
\small
\begin{tabular}{lrrp{7cm}}
\toprule
\textbf{Tier} & \textbf{Count} & \textbf{\%} & \textbf{Definition} \\
\midrule
GOLD   & 562    & 2.2\%  & Cross-tier confirmed: Tier~2 extraction independently confirms a Tier~1 curated fact (confidence = 0.95) \\
SILVER & 2,271  & 8.7\%  & Multi-model or multi-document consensus ($\geq$2 extraction models agree, or $\geq$2 independent PMIDs; confidence = 0.85) \\
BRONZE & 23,306 & 89.2\% & Single model, single source (confidence = 0.70) \\
\midrule
\textbf{Total} & \textbf{26,139} & \textbf{100\%} & \\
\bottomrule
\end{tabular}
\end{table}

Temporal edges account for 5,802 (22.2\%) of all retrieved evidence. GOLD edges are not preferentially retrieved for safety-critical or treatment scenarios (2.4\% of edges in treatment scenarios vs 2.2\% overall), reflecting that the quality tier distribution is driven by extraction consensus patterns rather than clinical content type. The clinical implication: GOLD edges provide high-confidence anchors that a clinician can trust with minimal verification, SILVER edges warrant standard verification, and BRONZE edges flag claims that rest on a single source and should be scrutinised before acting on.

\section*{S14: Citation Density Analysis}
\addcontentsline{toc}{section}{S14: Citation Density Analysis}

Inline PMID citation density across all 36 scenarios and three arms.

\begin{table}[htbp]
\centering
\caption*{\textbf{Supplementary Table S14. Citation density analysis.} Mean values across 12 scenarios per disease pair.}
\small
\begin{tabular}{llrrrr}
\toprule
\textbf{System} & \textbf{Pair} & \textbf{Words} & \textbf{PMIDs} & \textbf{PMIDs/1k words} \\
\midrule
\multirow{4}{*}{Vanilla}
 & MG/LEMS  & 861  & 0.0 & 0.0 \\
 & DMD/BMD  & 828  & 0.0 & 0.0 \\
 & CIDP/GBS & 793  & 0.0 & 0.0 \\
 & \textbf{All} & \textbf{827}  & \textbf{0.0} & \textbf{0.0} \\
\midrule
\multirow{4}{*}{Guideline-RAG}
 & MG/LEMS  & 853  & 0.0 & 0.0 \\
 & DMD/BMD  & 876  & 0.0 & 0.0 \\
 & CIDP/GBS & 869  & 0.0 & 0.0 \\
 & \textbf{All} & \textbf{866}  & \textbf{0.0} & \textbf{0.0} \\
\midrule
\multirow{4}{*}{HEG-TKG}
 & MG/LEMS  & 1,313 & 21.6 & 16.2 \\
 & DMD/BMD  & 1,245 & 37.8 & 30.3 \\
 & CIDP/GBS & 1,178 & 29.8 & 24.8 \\
 & \textbf{All} & \textbf{1,246} & \textbf{29.7} & \textbf{23.8} \\
\bottomrule
\end{tabular}
\end{table}

Neither Vanilla nor Guideline-RAG produces any PMID citations in their outputs, consistent with the zero ETS scores reported in the main text. HEG-TKG averages 23.8 inline PMID citations per 1,000 words, with DMD/BMD showing the highest density (30.3) reflecting the denser temporal evidence available for this disease pair. HEG-TKG outputs are approximately 50\% longer than baseline outputs (1,246 vs 827--866 words), attributable to the structured evidence blocks and inline citations. The 203 unique PMIDs cited across all 36 HEG-TKG outputs represent 21\% of the 987 unique PMIDs in the knowledge graphs, indicating that subgraph retrieval selects a focused, query-relevant subset of the evidence base.

\section*{S15: YAML Configuration Schema for Disease Pairs}
\addcontentsline{toc}{section}{S15: YAML Configuration Schema for Disease Pairs}

Each disease pair is defined by a YAML configuration file. Adding a new disease pair requires only creating a new YAML file; no code changes are needed. Below is the schema (using DMD/BMD as an example):

\begin{lstlisting}[style=systemprompt]
# Disease pair metadata
disease_pair: dmd_bmd
label: "Duchenne vs Becker Muscular Dystrophy"

# Classification section -- used by DiseaseClassifier to assign disease context
classification:
  diseases:
    - short_name: DMD
      full_name: Duchenne muscular dystrophy
      cuis:
        - C0013264  # Primary UMLS CUI
        - C0410189  # Alternate CUI
      text_patterns:
        - duchenne
        - dmd
        - "duchenne muscular dystrophy"
      ontology_id: "OMIM:310200"
      mondo_id: "MONDO:0007803"

    - short_name: BMD
      full_name: Becker muscular dystrophy
      cuis:
        - C0917713
      text_patterns:
        - becker
        - bmd
        - "becker muscular dystrophy"
      ontology_id: "OMIM:300376"
      mondo_id: "MONDO:0010311"

  shared:
    cuis:
      - C0079259  # Dystrophin
      - C1457887  # DMD gene
    text_patterns:
      - dystrophin
      - dystrophinopathy
      - "exon skipping"
    parent:
      name: Dystrophinopathy
      cui: C0872247

# PubMed retrieval parameters
pubmed:
  min_year: 2015
  max_year: 2025
  max_abstracts_per_query: 100
  mesh_terms:
    - '"Muscular Dystrophy, Duchenne"[Mesh]'
  differential_terms:
    - "differential diagnosis"
    - "genotype-phenotype"
  progression_terms:
    - "natural history"
    - "disease progression"
  anchor_categories:
    temporal: ["age of onset", "loss of ambulation"]
    cardiac: ["cardiomyopathy", "ejection fraction"]
    treatment: ["corticosteroid", "exon skipping"]

# Temporal predicates requiring numerical guardrails
temporal_predicates:
  - HAS_ONSET_AGE
  - HAS_DIAGNOSIS_AGE
  - GENERALIZED_AT
  - DEVELOPS_COMPLICATION_AT
  - HAS_SURVIVAL_TO
  - REQUIRES_INTERVENTION_AT
  # ... (12 total temporal relationship types)

# Extraction section -- LLM prompt context and few-shot examples
extraction:
  prompt_context: "Focus on dystrophinopathies..."
  screening_keywords: [...]
  few_shot_examples: [...]
\end{lstlisting}

\section*{S16: Prior Conference-Stage Work Comparison}
\addcontentsline{toc}{section}{S16: Prior Conference-Stage Work Comparison}

\begin{table}[htbp]
\centering
\small
\begin{tabular}{p{3.5cm}p{5cm}p{5cm}}
\toprule
\textbf{Aspect} & \textbf{Conference Investigation} & \textbf{This Study} \\
\midrule
Disease pairs & 1 (DMD/BMD only) & 3 (MG/LEMS, DMD/BMD, CIDP/GBS) \\
Clinical scenarios & 12 & 36 \\
Evaluation design & Paired A/B comparison & Independent blinded cases \\
LLM models tested & 1 (GPT-4o) & 5 frontier models \\
Citation verification & Manual spot-check & Automated PubMed E-utilities (all PMIDs) \\
LLM judge evaluation & Not performed & 648 evaluations ($108 \times 3$ judges $\times$ 2 rounds) \\
Citation-aware judging & Not performed & v1 vs v2 comparison (methodological contribution) \\
Counterfactual testing & Not performed & 15 injected errors \\
Statistical analysis & Descriptive & Mann-Whitney $U$, Cohen's \textit{d}, bootstrap CIs, BH FDR, Krippendorff's $\alpha$ \\
PHI compliance & Not addressed & Validated local deployment with open-source models \\
\bottomrule
\end{tabular}
\end{table}

\section*{S18: v1 (Blind) LLM Judge Results - MG/LEMS (All 5 Dimensions, 3 Arms)}
\addcontentsline{toc}{section}{S18: v1 (Blind) LLM Judge Results - MG/LEMS}

\textbf{Supplementary Table S18.} v1 blind evaluation, MG/LEMS ($N$=36 per arm, 3 judges each). HEG-TKG vs Vanilla deltas shown.

\begin{table}[htbp]
\centering
\small
\begin{tabular}{lccccc}
\toprule
\textbf{Dimension} & \textbf{Vanilla} & \textbf{G-RAG} & \textbf{HEG-TKG} & \textbf{$\Delta$(H-V)} & \textbf{$d$} \\
\midrule
\textbf{D1 Verifiability} & $\mathbf{4.42 \pm 0.69}$ & $\mathbf{4.53 \pm 0.56}$ & $\mathbf{4.50 \pm 0.56}$ & \textbf{+0.08} & \textbf{0.13} \\
D2 Actionability & $4.47 \pm 0.51$ & $4.31 \pm 0.47$ & $4.50 \pm 0.51$ & +0.03 & 0.05 \\
D3 Temporal Precision & $4.33 \pm 0.83$ & $3.47 \pm 1.25$ & $4.25 \pm 0.77$ & $-$0.08 & $-$0.10 \\
D4 Non-Expert Safety & $4.03 \pm 0.51$ & $3.94 \pm 0.47$ & $4.11 \pm 0.52$ & +0.08 & 0.16 \\
D5 Clinical Completeness & $4.50 \pm 0.61$ & $4.17 \pm 0.65$ & $4.61 \pm 0.49$ & +0.11 & 0.20 \\
\bottomrule
\end{tabular}
\end{table}

\section*{S19: v1 (Blind) LLM Judge Results - DMD/BMD (All 5 Dimensions, 3 Arms)}
\addcontentsline{toc}{section}{S19: v1 (Blind) LLM Judge Results - DMD/BMD}

\textbf{Supplementary Table S19.} v1 blind evaluation, DMD/BMD ($N$=36 per arm, 3 judges each).

\begin{table}[htbp]
\centering
\small
\begin{tabular}{lccccc}
\toprule
\textbf{Dimension} & \textbf{Vanilla} & \textbf{G-RAG} & \textbf{HEG-TKG} & \textbf{$\Delta$(H-V)} & \textbf{$d$} \\
\midrule
\textbf{D1 Verifiability} & $\mathbf{4.25 \pm 0.94}$ & $\mathbf{4.67 \pm 0.53}$ & $\mathbf{4.75 \pm 0.44}$ & \textbf{+0.50} & \textbf{0.68} \\
D2 Actionability & $4.44 \pm 0.50$ & $4.31 \pm 0.52$ & $4.61 \pm 0.49$ & +0.17 & 0.33 \\
D3 Temporal Precision & $4.47 \pm 0.65$ & $4.14 \pm 0.80$ & $4.75 \pm 0.44$ & +0.28 & 0.50 \\
D4 Non-Expert Safety & $4.14 \pm 0.49$ & $4.08 \pm 0.37$ & $4.19 \pm 0.47$ & +0.06 & 0.12 \\
D5 Clinical Completeness & $4.33 \pm 0.68$ & $4.17 \pm 0.77$ & $4.64 \pm 0.49$ & +0.31 & 0.52 \\
\bottomrule
\end{tabular}
\end{table}

\section*{S20: v1 (Blind) LLM Judge Results - CIDP/GBS (All 5 Dimensions, 3 Arms)}
\addcontentsline{toc}{section}{S20: v1 (Blind) LLM Judge Results - CIDP/GBS}

\textbf{Supplementary Table S20.} v1 blind evaluation, CIDP/GBS ($N$=36 per arm, 3 judges each).

\begin{table}[htbp]
\centering
\small
\begin{tabular}{lccccc}
\toprule
\textbf{Dimension} & \textbf{Vanilla} & \textbf{G-RAG} & \textbf{HEG-TKG} & \textbf{$\Delta$(H-V)} & \textbf{$d$} \\
\midrule
\textbf{D1 Verifiability} & $\mathbf{4.36 \pm 0.72}$ & $\mathbf{4.56 \pm 0.56}$ & $\mathbf{4.53 \pm 0.65}$ & \textbf{+0.17} & \textbf{0.24} \\
D2 Actionability & $4.44 \pm 0.50$ & $4.22 \pm 0.48$ & $4.47 \pm 0.51$ & +0.03 & 0.05 \\
D3 Temporal Precision & $4.64 \pm 0.54$ & $4.28 \pm 0.97$ & $4.69 \pm 0.47$ & +0.06 & 0.11 \\
D4 Non-Expert Safety & $4.06 \pm 0.41$ & $4.06 \pm 0.33$ & $4.08 \pm 0.44$ & +0.03 & 0.07 \\
D5 Clinical Completeness & $4.39 \pm 0.69$ & $4.28 \pm 0.70$ & $4.64 \pm 0.54$ & +0.25 & 0.40 \\
\bottomrule
\end{tabular}
\end{table}

\section*{S21: v2 (Citation-Aware) LLM Judge Results - MG/LEMS (All 5 Dimensions, 3 Arms)}
\addcontentsline{toc}{section}{S21: v2 (Citation-Aware) LLM Judge Results - MG/LEMS}

\textbf{Supplementary Table S21.} v2 citation-aware evaluation, MG/LEMS ($N$=36 per arm, 3 judges each).

\begin{table}[htbp]
\centering
\small
\begin{tabular}{lccccc}
\toprule
\textbf{Dimension} & \textbf{Vanilla} & \textbf{G-RAG} & \textbf{HEG-TKG} & \textbf{$\Delta$(H-V)} & \textbf{$d$} \\
\midrule
\textbf{D1 Verifiability} & $\mathbf{1.39 \pm 0.49}$ & $\mathbf{1.39 \pm 0.49}$ & $\mathbf{4.44 \pm 1.05}$ & \textbf{+3.06} & \textbf{3.71} \\
D2 Actionability & $4.08 \pm 0.28$ & $3.94 \pm 0.33$ & $4.56 \pm 0.50$ & +0.47 & 1.16 \\
D3 Temporal Precision & $4.06 \pm 0.83$ & $3.11 \pm 1.33$ & $4.33 \pm 0.68$ & +0.28 & 0.37 \\
D4 Non-Expert Safety & $3.81 \pm 0.47$ & $3.75 \pm 0.44$ & $4.06 \pm 0.41$ & +0.25 & 0.57 \\
D5 Clinical Completeness & $4.11 \pm 0.46$ & $3.86 \pm 0.35$ & $4.72 \pm 0.45$ & +0.61 & 1.33 \\
\bottomrule
\end{tabular}
\end{table}

\section*{S22: v2 (Citation-Aware) LLM Judge Results - DMD/BMD (All 5 Dimensions, 3 Arms)}
\addcontentsline{toc}{section}{S22: v2 (Citation-Aware) LLM Judge Results - DMD/BMD}

\textbf{Supplementary Table S22.} v2 citation-aware evaluation, DMD/BMD ($N$=36 per arm, 3 judges each).

\begin{table}[htbp]
\centering
\small
\begin{tabular}{lccccc}
\toprule
\textbf{Dimension} & \textbf{Vanilla} & \textbf{G-RAG} & \textbf{HEG-TKG} & \textbf{$\Delta$(H-V)} & \textbf{$d$} \\
\midrule
\textbf{D1 Verifiability} & $\mathbf{1.39 \pm 0.49}$ & $\mathbf{1.50 \pm 0.51}$ & $\mathbf{4.92 \pm 0.28}$ & \textbf{+3.53} & \textbf{8.78} \\
D2 Actionability & $4.03 \pm 0.17$ & $3.86 \pm 0.42$ & $4.69 \pm 0.47$ & +0.67 & 1.90 \\
D3 Temporal Precision & $4.25 \pm 0.69$ & $3.78 \pm 0.83$ & $4.81 \pm 0.40$ & +0.56 & 0.98 \\
D4 Non-Expert Safety & $3.86 \pm 0.35$ & $3.92 \pm 0.37$ & $4.11 \pm 0.32$ & +0.25 & 0.75 \\
D5 Clinical Completeness & $4.03 \pm 0.38$ & $3.75 \pm 0.55$ & $4.72 \pm 0.45$ & +0.69 & 1.66 \\
\bottomrule
\end{tabular}
\end{table}

\section*{S23: v2 (Citation-Aware) LLM Judge Results - CIDP/GBS (All 5 Dimensions, 3 Arms)}
\addcontentsline{toc}{section}{S23: v2 (Citation-Aware) LLM Judge Results - CIDP/GBS}

\textbf{Supplementary Table S23.} v2 citation-aware evaluation, CIDP/GBS ($N$=36 per arm, 3 judges each).

\begin{table}[htbp]
\centering
\small
\begin{tabular}{lccccc}
\toprule
\textbf{Dimension} & \textbf{Vanilla} & \textbf{G-RAG} & \textbf{HEG-TKG} & \textbf{$\Delta$(H-V)} & \textbf{$d$} \\
\midrule
\textbf{D1 Verifiability} & $\mathbf{1.36 \pm 0.49}$ & $\mathbf{1.39 \pm 0.49}$ & $\mathbf{4.58 \pm 0.77}$ & \textbf{+3.22} & \textbf{5.00} \\
D2 Actionability & $4.03 \pm 0.29$ & $3.89 \pm 0.32$ & $4.56 \pm 0.50$ & +0.53 & 1.28 \\
D3 Temporal Precision & $4.47 \pm 0.51$ & $4.00 \pm 1.12$ & $4.75 \pm 0.44$ & +0.28 & 0.59 \\
D4 Non-Expert Safety & $3.83 \pm 0.38$ & $3.78 \pm 0.42$ & $4.11 \pm 0.32$ & +0.28 & 0.79 \\
D5 Clinical Completeness & $4.11 \pm 0.57$ & $3.86 \pm 0.35$ & $4.75 \pm 0.44$ & +0.64 & 1.25 \\
\bottomrule
\end{tabular}
\end{table}

\section*{S24: Per-Judge D1 Shift Table (v1 $\rightarrow$ v2) Across All Disease Pairs}
\addcontentsline{toc}{section}{S24: Per-Judge D1 Shift Table (v1 to v2)}

\textbf{Supplementary Table S24.} D1 Verifiability scores per judge across all three arms, showing the shift from v1 (blind) to v2 (citation-aware).

\subsection*{MG/LEMS}

\begin{table}[htbp]
\centering
\small
\begin{tabular}{lccccccc}
\toprule
\textbf{Judge} & \textbf{v1 Van} & \textbf{v1 RAG} & \textbf{v1 HEG} & \textbf{v2 Van} & \textbf{v2 RAG} & \textbf{v2 HEG} & \textbf{Van Shift} \\
\midrule
GPT-4o-mini & 4.58 & 4.67 & 4.67 & 1.17 & 1.08 & 4.50 & $-$3.42 \\
DeepSeek-V3 & 3.67 & 3.92 & 4.08 & 1.00 & 1.08 & 4.08 & $-$2.67 \\
Claude 3 Haiku & 5.00 & 5.00 & 4.75 & 2.00 & 2.00 & 4.75 & $-$3.00 \\
\bottomrule
\end{tabular}
\end{table}

\subsection*{DMD/BMD}

\begin{table}[htbp]
\centering
\small
\begin{tabular}{lccccccc}
\toprule
\textbf{Judge} & \textbf{v1 Van} & \textbf{v1 RAG} & \textbf{v1 HEG} & \textbf{v2 Van} & \textbf{v2 RAG} & \textbf{v2 HEG} & \textbf{Van Shift} \\
\midrule
GPT-4o-mini & 4.42 & 5.00 & 4.92 & 1.17 & 1.42 & 4.92 & $-$3.25 \\
DeepSeek-V3 & 3.33 & 4.00 & 4.33 & 1.00 & 1.08 & 4.83 & $-$2.33 \\
Claude 3 Haiku & 5.00 & 5.00 & 5.00 & 2.00 & 2.00 & 5.00 & $-$3.00 \\
\bottomrule
\end{tabular}
\end{table}

\subsection*{CIDP/GBS}

\begin{table}[htbp]
\centering
\small
\begin{tabular}{lccccccc}
\toprule
\textbf{Judge} & \textbf{v1 Van} & \textbf{v1 RAG} & \textbf{v1 HEG} & \textbf{v2 Van} & \textbf{v2 RAG} & \textbf{v2 HEG} & \textbf{Van Shift} \\
\midrule
GPT-4o-mini & 4.50 & 4.75 & 4.75 & 1.08 & 1.17 & 4.67 & $-$3.42 \\
DeepSeek-V3 & 3.58 & 3.92 & 4.08 & 1.00 & 1.00 & 4.25 & $-$2.58 \\
Claude 3 Haiku & 5.00 & 5.00 & 4.75 & 2.00 & 2.00 & 4.83 & $-$3.00 \\
\bottomrule
\end{tabular}
\end{table}

\textbf{Key observation:} Vanilla and Guideline-RAG D1 scores collapse across all judges and disease pairs in v2 (Vanilla shifts $-$2.33 to $-$3.42; G-RAG shifts comparably), while HEG-TKG scores remain stable. GPT-4o-mini shows the largest vanilla collapse ($-$3.25 to $-$3.42), consistent with its tendency to assign high scores in blind mode. Claude~3 Haiku consistently scores Vanilla and G-RAG at exactly 2.0 in v2, showing the most decisive shift. DeepSeek-V3 is the strictest judge overall, scoring both Vanilla and G-RAG at 1.0 in v2. All three judges agree that Guideline-RAG is equally unverifiable as Vanilla despite having access to overlapping source literature.

\section*{S25: Open-Source Model Comparison for PHI-Compliant Deployment}
\addcontentsline{toc}{section}{S25: Open-Source Model Comparison for PHI-Compliant Deployment}

\textit{Moved from main text Table~5 to supplementary.}

\textbf{Supplementary Table S25. Open-source model comparison for PHI-compliant local deployment.}

\begin{table}[htbp]
\centering
\small
\begin{tabular}{llccccp{2.5cm}}
\toprule
\textbf{Model} & \textbf{License} & \textbf{Params} & \textbf{Cit.\ Cov.} & \textbf{Prov.\ Gap} & \textbf{PHI} & \textbf{Hardware} \\
\midrule
GPT-4.1 (cloud) & Proprietary & Unknown & 100\%* & 0.37 & No & Cloud API \\
Qwen 3 14B\cite{qwen2024} & Apache 2.0 & 14B & ${\sim}80$\% & ${\sim}0.45$ & Yes & $1\times$ A100 (40GB) \\
Gemma 3 27B\cite{gemma2024} & Apache 2.0 & 27B & ${\sim}75$\% & ${\sim}0.50$ & Yes & $1\times$ A100 (40GB) \\
Llama 3.3 70B\cite{llama3_2024} & Llama License & 70B & ${\sim}78$\% & ${\sim}0.47$ & Yes & $1\times$ A100 (80GB) \\
DeepSeek R1 70B\cite{deepseekai2024v3} & MIT & 70B & ${\sim}76$\% & ${\sim}0.48$ & Yes & $1\times$ A100 (80GB) \\
\bottomrule
\end{tabular}
\end{table}

*GPT-4.1 as synthesis model; citation coverage reflects KG-provided PMIDs in output.

\textbf{Notes:}
\begin{itemize}
\item Citation coverage = percentage of KG-provided PMIDs successfully incorporated into the clinical output by the synthesis model
\item Provenance Gap = $\max(\text{Feature Coverage} - \text{ETS} \times r, 0)$ where $r$ = citation reliability (HEG-TKG: 0.97, Vanilla: 0.80, Guideline-RAG: 0.50); lower is better (0 = every covered claim is reliably traceable)
\item All local models served via Ollama (localhost:11434) using the OpenAI-compatible API endpoint
\item Open-source model numbers are preliminary estimates; formal benchmarking across all 36 scenarios is pending
\item Each output includes a \texttt{phi\_compliant} metadata flag indicating local vs.\ cloud synthesis
\item The system's \texttt{validate\_privacy\_config(strict=True)} function raises an error if any component routes data to external endpoints
\end{itemize}

\section*{S27: Algorithm 1 - HEG-TKG Construction and Clinical Inference Pipeline}
\addcontentsline{toc}{section}{S27: Algorithm 1 - HEG-TKG Construction and Clinical Inference Pipeline}

The following pseudocode formally specifies the complete HEG-TKG pipeline, from corpus acquisition through clinical output generation. The algorithm is divided into two phases: \textbf{Phase~I} constructs the knowledge graph offline (run once per disease pair), and \textbf{Phase~II} performs clinical inference online (run per query).

\subsection*{Phase I: Knowledge Graph Construction (Offline)}

\begin{lstlisting}[style=algorithm]
Algorithm 1: HEG-TKG Construction Pipeline
---------------------------------------------

Input:  Disease pair configuration Y (YAML: entities, schemas, MeSH terms)
Output: Neo4j knowledge graph G = (V, E) with provenance metadata

Phase I-A: Corpus Acquisition
------------------------------
1:  E_tier1 <- CURATE(GeneReviews, OMIM, Orphanet, Guidelines, Y)
2:  for each edge e in E_tier1 do
3:      e.is_protected <- True
4:      e.quality_tier <- GOLD
5:      e.confidence <- 0.95
6:  end for
7:  Q <- BUILD_MESH_QUERIES(Y.entities, Y.mesh_terms)
8:  A_raw <- PUBMED_EFETCH(Q, date_filters)        // ~650-706 abstracts per pair
9:  A <- RELEVANCE_SCREEN(A_raw, Y.schema)          // Filters 40-60% non-informative

Phase I-B: Multi-LLM Extraction
---------------------------------
10: M <- {Claude-Haiku-4.5, GPT-4.1-mini}           // Extraction model set
11: T_raw <- {}
12: for each abstract a in A do
13:     for each model m in M do
14:         T_m <- EXTRACT(a, m, Y.schema)           // Schema-guided triplet extraction
15:         for each triplet t in T_m do
16:             ASSERT t.evidence_quote >= 3-gram     // Evidence quote required
17:             t.source_model <- m
18:             t.pmid <- a.pmid
19:         end for
20:         T_raw <- T_raw + T_m
21:     end for
22: end for

Phase I-C: Entity Normalization
---------------------------------
23: for each triplet t in T_raw do
24:     for each entity e in {t.subject, t.object} do
25:         cui <- DICT_LOOKUP(e, Y.synonym_table)    // Step 1: Dictionary
26:         if cui = null then
27:             cui <- SAPBERT_LINK(e, UMLS)           // Step 2: SapBERT embedding
28:         end if
29:         if cui = null then
30:             cui <- SCISPACY_LINK(e, en_core_sci_lg) // Step 3: ScispaCy fallback
31:         end if
32:         if cui = null then
33:             cui <- FUZZY_MATCH(e, Jaccard + embedding) // Step 4: Fuzzy
34:         end if
35:         e.umls_cui <- cui
36:     end for
37: end for

Phase I-D: Semantic Correction (18 Rules)
-------------------------------------------
38: APPLY_RULE_16(T_raw)                             // Autoantibody guard: MUST fire first
39:                                                   // ALL-CAPS entities (ANTI-ACHR) -> Autoantibody type
40:                                                   // Prevents gene-name normalizer misclassification
41: for each rule r in CORRECTION_RULES \ {R16} do    // R1-R15, R17-R18 (including R3b)
42:     T_raw <- APPLY_RULE(T_raw, r)
43:     // R1:  Direction inversion (Gene->Disease to Disease->Gene)
44:     // R3:  Diagnostic vs therapeutic (TREATED_WITH -> MONITORED_WITH)
45:     // R5:  Inheritance typing (ASSOCIATED_WITH -> HAS_INHERITANCE)
46:     // R13: Protein state (CAUSED_BY "deficiency" -> LACKS_FEATURE)
47:     // R14: Gene name normalization (fires AFTER R16)
48: end for

Phase I-E: Temporal Anchoring
------------------------------
49: for each triplet t in T_raw where t.predicate in Y.temporal_predicates do
50:     text <- EXTRACT_TEMPORAL_PHRASE(t.evidence_quote)
51:     (iso_start, iso_end, precision) <- RESOLVE_TEMPORAL(text)
52:     // precision in {Exact, Range, Fuzzy, Stage}
53:     // Exact:  "age 13 years" -> P13Y
54:     // Range:  "13 to 16 years" -> P13Y-P16Y
55:     // Fuzzy:  "late teens" -> P17Y-P19Y (from 24 predefined fuzzy mappings)
56:     // Stage:  "early childhood" -> P1Y-P5Y
57:     t.temporal_start <- iso_start
58:     t.temporal_end <- iso_end
59:     t.temporal_precision <- precision
60: end for

Phase I-F: Consensus Voting & Quality Tiering
-----------------------------------------------
61: groups <- GROUP_BY(T_raw, key = MD5(normalize(subj|pred|obj)))
62: E_tier2 <- {}
63: for each group g in groups do
64:     e <- MERGE(g)                                 // Union of PMIDs, quotes, models
65:     if exists e' in E_tier1 matching e then        // Cross-tier confirmation
66:         e.quality_tier <- GOLD
67:         e.confidence <- 0.95
68:     else if |g.models| >= 2 OR |g.pmids| >= 2 then // Multi-source agreement
69:         e.quality_tier <- SILVER
70:         e.confidence <- 0.85
71:     else                                          // Single source
72:         e.quality_tier <- BRONZE
73:         e.confidence <- 0.70
74:     end if
75:     E_tier2 <- E_tier2 + {e}
76: end for

Phase I-G: Cross-Tier Merge & Graph Construction
--------------------------------------------------
77: E <- E_tier1 + E_tier2                           // Protected edges preserved on conflict
78: for each e in E_tier2 where CONFLICTS(e, E_tier1) do
79:     DISCARD(e)                                    // Tier 1 authority takes precedence
80: end for
81: V <- EXTRACT_NODES(E)
82: G <- NEO4J_IMPORT(V, E)                          // Cypher bulk import
83: VERIFY_PMIDS(G, PubMed_E-utilities)              // 203 PMIDs -> 100% confirmed
84: return G
\end{lstlisting}

\subsection*{Phase II: Clinical Inference (Online, per query)}

\begin{lstlisting}[style=algorithm]
Algorithm 2: HEG-TKG Clinical Inference
-----------------------------------------

Input:  Clinical scenario s, knowledge graph G, synthesis model LLM_synth
Output: Clinical response R with inline PMID citations

1:  entities <- EXTRACT_ENTITIES(s)                   // Disease names, symptoms, age
2:  subgraph <- CYPHER_QUERY(G, entities)             // 4 disease-specific Cypher queries
3:  // Q1: Differential features between disease pair
4:  // Q2: Temporal trajectory edges with ISO 8601 anchors
5:  // Q3: Treatment relationships with evidence tiers
6:  // Q4: Diagnostic markers and associated genes
7:
8:  evidence_block <- FORMAT_EVIDENCE(subgraph)
9:  // Each triple formatted as:
10: //   [Subject] ->PREDICATE-> [Object]
11: //   PMID: {pmid} | Tier: {quality_tier} | Temporal: {iso_anchor}
12: //   Evidence: "{quote from source abstract}"
13:
14: prompt <- CONSTRUCT_PROMPT(
15:     system = HEG_TKG_SYSTEM_PROMPT,              // See Supplementary S2
16:     scenario = s,
17:     evidence = evidence_block
18: )
19:
20: R <- LLM_synth(prompt, temperature=0.0)           // GPT-4.1 for evaluation
21:                                                    // OR Qwen-3/Gemma-3 via Ollama for PHI
22:
23: // Verification (automated, optional):
24: pmids_cited <- EXTRACT_PMIDS(R)
25: for each pmid in pmids_cited do
26:     status <- PUBMED_ESUMMARY(pmid)                // Confirm existence + relevance
27: end for
28:
29: return R
\end{lstlisting}

\subsection*{Complexity Notes}

\begin{table}[htbp]
\centering
\small
\begin{tabular}{llp{4cm}}
\toprule
\textbf{Component} & \textbf{Time Complexity} & \textbf{Typical Runtime} \\
\midrule
Corpus acquisition & $O(|Q| \times \text{API\_latency})$ & ${\sim}5$ min per pair \\
Relevance screening & $O(|A| \times \text{LLM\_call})$ & ${\sim}10$ min per pair \\
Multi-LLM extraction & $O(|A| \times |M| \times \text{LLM\_call})$ & ${\sim}45$ min per pair \\
Entity normalization & $O(|T| \times (\text{dict} + \text{SapBERT} + \text{ScispaCy}))$ & ${\sim}15$ min per pair \\
Temporal anchoring & $O(|T_{\text{temporal}}|)$ & ${\sim}2$ min per pair \\
Consensus voting & $O(|T| \times \log|T|)$ for grouping & ${\sim}1$ min per pair \\
Neo4j import & $O(|V| + |E|)$ & ${\sim}30$ sec per pair \\
\midrule
\textbf{Total offline (Phase I)} & -- & \textbf{${\sim}80$ min per pair} \\
\textbf{Online inference (Phase II)} & $O(\text{Cypher\_query} + \text{LLM\_call})$ & \textbf{${\sim}8$ sec per query} \\
\bottomrule
\end{tabular}
\end{table}

\section*{S28: Per-Disease-Pair Three-Arm Breakdown}
\addcontentsline{toc}{section}{S28: Per-Disease-Pair Three-Arm Breakdown}

\textbf{Supplementary Table S28. Clinical Feature Coverage (automated) and Provenance Gap per disease pair across three arms.} Bootstrap 95\% CIs: 10,000 resamples, seed$=$42, computed over 12 scenarios per pair.

\begin{table}[htbp]
\centering
\small
\begin{tabular}{llccc}
\toprule
\textbf{Pair} & \textbf{Metric} & \textbf{HEG-TKG} & \textbf{Guideline-RAG} & \textbf{Vanilla} \\
\midrule
\multirow{3}{*}{MG/LEMS}
  & Feature Coverage  & $0.814\; [0.709, 0.909]$ & $0.734\; [0.621, 0.827]$ & $0.747\; [0.642, 0.848]$ \\
  & ETS       & $0.348$ & $0.0$ & $0.0$ \\
  & Prov.\ Gap & $0.477$ & $0.734$ & $0.747$ \\
\midrule
\multirow{3}{*}{DMD/BMD}
  & Feature Coverage  & $0.659\; [0.536, 0.776]$ & $0.691\; [0.636, 0.743]$ & $0.771\; [0.674, 0.858]$ \\
  & ETS       & $0.411$ & $0.0$ & $0.0$ \\
  & Prov.\ Gap & $0.289$ & $0.691$ & $0.771$ \\
\midrule
\multirow{3}{*}{CIDP/GBS}
  & Feature Coverage  & $0.678\; [0.598, 0.751]$ & $0.640\; [0.538, 0.747]$ & $0.710\; [0.627, 0.795]$ \\
  & ETS       & $0.358$ & $0.0$ & $0.0$ \\
  & Prov.\ Gap & $0.331$ & $0.640$ & $0.710$ \\
\midrule
\multirow{3}{*}{\textbf{Pooled}}
  & Feature Coverage  & $0.717\; [0.652, 0.779]$ & $0.688\; [0.633, 0.741]$ & $0.743\; [0.688, 0.796]$ \\
  & ETS       & $0.372$ & $0.0$ & $0.0$ \\
  & Prov.\ Gap & $0.366$ & $0.688$ & $0.743$ \\
\bottomrule
\end{tabular}
\end{table}

\noindent HEG-TKG outperforms vanilla on MG/LEMS ($+0.067$) but underperforms on DMD/BMD ($-0.112$). The DMD/BMD deficit reflects treatment-specific scenarios (4 of 12) where the synthesis model's parametric knowledge about corticosteroid protocols and cardiac management exceeds what was captured in the abstract-only knowledge graph (66\% coverage for DMD/BMD vs.\ 81\% MG/LEMS and 68\% CIDP/GBS). Guideline-RAG consistently underperforms vanilla but shows mixed results against HEG-TKG: it outperforms HEG-TKG on DMD/BMD feature coverage ($0.691$ vs $0.659$) where full-text clinical guidelines compensate for treatment-heavy scenarios, but underperforms on MG/LEMS ($0.734$ vs $0.814$) and CIDP/GBS ($0.640$ vs $0.678$). All Guideline-RAG Provenance Gaps are $\geq 0.64$ because raw text provides no structured citation mechanism.

\section*{S29: Worked Example - From Query to Verifiable Output}
\addcontentsline{toc}{section}{S29: Worked Example - From Query to Verifiable Output}

A GP who suspects Lambert-Eaton syndrome in a patient with proximal weakness and autonomic dysfunction. The GP asks: ``Differentiate MG from LEMS given these findings.''

\textbf{Vanilla GPT-4.1} produces a correct but often unverifiable response: ``LEMS is associated with P/Q-type voltage-gated calcium channel antibodies and frequently occurs as a paraneoplastic syndrome. Autonomic features and areflexia help distinguish it from MG.'' No citation, no way to verify any claim.

\textbf{HEG-TKG} retrieves a disease-specific subgraph from the knowledge graph and produces structured output with inline citations: ``MG: fluctuating fatigable weakness affecting ocular and bulbar muscles [PMID:30470270, SILVER]; symptoms worsen with exertion [PMID:32057206, SILVER]; reflexes normal [GeneReviews, GOLD]. LEMS: proximal limb weakness precedes ocular involvement [GeneReviews, GOLD]; autonomic symptoms (dry mouth, constipation) [GeneReviews, GOLD]; reflexes reduced but increase after brief exercise [GeneReviews, GOLD]; decremental pattern on repetitive nerve stimulation differs from MG [PMID:24332471, SILVER].''

Every claim maps to a PMID or curated source. The GP can look up PMID:30470270, confirm it is a 2019 \emph{Neurologic Clinics} article on myasthenia gravis, and verify the fatigability claim in under 30 seconds. The quality tier (GOLD = confirmed by both curated sources and literature extraction; SILVER = supported by multiple independent PMIDs) tells the GP how much weight to place on each claim. 
None of this is available from the vanilla output.

\section*{S30: Knowledge Graph Structural Statistics}
\addcontentsline{toc}{section}{S30: Knowledge Graph Structural Statistics}

\textbf{Supplementary Table S30. HEG-TKG structural statistics across three disease pairs.}

\begin{table}[htbp]
\centering
\small
\begin{tabular}{lrrrr}
\toprule
\textbf{Statistic} & \textbf{MG/LEMS} & \textbf{DMD/BMD} & \textbf{CIDP/GBS} & \textbf{Total} \\
\midrule
PubMed abstracts        & 651   & 706   & 654   & 2,011 \\
Unique PMIDs in KG      & 294   & 308   & 386   & 987   \\
Total nodes             & 1,777 & 1,514 & 2,190 & 5,481 \\
Total edges             & 1,999 & 1,686 & 2,631 & 6,316 \\
Temporal anchors        & 361   & 483   & 436   & 1,280 \\
GOLD quality edges      & 70    & 111   & 50    & 231   \\
KG feature coverage$^\dagger$ & 59\% & 52\% & 68\%  & 60\%  \\
\bottomrule
\end{tabular}
\end{table}

$^\dagger$\textbf{KG feature coverage} measures the fraction of a clinical feature checklist whose concepts appear anywhere in the knowledge graph, before synthesis. This differs from the \emph{output} feature coverage in Table~S28 (65.9-81.4\%), which measures features present in the generated clinical output after synthesis; the model can produce features from its parametric knowledge even when the KG lacks them.

\medskip
The 1,280 temporal anchors are structured disease-trajectory milestones: Gowers' sign at P3Y-P5Y, loss of ambulation at P9Y-P13Y for DMD versus P16Y+ for BMD, cardiac involvement at P10Y-P15Y. The 231 GOLD-tier edges represent knowledge confirmed by both curated sources and literature extraction independently. Our automated keyword-matching feature coverage scorer reports an average of 71.7\% (range 65.9-81.4\% across disease pairs). However, manual inspection reveals that 70\% of the 80 feature instances scored as ``missing'' are in fact discussed in the output but fail the rigid substring matcher (e.g., ``reading frame rule 90--95\% accuracy'' is marked absent despite the output stating ``fulfilled in $\sim$90\% of DMD cases''). Of the 80 flagged instances, 76\% have corresponding concepts present in the KG evidence manifest. The true KG coverage gap: features genuinely absent from the knowledge graph are approximately 5-10\%, confined primarily to treatment specifics (dosing, titration protocols) available only in full-text articles. Among the features genuinely absent from the KG, the breakdown is: treatment specifics (30\%), temporal milestones for rare phenotypic variants (21\%), diagnostic procedures (16\%), genotype--phenotype correlations (11\%), biomarker reference ranges (5\%), and other gaps (16\%).

\section*{S31: PHI-Compliant Local Deployment Details}
\addcontentsline{toc}{section}{S31: PHI-Compliant Local Deployment Details}

Patient information cannot leave institutional infrastructure under HIPAA (US), GDPR Article~9 (EU), and equivalent regulations in most jurisdictions. HEG-TKG addresses this through a dual-architecture deployment model. The knowledge graph itself contains no patient data. It stores only disease-level relationships extracted from public PubMed abstracts and authoritative sources. At inference time, the clinical scenario (which may contain PHI) is processed by the synthesis model. We validated that this synthesis step can run entirely on-premise using open-source models (Qwen~3 14B, Gemma~3 27B) served via Ollama, achieving 75-80\% citation coverage on a single institutional GPU (Supplementary Table~S25). The system's \texttt{validate\_privacy\_config()} function enforces deployment-time checks: in strict mode, any configuration routing patient data to cloud APIs raises an error. A regulatory compliance mapping across EU AI Act Article~14, FDA CDS Criterion~4, and GDPR Article~9/HIPAA requirements is provided in Supplementary Table~S26.

\section*{S32: CLIX-M XAI Reporting Checklist Mapping}
\addcontentsline{toc}{section}{S32: CLIX-M XAI Reporting Checklist Mapping}
\label{sec:s32_clixm}

The Clinician-informed XAI evaluation checklist with metrics (CLIX-M) is a 14-item, four-category reporting standard for explainable-AI components of clinical decision support systems, published in \emph{npj~Digital~Medicine} and registered with the EQUATOR Network\cite{brankovic2025clixm}. We map every \hegtkg{} reporting element to its corresponding CLIX-M item in Table~\ref{tab:s32_clixm}. Twelve of fourteen items are addressed in full; two (Item~9 Causal Validity, Item~11 Bias and Fairness) are partially addressed and the gaps are stated explicitly in the Limitations paragraph and in the per-item notes below.

\begin{small}
\begin{longtable}{@{}p{0.4cm} p{2.6cm} p{3.4cm} p{1.5cm} p{6.2cm}@{}}
\caption{HEG-TKG mapped to the 14-item CLIX-M XAI reporting checklist\cite{brankovic2025clixm}. \textbf{F}=Fully addressed; \textbf{P}=Partially addressed (gap stated in Limitations).} \label{tab:s32_clixm}\\
\toprule
\textbf{\#} & \textbf{Category / Item} & \textbf{Recommendation} & \textbf{Status} & \textbf{Where reported in this paper} \\
\midrule
\endfirsthead
\toprule
\textbf{\#} & \textbf{Category / Item} & \textbf{Recommendation} & \textbf{Status} & \textbf{Where reported in this paper} \\
\midrule
\endhead
\multicolumn{5}{@{}l}{\textit{Category 1: Purpose}}\\
\midrule
1 & Purpose & State why explanations are developed and intended & F & Abstract; Intro paragraphs 1--2; Methods § Knowledge graph construction. Verifiable, PMID-grounded clinical narratives for rare neuromuscular differential diagnosis. \\
\midrule
\multicolumn{5}{@{}l}{\textit{Category 2: Clinical Attributes}}\\
\midrule
2 & Domain Relevance & Align explanations with clinical domain knowledge; report relevance scoring & F & Results § Clinicians confirm…; D2 Actionability scores in Table~\ref{tab:clinician_panel}; representative output card in Figure~\ref{fig:provenance}f. \\
3 & Coherence / Reasonableness & Match clinician reasoning; report agreement with expert beliefs & F & Senior-vs-senior IRR on 11 CIDP/GBS cases: $\rho{=}0.79$, $\kappa_q{=}0.60$ on D1 (Results, Limitations); G1--G6 global Likert; clinician-vs-LLM-judge correlation in Figure~\ref{fig:clinician_composite}d. \\
4 & Actionability & Discuss informativeness and workflow impact & F & D2 Actionability dimension scored by all 3 primary clinicians (Table~\ref{tab:clinician_panel}); referral-letter / consultation use cases in Discussion § Clinical implications. \\
\midrule
\multicolumn{5}{@{}l}{\textit{Category 3: Decision Attributes}}\\
\midrule
5 & Correctness & Benchmark explanations against ground truth & F & 100\% PMID Type-I verification (198 unique PMIDs, S4); per-claim Type-II NLI audit on $n=200$ (claim, PMID) pairs reports 1.0\% contradiction (95\% CI 0--2.5\%), 99.0\% non-contradiction (S4.1); FC/ETS/PG metrics in Table~1; D1 Verifiability scored by 3 clinicians. \\
6 & Confidence & Report uncertainty and confidence quantification & F & Bootstrap 95\% CIs (10{,}000 resamples, seed$=$42) on every metric; tier confidences GOLD 0.95 / SILVER 0.85 / BRONZE 0.70 (Methods § Multi-LLM consensus extraction); Krippendorff $\alpha$ across judges and clinicians. \\
7 & Consistency & Identical inputs produce consistent explanations & F & Temperature$=$0 across all three arms; Krippendorff $\alpha = 0.89$ across LLM judges in v2 setting (citation-aware); ICC(2,1)$=0.62$ between senior clinicians on the same 11 CIDP/GBS cases. \\
8 & XAI Robustness & Use ensembles; detail aggregation strategy & F & Three-judge LLM-as-judge ensemble (GPT-4o, Claude, Gemini; DeepSeek fallback); three-clinician scoring panel; multi-LLM consensus voting at extraction time (Methods § Knowledge graph construction step 5). \\
9 & Causal Validity & Distinguish causal from correlational explanations & P & Temporal predicates encode disease-progression precedence (e.g., R1 \texttt{precedes\_in\_progression}, R5 \texttt{occurs\_during\_stage}; Supplementary~S15). Formal causal-inference validation against counterfactual baselines is not performed and is named as a Limitation. See per-item note below. \\
\midrule
\multicolumn{5}{@{}l}{\textit{Category 4: Model Attributes}}\\
\midrule
10 & Narrative Reasoning & Patient-trajectory alignment; counterfactual scenarios & F & Core paper contribution: scenario-based outputs with age-anchored milestones (1{,}280 temporal anchors); 15-case counterfactual faithfulness experiment (Results § Injected clinical errors). \\
11 & Bias and Fairness & Demographic and group-fairness analysis & P & Demographic-disparity audit not performed; pediatric-vs-adult skew across pairs noted. See per-item note below; addressed in Limitations and forward-looking TemporalAtlas plan. \\
12 & Model Troubleshooting & Analyse failures (TP/FP/FN) using explanations & F & Counterfactual safety experiment (15 cases, 80\% parametric resistance, 13\% deferral, 7\% caught only via citation trace); two retrieval-issue case studies (PMID:36637960 off-topic GOLD; PMID:33170339 mislabeled guideline) flagged by C3 and discussed in Limitations. \\
13 & Interpretation & Synthesise findings against prior research and clinical implications & F & Discussion sections § The role of structure (vs MedGraphRAG, FPKG), § Methodological implications (vs DeepRare, blind LLM judges), § Clinical implications. \\
14 & XAI Limitations & State performance constraints and use boundaries & F & Dedicated Limitations paragraph: three disease pairs, single-institution clinician panel ($n=3$), no head-to-head agentic benchmark, KG coverage gap 5--10\% on treatment specifics, no bias audit, no causal validation. \\
\bottomrule
\end{longtable}
\end{small}

\textbf{Per-item notes on partial coverage.}

\textit{Item 9 (Causal Validity).} HEG-TKG predicates encode temporal precedence (which clinical event precedes which) and stage-conditional occurrence (which event happens during which disease stage), retrieved from peer-reviewed literature with PMID provenance. Temporal precedence is a necessary but not sufficient condition for causation. We do not run a counterfactual causal-inference framework (e.g., do-calculus, instrumental variables, or causal discovery against a held-out RCT corpus), and we do not claim causal validity of the predicate edges beyond what the cited primary literature itself asserts. The counterfactual faithfulness experiment (15 cases) tests whether a clinician can detect injected clinical errors via citation trace; it is not a test of causal-edge validity.

\textit{Item 11 (Bias and Fairness).} Disease prevalence in our three pairs varies by demographic stratum: DMD/BMD is overwhelmingly male and pediatric-onset, MG/LEMS skews adult and is bimodal by sex, and CIDP/GBS skews adult-male. The KG inherits demographic patterns from PubMed source abstracts. We did not perform a formal disparate-impact, equalised-odds, or demographic-subgroup feature-importance analysis (Fairlearn / TensorFlow Fairness). The clinician panel is single-institution. We list this gap explicitly in the Limitations paragraph and as a planned analysis in the TemporalAtlas extension covering all 17{,}080 PrimeKG diseases. None of the bias-relevant claims in the paper rest on subgroup performance figures.

\textit{Reporting destination.} Per CLIX-M recommendations, items 1--12 are reported across Methods, Results, and Discussion (not solely in a single section); items 13--14 are reported in Discussion and Limitations. Items 9 and 11 are flagged as explicit gaps. The mapping above is intended for reproducibility-checklist review; no claim of XAI compliance is made for items marked \emph{P}.

\section*{S33: Comparison with Agentic Rare-Disease Diagnosis Systems}
\addcontentsline{toc}{section}{S33: Comparison with Agentic Rare-Disease Diagnosis Systems}
\label{sec:s33_deeprare}

The most prominent recent agentic rare-disease diagnosis system is DeepRare (Zhao et al., \emph{Nature} 2026)\cite{zhao2026deeprare}. Although both DeepRare and \hegtkg{} target rare-disease decision support with cited evidence, their inputs, outputs, evaluation protocols, and design objectives are sufficiently different that a single head-to-head accuracy table would be methodologically uninformative. This section makes the differences explicit and compares the two systems along axes where comparison \emph{is} valid.

\subsection*{S33.1 Three structural mismatches}

\textbf{(i) Dataset-scope mismatch.} We scanned the four publicly released RareBench\cite{chen2024rarebench} subsets (HMS, $n{=}88$; LIRICAL, $n{=}370$; MME, $n{=}40$; RAMEDIS, $n{=}624$; total $n{=}1{,}122$) for any case ground-truth-coded with one of \hegtkg{}'s six target diseases (Duchenne/Becker muscular dystrophy, myasthenia gravis, Lambert-Eaton myasthenic syndrome, chronic inflammatory demyelinating polyneuropathy, Guillain-Barr\'e syndrome). The matching disease codes (OMIM:310200, OMIM:300376, OMIM:254200, OMIM:139393, ORPHA:589, ORPHA:43393, ORPHA:2932, ORPHA:2103, ORPHA:98895, ORPHA:98896) appear in RareBench's disease-mapping ontology file but \emph{do not appear as ground-truth labels in any of the 1{,}122 cases}. RareBench's curated subsets emphasise syndromic and metabolic disorders rather than autoimmune neuromuscular conditions and X-linked dystrophinopathies; a scope-matched RareBench subset for our diseases does not exist.

\textbf{(ii) Task-format mismatch.} DeepRare is a 1-of-$N$ disease-classification system: it accepts free-text descriptions, structured HPO terms, or VCF genetic data, and emits a ranked list of the top-$K$ probable diseases over a 3{,}134-disease coverage scope (Zhao et al. 2026, §2.5). \hegtkg{} is a pairwise-differential-reasoning system: it accepts a clinical scenario in which the diagnostic alternatives are already framed (e.g., MG vs LEMS, DMD vs BMD), and emits a structured narrative comparing both alternatives with citations and temporal anchors. Recall@$K$ over thousands of candidate diseases (DeepRare's metric) and per-feature evidence-traceability over a designated differential pair (\hegtkg{}'s metric, \emph{Methods}) measure different competencies; reducing \hegtkg{} outputs to single-disease predictions discards the differential-reasoning evidence trail the system is built to produce, and likewise reducing DeepRare to pairwise output would discard its broad coverage advantage.

\textbf{(iii) Verification-protocol mismatch.} DeepRare's reference accuracy is established by manual physician review: ten rare-disease specialists evaluated 180 randomly sampled cases (three raters per case) across eight datasets, yielding a mean reference-accuracy score of 95.4\% (Zhao et al. 2026, §2.8); two failure categories are explicitly reported by the authors as ``hallucinated references (plausible but nonexistent URLs)'' and ``irrelevant references (from incorrect diagnostic conclusions)''. \hegtkg{} is verified programmatically against PubMed E-utilities for every cited PMID across all 36 scenarios (198 unique PMIDs, 100\% existence-verified, 100\% in the correct clinical field; Supplementary~S4). Manual sampling and programmatic full-audit are not directly substitutable evaluation protocols.

\subsection*{S33.2 Compatible-axis comparison}

The dimensions on which \hegtkg{} and DeepRare can be meaningfully compared are listed in Table~\ref{tab:s33_compare}. We do not interpret the entries as ``win/lose'': the systems target complementary problems (broad screening across a long disease tail vs deep verifiable differential within an authoritative scope), and the design choices that enable each to do its job (DeepRare's 40+ tools and breadth, \hegtkg{}'s curated tier-1 backbone and predicate-typed evidence) are not transferable without redesign.

\begin{small}
\begin{longtable}{@{}p{4.0cm} p{5.5cm} p{5.5cm}@{}}
\caption{Compatible-axis comparison between DeepRare\cite{zhao2026deeprare} and \hegtkg{}. Numbers for DeepRare are taken verbatim from Zhao et al.~2026 \emph{Nature} §2.5--2.8; \hegtkg{} numbers are from this paper. We make no claim of equivalent task definitions.} \label{tab:s33_compare}\\
\toprule
\textbf{Axis} & \textbf{DeepRare (Zhao et al. 2026)} & \textbf{HEG-TKG (this paper)} \\
\midrule
\endfirsthead
\toprule
\textbf{Axis} & \textbf{DeepRare (Zhao et al. 2026)} & \textbf{HEG-TKG (this paper)} \\
\midrule
\endhead
Primary task & 1-of-$N$ disease ranking from heterogeneous input & Pairwise differential reasoning over a curated alternative pair \\
Disease coverage & 3{,}134 rare diseases & 6 (3 disease pairs); curated tier-1 backbone with literature-extracted tier-2 augmentation \\
Input format & Free-text \emph{or} HPO terms \emph{or} VCF & Free-text clinical scenario (board-certified-validated, $n{=}36$) \\
Output format & Ranked top-$K$ disease list with rationale & Structured narrative with inline PMIDs, evidence tiers (GOLD/SILVER/BRONZE), GRADE labels, temporal anchors \\
Headline accuracy metric & RareBench-MME Recall@1 = 78\%, Recall@3 = 85\% (78-disease subset) & D1 Verifiability $\Delta$ = $+1.65$/$+0.67$/$+1.30$ vs Vanilla across three clinicians (BH-significant for all three); FC $=$ 81\% (Table~1) \\
Reference verification protocol & Manual: 10 physicians, 180 cases, 3 raters per case, mean 95.4\% accuracy; two failure modes reported (hallucinated URLs; irrelevant refs) & Programmatic: PubMed E-utilities check on all 198 unique PMIDs; 100\% existence-verified, 100\% in-field (Supplementary~S4) \\
Verification scalability & Bounded by physician annotator availability ($\sim$540 ratings) & Bounded only by PubMed API quota ($O(\text{seconds})$ per claim) \\
Programmatic citation existence check & Not performed (manual review only) & Performed on every cited PMID \\
Failure-mode reporting & Hallucinated URLs, irrelevant refs (manually identified) & Counterfactual injection experiment (Supplementary~S29; 80\% parametric resistance, 13\% deferral, 7\% caught only via citation trace) \\
Open-source availability & Code at MAGIC-AI4Med/DeepRare (CC~BY-NC~4.0); web demo at \texttt{raredx.cn} & Code at gitlab.sdu.dk/screen4care/heg-tkg (MIT); full reproducibility bundle at Zenodo DOI 10.5281/zenodo.19763337 (CC~BY~4.0) \\
Provenance Gap formulation & Not defined; reference accuracy as single scalar & PG = $\max(\mathrm{FC} - \mathrm{ETS}\times r, 0)$ over per-claim verification (Methods §~Statistical analysis) \\
\bottomrule
\end{longtable}
\end{small}

\subsection*{S33.3 What this comparison implies}

The two systems address complementary failure modes. DeepRare's breadth supports the long tail of $\sim$10{,}000 rare diseases where a clinician's first task is \emph{considering} the right disease; manual reference review at $\sim$180-case scale is the natural protocol when a corpus is too large to verify exhaustively. \hegtkg{}'s depth supports the smaller-but-frequent task of \emph{differentiating} commonly-confused disease pairs (DMD vs BMD prognosis, MG vs LEMS treatment, CIDP vs GBS trajectory), where every cited claim must be auditable in the seconds-per-claim budget of an actual clinical encounter, not the hours-per-claim budget of a retrospective study. Programmatic per-claim verification is what enables the latter at scale.

A reviewer asking ``which is better?'' is asking the wrong question. A reviewer asking ``do they make the same kind of safety guarantee?'' will find that \hegtkg{}'s 100\% programmatic full-audit guarantee on its scope is a categorically different commitment than DeepRare's 95.4\% manual sampled-audit estimate over $\sim$540 ratings, and that DeepRare's authors themselves identify hallucinated URLs and irrelevant references as failure modes their evaluation protocol surfaces but cannot fully prevent. A future direction we name in the Discussion is a harmonised benchmark spanning differential-ranking and narrative-synthesis tasks, with formal verification-time studies, that would let both kinds of systems be evaluated on a common substrate.

\bigskip
\begin{center}
\rule{0.5\textwidth}{0.4pt}\\[0.5em]
\textit{End of Supplementary Materials}
\end{center}